\RequirePackage{fix-cm}
\documentclass[twocolumn]{svjour3}          
\smartqed  
\usepackage{graphicx}
\journalname{Int J Comput Vis}
\usepackage{hhline}
\usepackage{hyperref} 
\usepackage{color}
\usepackage{colortbl} 
\usepackage{amsmath} 
\usepackage{amssymb}
\definecolor{light-gray}{gray}{0.6}
\usepackage{multirow}
\newcommand\norm[1]{\left\lVert#1\right\rVert}

\newcommand{\revise}[1]{\textcolor{black}{#1}}

\usepackage{amsmath}
\usepackage[caption=false,font=footnotesize]{subfig}
 
\usepackage{stfloats}
\usepackage{fixltx2e}
\usepackage{array}
\usepackage{soul}
\usepackage{natbib}
\usepackage[misc]{ifsym}

\begin{document}
\newcolumntype{L}[1]{>{\raggedright\arraybackslash}p{#1}}
\newcolumntype{C}[1]{>{\centering\arraybackslash}p{#1}}
\newcolumntype{R}[1]{>{\raggedleft\arraybackslash}p{#1}}

\title{ARBEE: Towards Automated Recognition of Bodily Expression of Emotion In the Wild 
}


\author{Yu Luo\and
Jianbo Ye\and
Reginald B. Adams, Jr.\and
Jia Li\and
Michelle~G.~Newman\and
James~Z.~Wang
}


\institute{\footnotesize
Y. Luo~(\Letter) \and J. Ye\and J.Z. Wang~(\Letter) \at
College of Information Sciences and Technology,\\
The Pennsylvania State University, University Park, PA, USA.\\
\email{yzl5709@psu.edu}        
\and
J. Ye\at
\email{yelpoo@gmail.com}\\        
Ye is currently with Amazon Lab126, Sunnyvale, CA, USA. 
\and
J.Z. Wang\at
\email{jwang@psu.edu}        
\and
R.B. Adams Jr.\and M.G. Newman \at
Department of Psychology,\\
The Pennsylvania State University,
University Park, PA, USA.\\
\email{radams@psu.edu}        
\and
M.G. Newman\at
\email{mgn1@psu.edu}        
\and
J. Li \at
Department of Statistics,\\
The Pennsylvania State University,
University Park, PA, USA.\\
\email{jiali@psu.edu}   
}

\date{Received: date / Accepted: date}

\maketitle

\begin{abstract}
Humans are arguably innately prepared to comprehend others' emotional expressions from subtle body movements. If robots or computers can be empowered with this capability, a number of robotic applications become possible. Automatically recognizing human bodily expression in unconstrained situations, however, is daunting 
given the incomplete understanding of the relationship between emotional expressions and body movements. The current research, as a  multidisciplinary effort among computer and information sciences, psychology, and statistics, proposes a scalable and reliable crowdsourcing approach for collecting in-the-wild perceived emotion data for computers to learn to recognize body languages of humans.
To accomplish this task, a large and growing annotated dataset with 9,876  video clips of body movements and 13,239 human characters, named BoLD (Body Language Dataset), has been created. Comprehensive statistical analysis of the dataset revealed many interesting insights. A system to model the emotional expressions based on bodily movements, named ARBEE (Automated Recognition of Bodily Expression of Emotion), has also been developed and evaluated. Our analysis shows the effectiveness of Laban Movement Analysis (LMA) features in characterizing arousal, and our experiments using \revise{LMA features} further demonstrate computability of bodily expression. 
\revise{We report and compare results of several other baseline methods which were developed for action recognition based on two different modalities, body skeleton and raw image.}
The dataset and findings presented in this work will likely serve as a launchpad for future discoveries in body language understanding that will enable future robots to interact and collaborate more effectively with humans.
\end{abstract}


\section{Introduction}\label{sec:introduction}
Many future robotic applications, including personal assistant robots, social robots, and police robots demand close collaboration with and comprehensive understanding of the humans around them. Current robotic technologies for understanding human behaviors beyond their basic activities, however, are limited. 
Body movements and postures encode rich information about a person's status, including their awareness, intention, and emotional state~\citep{shiffrar2011seeing}. Even 
at a young age, humans can ``read'' another's body language, decoding movements and facial expressions as emotional keys. {\it How can a computer program be trained to recognize human emotional expressions from body movements?} This question drives our current research effort.

Previous research on computerized body movement analysis has largely focused on
recognizing human activities ({\it e.g.}, the person is running). Yet, a person's emotional 
state is another important characteristic that is often conveyed through  
body movements. Recent studies in psychology have suggested that movement and
postural behavior are useful features for identifying human 
emotions~\citep{wallbott1998bodily,meeren2005rapid,de2006towards,aviezer2012body}. 
For instance, researchers found that human participants of a study 
could not correctly identify facial expressions associated with winning or losing a point in a professional tennis game when facial images were presented alone, whereas they were
able to correctly identify this distinction with images of just the body or images that included both the body
and the face~\citep{aviezer2012body}. More interestingly, when the face part of an
image was paired with the body and edited to an opposite
situation face ({\it e.g.}, winning face paired with losing body), people still used the body to identify the outcome. A valuable insight
from this psychology study is that the human body may be more diagnostic than the
face in terms of emotion recognition. In our work, {\it bodily expression} is
defined as  human affect expressed by body movements and/or postures. 

Our earlier work studied the computability of evoked emotions~\citep{lu2012shape,luinvestigation,ye2017probabilistic} from visual stimuli using
computer vision and machine learning. In this work, we investigate whether bodily
expressions are computable. In particular, we explore whether modern computer 
vision techniques can match the cognitive ability of typical humans in recognizing 
bodily expressions in the wild, {\it i.e.}, from real-world unconstrained situations.

\begin{figure}
\includegraphics[height=1in, trim=20 0 0 0, clip]{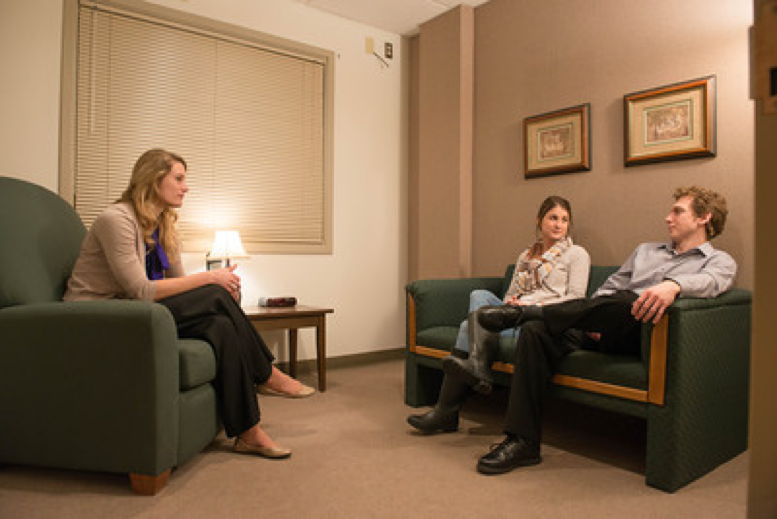}
\includegraphics[height=1in, trim=100 0 70 0, clip]{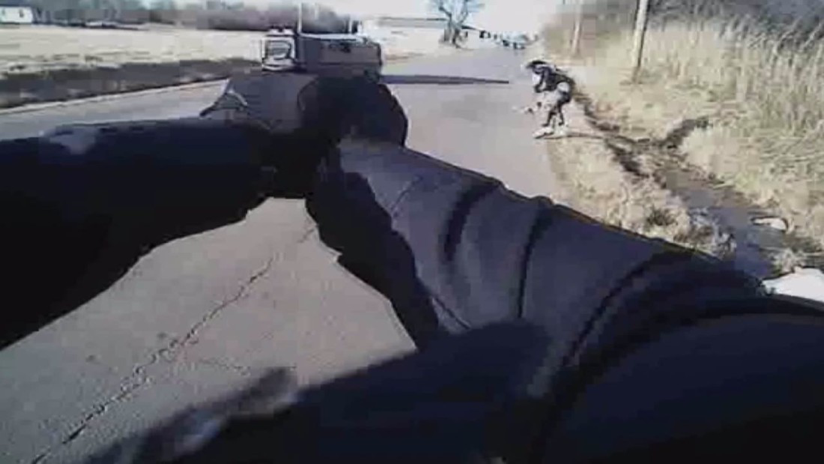}
\includegraphics[height=1in]{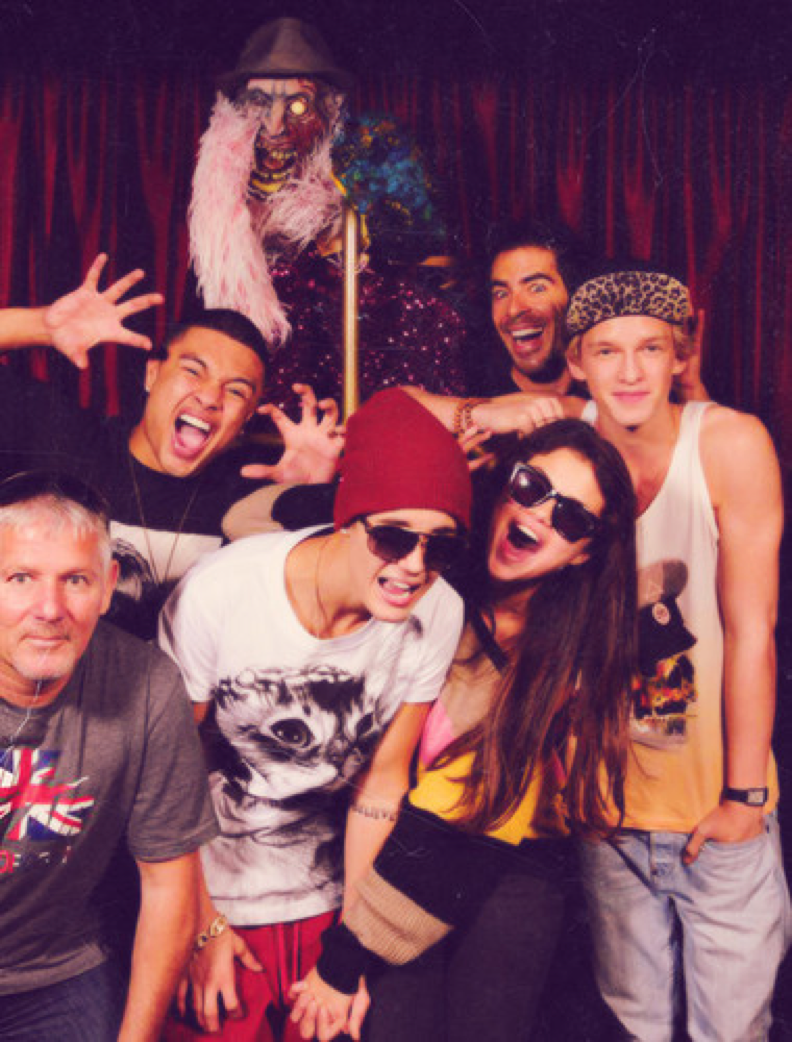}
\caption{Examples of possible scenarios where computerized bodily expression recognition can be useful.
From left to right: psychological clinic assistance, public safety and law enforcement, and social robot or social media.}
\end{figure}

Computerized bodily expression recognition capabilities have the potential
to enable a large number of innovative applications including information
management and retrieval, public safety, patient care, and
social media~\citep{krakovsky2018artificial}. For instance, such systems
can be deployed in public areas such as airports, metro or bus
stations, or stadiums to help police identify potential threats. Better results might 
be obtained in a population with a high rate of emotional instability.
A psychology clinic, for example, may install such systems to
help assess and evaluate disorders, {including} anxiety and depression{, either} to predict danger to self and others from patients, or to track
the progress of patients over time. 
Similarly, police may use such technology to help assess the identity
of suspected criminals in naturalistic settings and/or their emotions
and deceptive motives during an interrogation. Well-trained and
experienced detectives and interrogators rely on a combination of body
language, facial expressions, eye contact, speech patterns,
and voices to differentiate a liar from a truthful person. An effective
assistive technology based on emotional understanding could substantially
reduce the stress of police officers as they carry out their work.
{Improving the bodily expression recognition of assistive robots will enrich human-computer interactions.} Future assistive robots can better assist those who may suffer 
emotional stress or mental illness, {\it e.g.,} assistive robots may detect early warning signals of manic episodes. In social media, recent popular social 
applications such as Snapchat and Instagram allow users to upload short clips
of self-recorded and edited videos. A crucial analysis from an advertising
perspective is to better identify the intention of a specific uploading act by understanding
the emotional status of a person in the video. For example, a user who wants to share
the memory of traveling with his family would more likely upload a video capturing the best interaction moment
filled with joy and happiness. Such analysis helps companies to
better {personalize} the services or to provide advertisement {more effectively} for their users, {\it e.g.,} through showing travel-related products or services as opposed to business-related ones. 

Automatic bodily expression recognition as a research problem is highly {\bf challenging} for three primary reasons.
First, it is difficult to collect a bodily expression dataset with high quality 
annotations. The understanding and perception of emotions from concrete 
observations is often subject to context, interpretation, ethnicity and culture. 
There is often no gold standard label for emotions, especially for bodily expressions.
In facial analysis, the expression could be encoded with movements of individual  muscles, a.k.a., Action Units (AU) in facial action coding system (FACS)~\citep{ekman1977facial}. However, psychologists have not developed an analogous notation system that directly 
encodes correspondence 
between bodily expression and body movements. {This} lack of such 
empirical guidance {leaves} even professionals {without}  
complete agreement {about} annotating 
bodily expressions. To date, research on bodily expression is limited to acted and constrained lab-setting video data~\citep{gunes2007bi,kleinsmith2006cross,schindler2008recognizing,dael2012emotion}, 
which are usually of small size due to lengthy human subject study {regulations}.
Second, bodily expression is subtle and composite. 
According to~\citep{karg2013body}, body movements have three categories, functional
movements ({\it e.g.} walking), artistic movements\break({\it e.g.} dancing), and communicative 
movements ({\it e.g.} gesturing while talking). In a real-world setting, bodily expression
can be strongly coupled with functional movements. For example, people may
represent different emotional states in the same functional movement, {\it e.g.} walking.
Third, an articulated pose has many degrees of freedom. Working with real-world video 
data poses additional technical challenges {such as the} high level of heterogeneity in people’s 
behaviors, the highly cluttered background, and {the often substantial differences in} scale, camera perspective, and 
pose of the person in the frame. 

In this work, we investigate the feasibility of crowdsourcing bodily expression data collection and study the
computability of bodily expression using the collected data. We summarize the primary {\bf contributions} as follows.
\begin{itemize}
\item We propose a scalable and reliable crowdsourcing pipeline for collecting in-the-wild perceived emotion data.
With this pipeline, we collected a large dataset with 9,876 clips that have body movements and over 13,239 human characters. We named the dataset the {\bf BoLD} ({\bf Bo}dy {\bf L}anguage {\bf D}ataset). Each short video
clip in BoLD has been annotated for emotional expressions as perceived by the viewers. To our knowledge, BoLD is the first large-scale video dataset for 
bodily emotion in the wild. 
\item We conducted comprehensive agreement analysis on the crowdsourced annotations. The results demonstrate the
validity of the proposed data collection pipeline. We also evaluated human performance on
emotion recognition on a large and highly diverse population. Interesting insights have been found in these analyses.
\item We investigated Laban Movement Analysis (LMA) features and action recognition-based methods using the BoLD
dataset. From our experiments, hand acceleration shows strong correlation with one particular
dimension of emotion --- arousal, a result that is intuitive. We further show that existing action recognition-based models can yield promising results. Specifically, deep models achieve remarkable performance on emotion recognition tasks.
\end{itemize}

\revise{In our work, we approach the bodily expression\break recognition problem with the focus of addressing the first challenge mentioned earlier. Using our proposed data collection pipeline, we have collected high quality affect annotation. With the state-of-the-art computer vision techniques, we are able to address the third challenge to a certain extent. To properly address the second challenge, regarding the subtle and composite nature of bodily expression, requires breakthroughs in computational psychology. Below,
we detail some of the remaining technical difficulties on the bodily expression recognition problem that the computer vision community can potentially address. }

\revise{
Despite significant progress recently in 2D/3D pose estimation~\citep{cao2017realtime, martinez2017simple}, these techniques are limited compared with Motion Capture (MoCap) systems, which rely on placing active or passive optical markers on the subject's body to detect motion, because of two issues. First, these vision-based estimation methods are noisy in terms of the jitter errors~\citep{ruggero2017benchmarking}. While high accuracy has been reported on pose estimation benchmarks, the criteria used in the benchmarks are not designed for our application which demands substantially higher precision of landmark locations. Consequently, the errors in the results generated through those methods propagate in our pipeline, as pose estimation is a first-step in analyzing the relationship between motion and emotion.}

\revise{Second, vision-based methods ({\it e.g.},~\citet{martinez2017simple}) usually address whole-body poses, which have no missing landmarks, and only produce relative coordinates of the landmarks from the pose ({\it e.g.}, with respect to the barycenter of the human skeleton) instead of the actual 
coordinates in the physical environment. In-the-wild videos, however, often contain upper-body or partially-occluded poses. Further, the interaction between human and the environment, such as a lift of the person's barycenter or when the person is pacing between two positions, is often critical for bodily expression recognition. Additional modeling on the environment together with that for the human would be useful in understanding body movement. }

\revise{In addition to these difficulties faced by the computer vision community broadly, the computation psychology community also needs some breakthroughs. For instance, state-of-the-art end-to-end action recognition methods developed in the computer vision community offer insufficient interpretability of bodily expression. While the LMA features that we have developed in this work has better interpretability than the action recognition based methods, to completely address the problem of body language interpretation, we believe it will be important to have comprehensive motion protocols defined or learned, as a counterpart of FACS for bodily expression.}

The rest of this paper is structured as follows. Section~\ref{sec:related} reviews related work in the literature. The data collection pipeline and statistics of the BoLD dataset are introduced in Section~\ref{sec:dataset}. We describe our modeling processes on BoLD and demonstrate findings in Section~\ref{sec:method}, and conclude in Section~\ref{sec:conclusion}.

\section{Related Work} \label{sec:related}
{After} first reviewing basic concepts on bodily expression and related datasets, we then discuss related work on crowdsourcing subjective affect annotation and automatic bodily expression modeling.
\subsection{Bodily Expression Recognition}
Existing automated bodily expression recognition studies mostly build on two theoretical
models for representing affective states, the {\it categorical} and the {\it dimensional}
models. The categorical model represents affective states into several emotion categories. In~\citep{ekman1986new,ekman1992there}, Ekman {\it et al.} proposed six basic emotions, {\it i.e.},
anger, happiness, sadness, surprise, disgust, and fear. However, as suggested 
by~\citet{carmichael1937study} and~\citet{karg2013body}, bodily expression is not limited to basic emotions.
When we restricted {interpretations} to only basic emotions at a preliminary data collection pilot study, 
the participants provided feedback
that they often found none of the basic emotions as suitable for the given video sample.
A dimensional model of affective states is the PAD model by~\cite{mehrabian1996pleasure}, which describes an 
emotion in three dimensions, pleasure (valence), arousal, and dominance. {In the PAD model,} valence characterizes~the positivity versus negativity of an emotion, {while} arousal characterizes the level of activation and energy of an emotion,
{and} dominance characterizes the extent of controlling others or surroundings. As summarized 
in \citep{karg2013body,kleinsmith2013affective}, most bodily expression-related studies
focus on {either} a small set of categorical emotions or two dimensions of valence and arousal in the PAD model. In our work, we adopt both measurements in order to acquire complementary emotion annotations.

Based on how emotion is generated, emotions can be categorized into acted or elicited emotions, and 
spontaneous emotions. Acted emotion refers to actors' performing a certain emotion under given
contexts or scenarios. Early work was mostly built on acted emotions \citep{wallbott1998bodily,dael2012emotion,gunes2007bi,schindler2008recognizing}. 
\citet{wallbott1998bodily} analyzes videos recorded on recruited actors and established bodily 
emotions as an important modality of emotion recognition. 
In~\citep{douglas2007humaine}, a human subject's emotion is elicited via interaction with computer avatar of its operator. 
\citet{luinvestigation} crowdsourced emotion responses with image stimuli. Recently, natural or authentic emotions have generated more interest in the research
community. In~\citep{kleinsmith2011automatic}, body 
movements are recorded while human subjects play body movement-based video games. 

Related work can be categorized based on raw data types, namely MoCap data or 
image/video data. For lab-setting studies such as \citep{kleinsmith2006cross,kleinsmith2011automatic,aristidou2015emotion}, collecting motion 
capture data is usually feasible. \citet{gunes2007bi} collected a dataset
with upper body movement video recorded in a studio. Other work \citep{gunes2007bi,schindler2008recognizing,douglas2007humaine} used image/video data capturing the frontal view of the poses. 

Humans perceive and understand emotions from multiple modalities, such as face, body language,
touch, eye contact, and vocal cues. We review the most related vision-based facial expression analysis
here. Facial expression is an important modality in emotion recognition and automated facial
expression recognition is more successful compared with other modalities. The main reasons for this success are 
two-fold. First, the discovery of FACS made facial expression less subjective. Many recent works 
on facial expression recognition focus on Action Unit detection, {\it e.g.}, \citep{eleftheriadis2015discriminative,fabian2016emotionet}. Second, the face has {fewer} degrees of 
freedom compared with the whole body \citep{schindler2008recognizing}. To address the comparatively broader freedom of bodily movement, \citet{karg2013body} suggest
the use of a movement notation system may help {identify} bodily expression. {Other research has considered microexpressions,} {\it e.g.}, \citep{xu2017microexpression}, suggesting additional nuances in facial expressions.
{To our knowledge, no} vision-based study or dataset on complete measurement of
natural bodily emotions {exists}.

\subsection{Crowdsourced Affect Annotation}
Crowdsourcing from the Internet as a data collection process has been originally proposed to collect objective, non-affective
data and received popularity in the machine learning community to acquire large-scale\break ground~truth~datasets. 
A school of data quality control methods has been
proposed for crowdsourcing. Yet, crowdsourcing affect annotations 
is highly challenging due to the intertwined subjectivity of affect
and uninformative participants. Very few studies report on the limitations and complexity of crowdsourcing affect annotations. 
As suggested by \citet{ye2017probabilistic}, inconsistency of crowdsourced affective
data exists due to two factors. The first is the possible {untrustworthiness} of recruited participants due to the discrepancy between
the purpose of study (collecting high quality data) and the incentive for participants (earning cash rewards). The second
is the natural variability of humans’ perceiving others' affective expressions, as was discussed earlier.
\citet{biel2013youtube} crowdsourced personality attributes. Although they 
analyzed agreements among different participants, they did not conduct quality control, 
catering to the two stated factors in the crowdsourcing. 
\citet{kosti2017emotion}, however, used an {\it ad hoc} gold standard to control {annotation} 
quality and each sample in the training set was only annotated once.
\citet{luinvestigation} crowdsourced evoked emotions of stimuli images. Building on \citet{luinvestigation}, 
\citet{ye2017probabilistic} proposed a probabilistic model, named the GLBA, to jointly model each worker's reliability and regularity --- the two factors contributing
to the inconsistent annotations --- in order to improve the quality of affective data collected.
Because the GLBA methodology is applicable for virtually 
any crowdsourced affective data, 
we use it for our data quality control pipeline as well. 

\subsection{Automatic Modeling of Bodily Expression}

\revise{Automatic modeling of bodily expression (AMBE) typically requires three steps: human detection, pose estimation and tracking, and representation learning. In such a pipeline, human(s) are detected frame-by-frame in a video and their body landmarks are extracted by a pose estimator. Subsequently, if multiple people appear in the scene, the poses of the same person are associated along all frames~\citep{iqbal2017posetrack}. With each person's pose identified and associated across frames, an appropriate feature representation of each person is extracted.}

\revise{Based on the way data is collected, we divide AMBE methods into video-based and non-video-based. For\break video-based methods, data are collected from a camera, in the form of color videos. In \citep{gunes2005affect,nicolaou2011continuous}, videos are collected in a lab setting with a pure-colored background and a fixed-perspective camera. They could detect and track hands and other landmarks with simple thresholding and grouping of pixels. \citet{gunes2005affect} additionally defined motion protocols, such as whether the hand is facing up, and combined them with landmark displacement as features. \citet{nicolaou2011continuous} used the positions of shoulders in the image frame, facial expression, and audio features as the input of a neural network. Our data, however, is not collected under such controlled settings, thus has variations in viewpoint, lighting condition, and scale.}

\revise{For non-video-based methods, locations of body\break markers are inferred by the MoCap system~\citep{kleinsmith2011automatic,kleinsmith2006cross,aristidou2015emotion, schindler2008recognizing}. 
The first two steps, {\it i.e.,} human detection, and pose estimation and tracking, are solved directly by the MoCap system. Geometric features, such as velocity, acceleration, and orientation of body landmarks, as well as motion protocols can then be conveniently developed and used to build predictive models~\citep{kleinsmith2011automatic,kleinsmith2006cross,aristidou2015emotion}. For a more comprehensive survey of automatic modeling of bodily expression, readers are referred to the three surveys~\citep{karg2013body,kleinsmith2013affective,corneanu2018survey}.}


\revise{Related to AMBE, human behaviour understanding (a.k.a. action recognition) has attracted a lot of attention. The emergence of large-scale annotated video datasets~\citep{soomro2012ucf101,caba2015activitynet,kay2017kinetics} and advances in deep learning~\citep{krizhevsky2012imagenet} have accelerated the development in action recognition. To our knowledge, two-stream ConvNets-based models have been leading on this task~\citep{simonyan2014two,wang2016temporal,carreira2017quo}. The approach uses two networks with an image input stream and an optical flow input stream to characterize appearance and motion, respectively. Each stream of ConvNet learns human-action-related features in an end-to-end fashion. Recently, some researchers have attempted to utilize human pose information. \citet{yan2018spatial}, for example, modeled human skeleton sequences using a spatiotemporal graph convolutional network. \citet{luvizon20182d} leveraged pose information using a multitask-learning approach. In our work, we extract LMA features based on skeletons and use them to build predictive models.}






\section{The BoLD Dataset}\label{sec:dataset}

In this section, we describe how we created the BoLD dataset and provide results of our statistical analysis of the data.
\begin{figure*}[t!]
  \centering
    \includegraphics[width=0.9\textwidth]{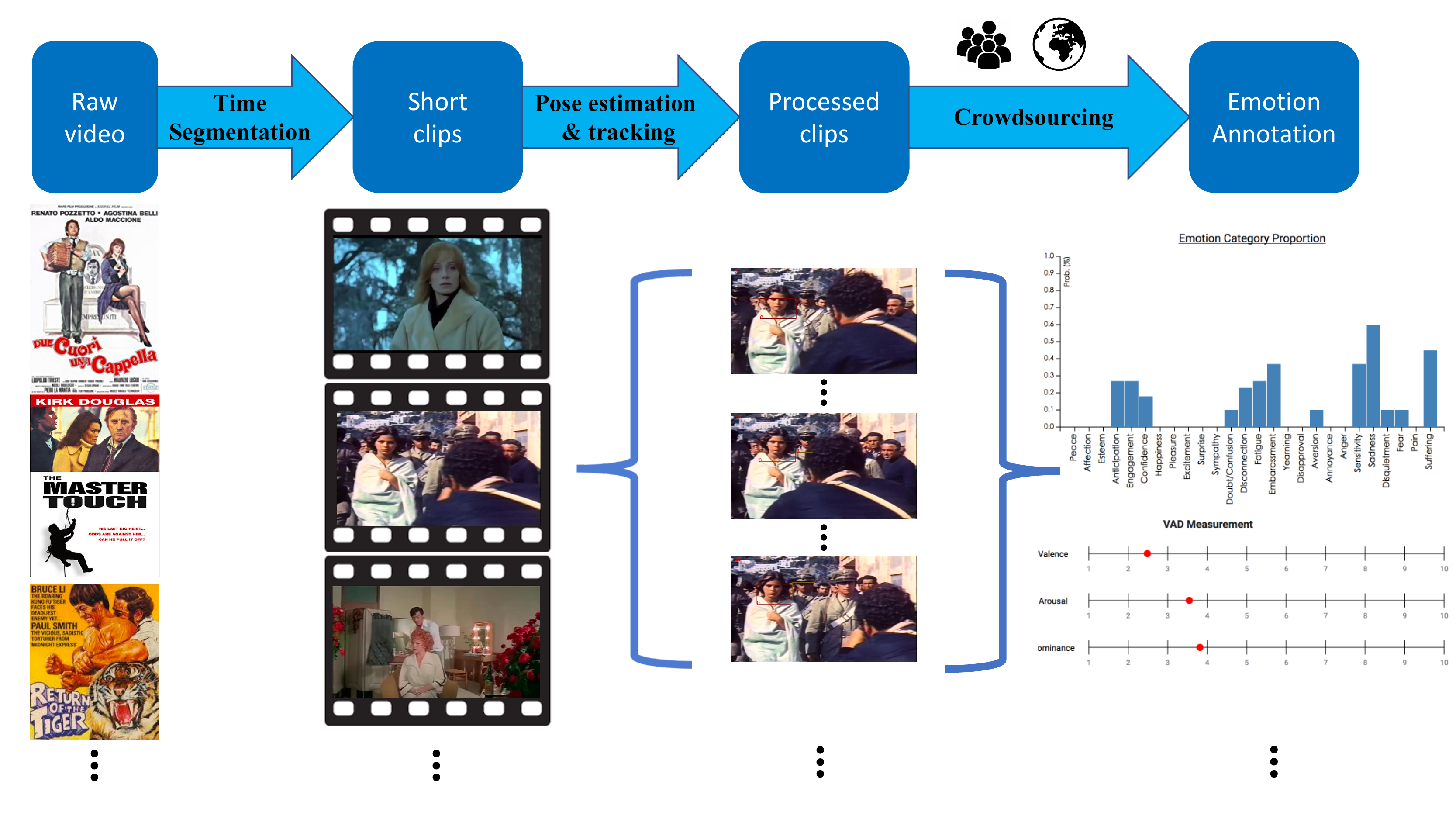}
   \caption{Overview of our data collection pipeline. The process involves crawling
   movies, segmenting them into clips, estimating the poses, and emotion annotation.}
   \label{fig:data_collection}
\end{figure*}
\subsection{\revise{Dataset Construction}}
The dataset construction process, detailed below, consists of three stages: movie selection and time segmentation, pose estimation and tracking, and emotion annotation. Fig.~\ref{fig:data_collection} illustrates our dataset construction\break pipeline. 
We chose the movies included in a
public\break dataset, the AVA dataset~\citep{gu2018ava}, which contains a list of YouTube movie IDs. To respect the
copyright of the movies, we provide the movie ID in the same way
as in the AVA dataset when the data is shared to the research
community. Any raw movies will be kept only for feature extraction and
research in the project and will not be distributed.
Given raw movies crawled from Youtube, we first partitioned each into several short scenes before using other vision-based methods to locate and track each person across different frames in the scene. To facilitate tracking, the same person in each clip was marked with a unique ID number. Finally, we obtained emotion annotations of each person in these ID-marked clips by employing independent contractors (to be called participants hereafter) from the online crowdsourcing platform, the Amazon Mechanical Turk (AMT).

\subsubsection{Movie Selection and Time Segmentation}
The Internet has vast {\it natural} human-to-human interaction videos, which serves as a rich source for our data. 
A large collection of video clips from daily lives is an ideal dataset for developing affective recognition capabilities 
because they match closely with our common real-world situations. However, a majority of those user-uploaded, in-the-wild videos suffer from poor camera perspectives and may not cover a variety of emotions. We consider it beneficial to use movies and TV
shows, {\it e.g.}, reality shows or uploaded videos in social media, that are unconstrained but offer highly interactive and emotional content. Movies and TV shows are typically of high quality in terms of filming techniques and the richness of plots. Such shows are thus more representative in reflecting characters' emotional states than some other categories of videos such as DIY instructional videos and news event videos, some of which were collected recently~\citep{abu2016youtube,thomee2016yfcc100m}. In this work, we have crawled 150 movies (220 hours in total) from YouTube by
the video IDs curated in the AVA dataset~\citep{gu2018ava}. 

Movies are typically filmed so that shots in one scene demonstrate characters' specific activities, verbal communication, and/or emotions. To make these videos manageable for further human annotation, we partition each video into short video clips using the kernel temporal segmentation (KTS) method~\citep{potapov2014category}. KTS detects shot boundary by keeping variance of visual descriptors within a temporal segment small. Shot boundary can be either a change of scene or a change of camera perspective within the same scene. To avoid confusion, we will use the term scene to indicate both cases. 

\begin{figure}[t!]
  \centering
    \includegraphics[width=0.4\textwidth]{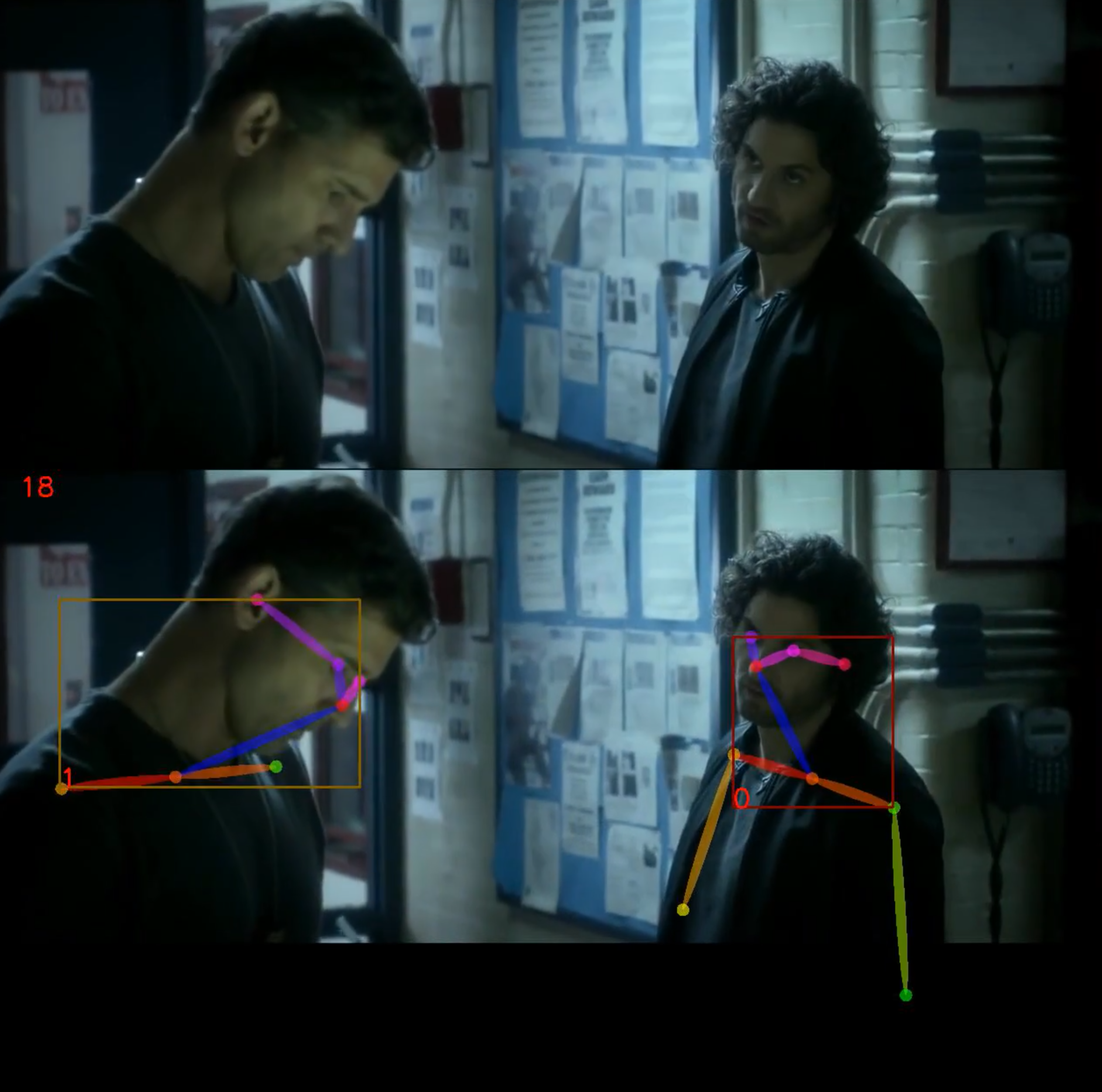}
   \caption{A frame in a video clip, with different characters numbered with an ID ({\it e.g.}, 0 and 1 \revise{at the bottom left corner of red bounding boxes}) and the body and/or facial landmarks detected (indicated with the stick figure).}
   \label{fig:demo1}
\end{figure}

\subsubsection{Pose Estimation and Tracking}
We adopted an approach to detect human body landmarks and track each character at the same
time (Fig.~\ref{fig:demo1}). Because not all short clips contain human characters, we removed those
clips without humans via pose estimation \citep{cao2017realtime}. Each clip was processed by a pose
estimator\footnote{\url{https://github.com/CMU-Perceptual-Computing-Lab/caffe_rtpose}} frame-by-frame to acquire human body landmarks. 
Different characters in one clip correspond to different samples. Each character in the clip is marked as a different sample.
To make the correspondence clear, we track each character and designate them with a unique
ID number. Specifically, tracking was conducted on the upper-body bounding box with the Kalman Filter and
Hungarian algorithm as the key component~\citep{Bewley2016_sort}\footnote{\url{https://github.com/abewley/sort}}. In our implementation, the upper-body bounding box
was acquired with the landmarks on face and shoulders. \revise{
Empirically, to ensure reliable tracking results when presenting to the annotators, we removed short trajectories that had less than $80\%$ of the total frames.}

\subsubsection{Emotion Annotation}
\begin{figure*}[ht!]
  \centering
  \begin{tabular}{llll}
  \subfloat[video data quality check]{
    \includegraphics[height=1.5in]{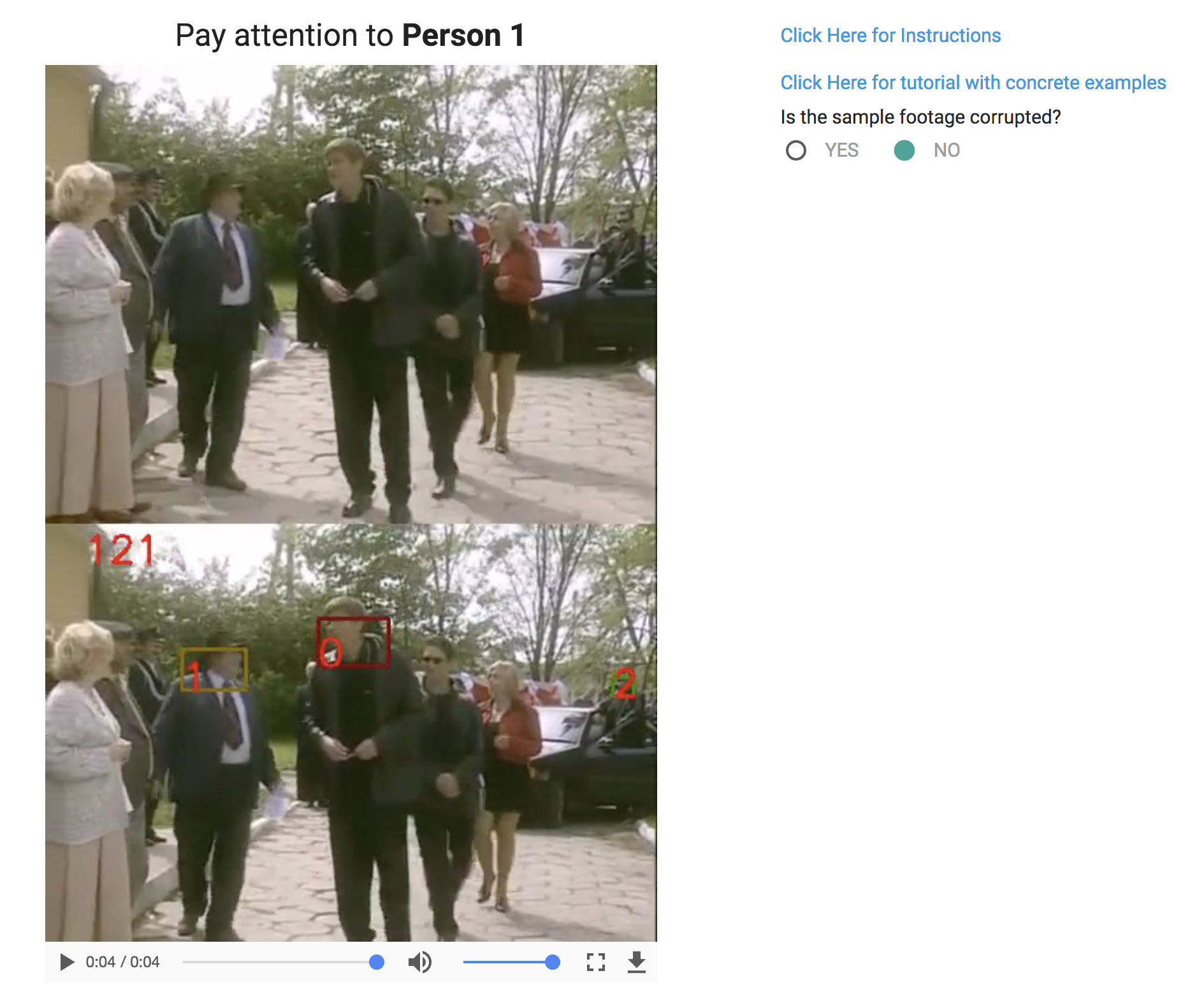}}&
  \subfloat[categorical emotion labeling]{
\includegraphics[trim={6.5in 0 0 0},clip,height=1.5in]{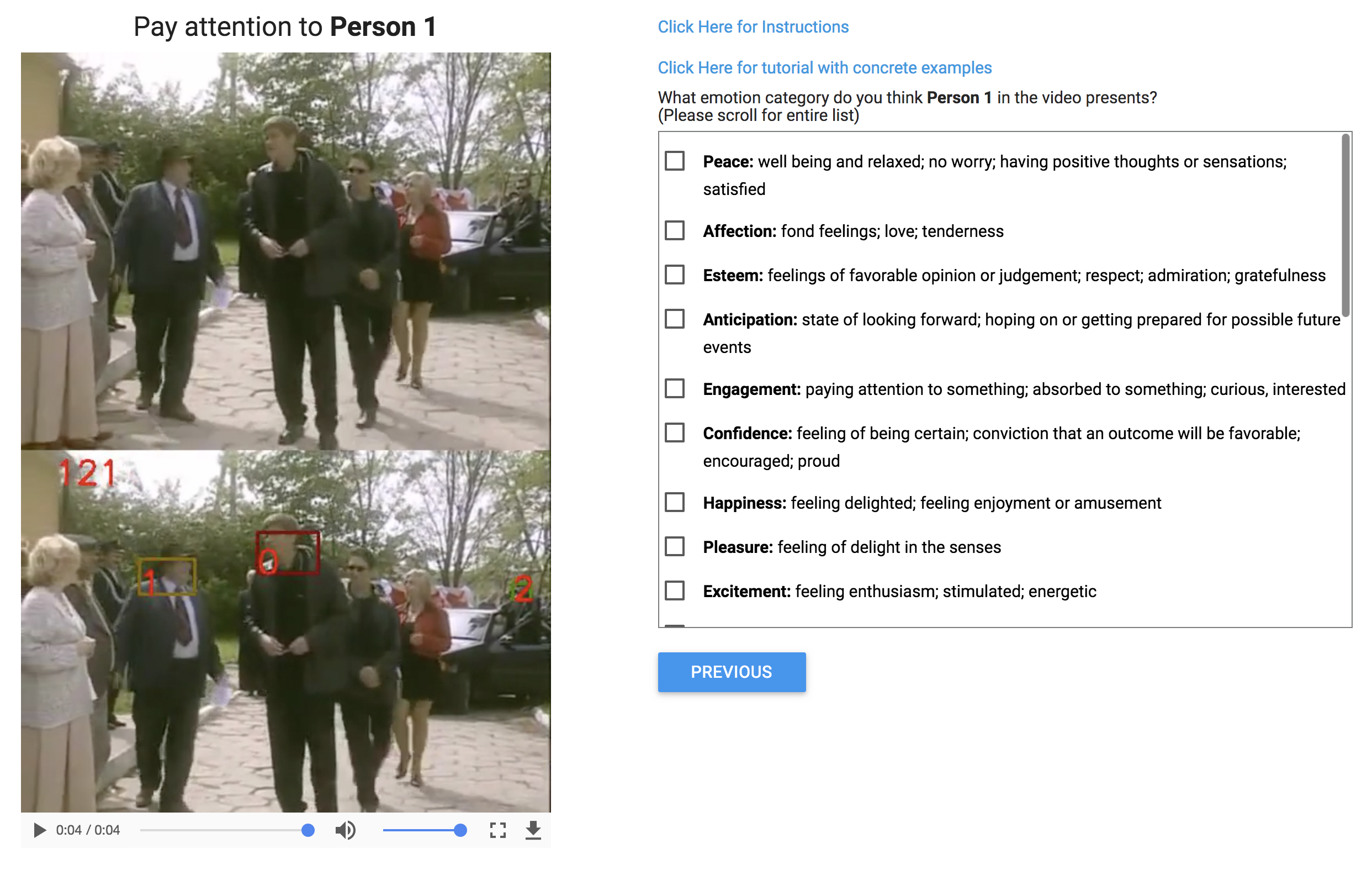}}&
  \subfloat[dimensional emotion and demographic labeling]{
    \includegraphics[trim={6.7in 0 0 0},clip,height=1.5in]{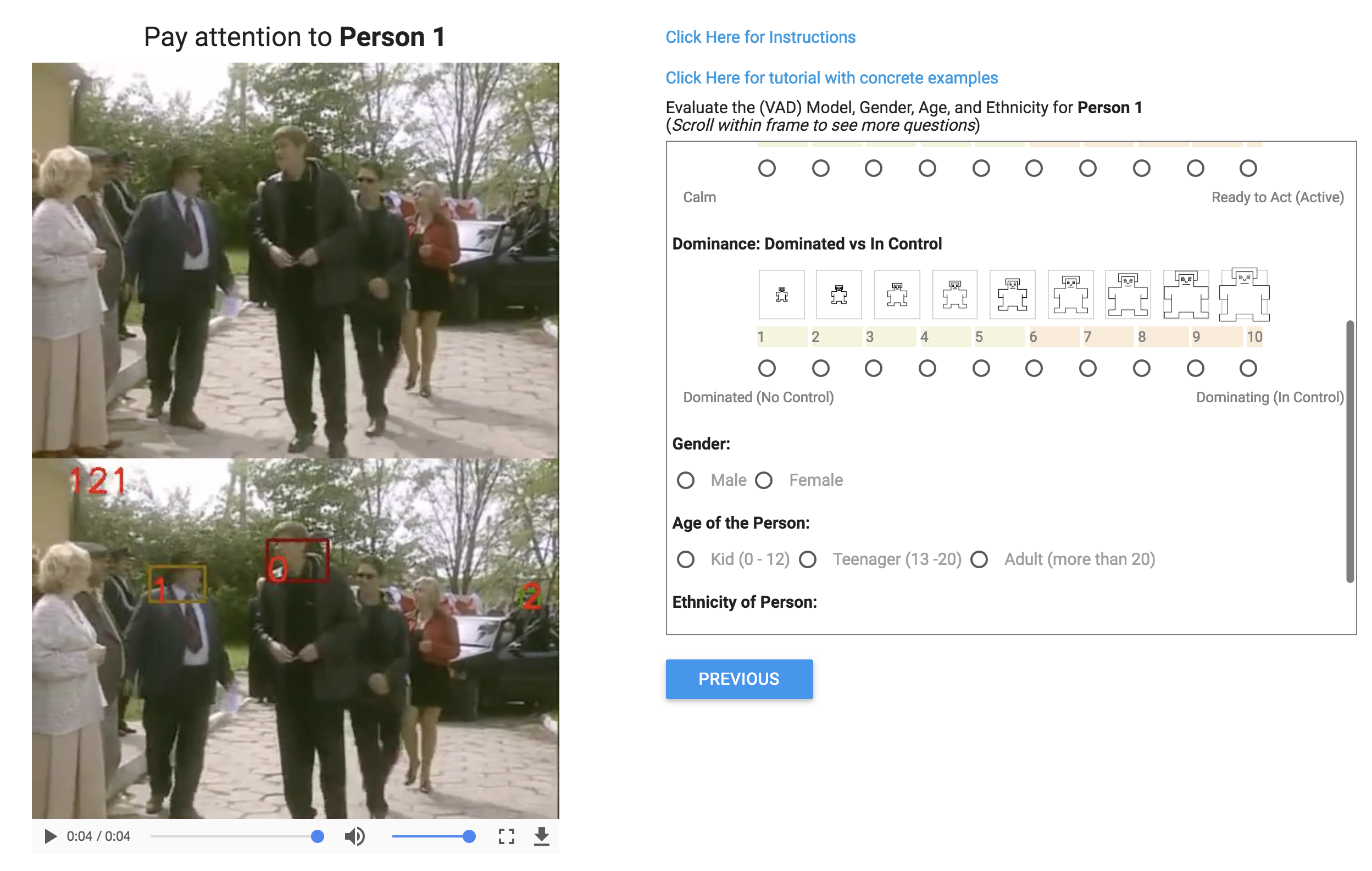}}&
  \subfloat[frame range identification]{
    \includegraphics[trim={6.5in 0 0 0},clip,height=1.5in]{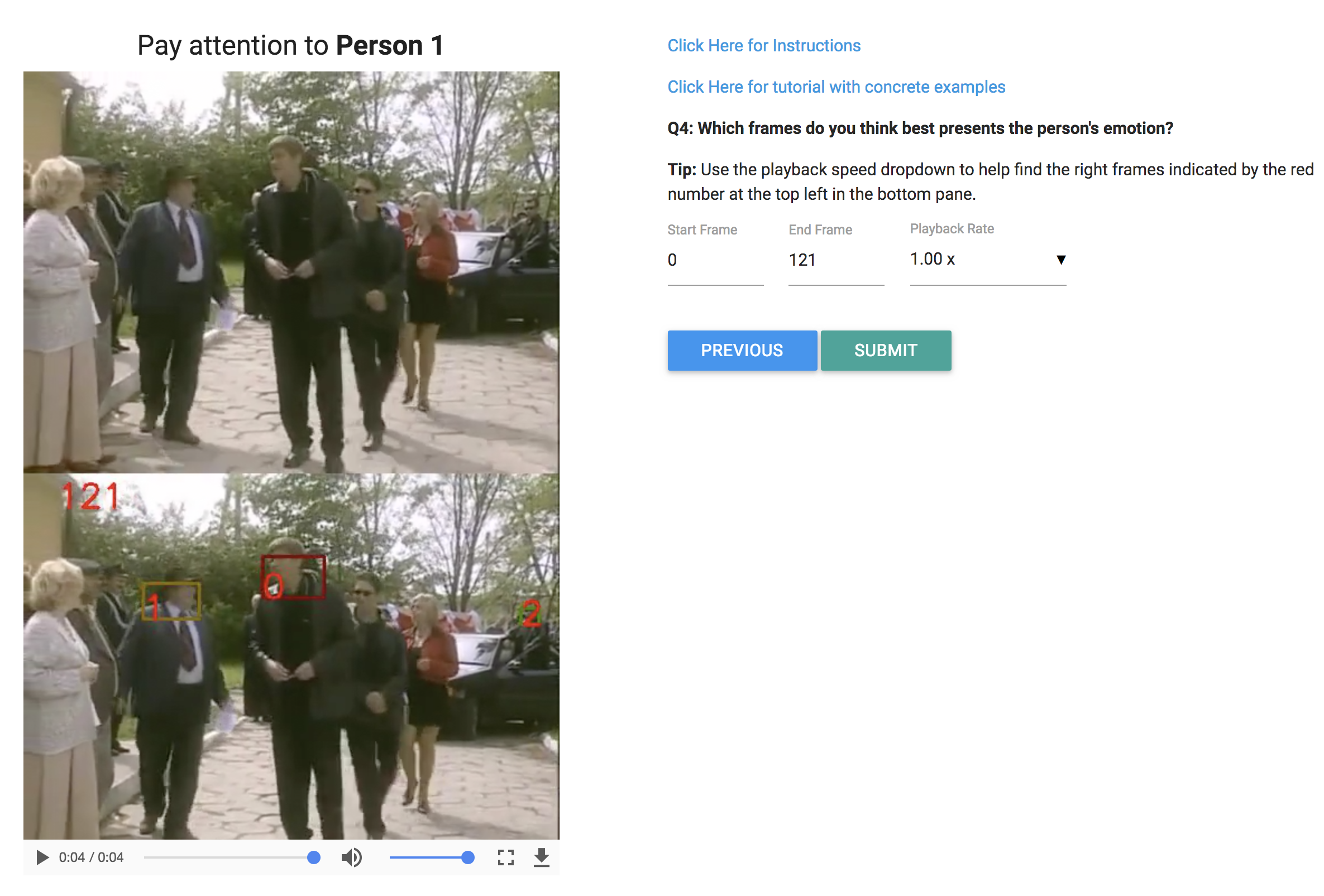}}
    \end{tabular}
   \caption{The web-based crowdsourcing data collection process. Screenshots of the four steps are shown. For each video clip, participants are directed to go through a sequence of screens with questions step-by-step.}
   \label{fig:website}
\end{figure*}

Following the above steps, we generated $122,129$ short clips from these movies.
We removed facial close-up clips using results from pose estimation. Concretely, we included a clip in our annotation list if the character in it has at least three visible landmarks out of the six upper-body landmarks, {\it i.e.}, wrists, elbows, and shoulders on both body sides (left and right). We further select those clips with 
between $100$ and $300$ frames for manual annotation by the participants. 
An identified character with landmark tracking in a single clip is called an {\it instance}. 
We have curated a total of 48,037 instances for annotation from a total of 26,164 video clips. 

We used the AMT for crowdsourcing emotion annotations of the 48,037 instances. 
For each Human Intelligence Task (HIT), a human participant completes 
emotion annotation assignments for 20 different instances. Each of which was drawn 
randomly from the instance pool. Each instance is expected to be annotated by five 
different participants.

We asked human annotators to finish four annotation tasks per instance.
Fig.~\ref{fig:website} shows screenshots of our crowdsourcing website design.
As a first step, participants must check if the instance is corrupted. An instance is considered corrupted if 
landmark tracking of the character is not consistent or the scene is not realistic in
daily life, \revise{such as science fiction scenes}. If an instance is not corrupted, participants are asked to annotate the character's emotional expressions
according to both categorical emotions and dimensional emotions ({\it i.e.}, valence, arousal, dominance (VAD) in dimensional emotion state model \citep{mehrabian1980basic}). For categorical emotions, we used the
list in \citep{kosti2017emotion}, which contains 26 categories
and is a superset of the six basic emotions \citep{ekman1993facial}. \revise{Participants are asked to annotate these categories in the way of multi-label binary classifications.} For each dimensional emotion, we used \revise{integers} that scales from $1$ to $10$. \revise{These annotation tasks are meant to reflect the truth revealed in the visual and audio data --- movie characters' emotional expressions --- and do not involve the participants’ emotional feelings.}
In addition to these tasks, participants are asked to specify a time interval \revise{({\it i.e.}, the start and end frames)}
over the clip that best represents the selected emotion(s) or has led to
their annotation. Characters' and participants' demographic information (gender, age, and ethnicity) is also \revise{annotated/}collected for complementary analysis. Gender categories are male and female. Age categories are defined as kid (aged up to 12 years), teenager (aged 13-20), and adult (aged over 20). Ethnicity categories are American Indian or Alaska Native, Asian, African American, Hispanic or Latino, Native Hawaiian or Other Pacific Islander, White, and Other. 

The participants are permitted to hear the audio of the clip, which can include a conversation in English or some other language. While the goal of this research is to study the computability of body language, we allowed the participants to use all sources of information (facial expression, body movements, sound, and limited context) in their annotation in order to obtain as high
accuracy as possible in the data collected. Additionally, the participants can play the clip back-and-forth during the entire annotation process for that clip.

To sum up, we crowdsourced the annotation of categorical and dimensional emotions, time interval of interest, and character demographic information.

\subsubsection{Annotation Quality Control}
\label{sec:qc}
Quality control has always been a necessary component for crowdsourcing to identify dishonest participants, but it is much more difficult for affect data.
Different people may not perceive affect in the same way, and their understanding may be influenced by their cultural background, current mood, gender, and personal experiences. An honest participant could also be uninformative in affect annotation, and consequently, their annotations can be poor in quality. In our study, the variance in acquiring affects usually comes from two kinds of participants, {\it i.e.}, dishonest ones, who give useless annotations for economic motivation, and exotic ones, who give inconsistent annotations compared with others. Note that exotic participants come with the nature of emotion, and annotations from exotic participants could still be useful when aggregating final ground truth or investigating cultural or gender effects of affect. In our crowdsourcing task, we want to reduce the variance caused by dishonest participants. In the meantime, we do not expect too many exotic participants because that would lead to low consensus.

Using gold standard examples is a common practice in crowdsourcing to identify uninformative participants. This approach involves curating a set of instances with known ground truth and removing those participants who answer incorrectly. For our task, however, this approach is not as feasible as in conventional crowdsourcing tasks such as image object classification. To accommodate subjectivity of affect, gold standard has to be relaxed to a large extent. Consequently, the recall of dishonest participants is lower.

To alleviate the aforementioned dilemma, we used four complementary mechanisms for quality control, including three online approaches ({\it i.e.}, analyzing while collecting the data) and an offline one ({\it i.e.,} post-collection analysis). \revise{The online approaches are participant screening, annotation sanity check, and relaxed gold standard test, while the offline one is reliability analysis.} 
\begin{itemize}
\item \textit{Participant screening.} First-time participants in our HIT must take a short empathy quotient (EQ) test\break\citep{wakabayashi2006development}. Only those who have above-average EQ are qualified. This approach aims to reduce the number of exotic participants from the beginning.

\item \textit{Annotation sanity check.} During the annotation process, the system checks consistency between categorical emotion and dimensional emotion annotations as they are entered. \revise{Specifically, we expect an ``affection'', ``esteem'', ``happiness'', or ``pleasure'' instance to have an above-midpoint valence score; a ``disapproval'', ``aversion'', ``annoyance'', ``anger'', ``sensitivity'', ``sadness'', ``disquietment'', ``fear'', ``pain'', or ``suffering'' instance to have a below-midpoint valence score; a ``peace'' instance to have a below-midpoint arousal score; and an ``excitement'' instance to have an above-midpoint arousal score.} As an example, if a participant chooses ``happiness'' and a valence rating between 1 and 5 (out of 10) for an instance, we treat the annotation as inconsistent. In each HIT, a participant fails this annotation sanity check if there are two inconsistencies among twenty instances. 

\item \textit{Relaxed gold standard test.} One control instance (relaxed gold standard) is randomly inserted in each HIT to monitor the participant's performance. We collect control instances in our trial
run within a small trusted group and choose instances with very high
consensus. We manually relax the acceptable range of each control instance to avoid false alarm. For example, for an indisputable sad emotion instance, we accept an annotation if valence is not higher than 6. An annotation that goes beyond the acceptable range is treated as failing the gold standard test. We selected nine control clips \revise{and their relaxed annotations} as the gold standard. We did not use more control clips because the average number of completed HITs per participant is much less than nine and the gold standard is rather relaxed and inefficient in terms of recall.

\item \textit{Reliability analysis.}
\revise{To further reduce the noise introduced by dishonest participants, we conduct reliability analysis over all participants.} 
We adopted the method by \citet{ye2017probabilistic} to properly handle the intrinsic subjectivity in affective data. Reliability and regularity of participants are jointly modeled. Low-reliability-score participant corresponds to dishonest participant, and low-regularity participant corresponds to exotic participant. 
This method was originally developed for improving the quality of dimensional annotations 
based on modeling the agreement multi-graph built from all participants and their annotated instances.
For each dimension of VAD, this method estimates participant $i$'s reliability score, {\it i.e.}, $r_i^v, r_i^a,r_i^d$. According to \citet{ye2017probabilistic}, the valence and arousal dimensions are empirically meaningful for ranking  participants' reliability scores.
Therefore, we ensemble the reliability score as $r_i = (2 r_i^v + r_i^a)/3$. 
We mark participant $i$ as failing in reliability analysis if $r_i$ is less than $\frac 13$ with enough effective sample size.
\end{itemize}

Based on these mechanisms, we restrain those participants deemed `dishonest.' After each HIT, participants with low performance are blocked for one hour. Low-performance participant is defined as either failing the annotation sanity check or the relaxed gold standard test. We reject the work if it shows low performance and fails in the reliability analysis. In addition to these constraints, we also permanently exclude participants with a low reliability score from participating our HITs again.

\subsubsection{Annotation Aggregation}
Whenever a single set of annotations is needed for a clip, proper aggregation is necessary to obtain a consensus annotation from multiple participants. The Dawid-Skene method \citep{dawid1979maximum}, which is typically used to combine noisy categorical observations, computes an estimated score (scaled between $0$ and $1$) for each instance. We used the method to aggregate annotations on each categorical emotion annotation and categorical demographic annotation. Particularly, we used the notation $s_i^c$ to represent the estimated score of the binary categorical variable $c$ for the instance $i$. We set a threshold of $0.5$ for these scores when binary categorical annotation is needed. For dimensional emotion, we averaged the set of annotations for a clip with their annotators' reliability score~\citep{ye2017probabilistic}. Considering a particular instance, suppose it has received $n$ annotations. The score $s_i^d$ is annotated by participant $i$ with reliability score $r_i$ for dimensional emotion $d$, where $i \in \{1, 2,\dots, n\}$ and $d \in \{\text{V, A, D}\}$ in the VAD model. The final annotation is then aggregated as
\begin{equation}
s^d = \frac{\Sigma_{i=1}^{n} {r_i s_i^d} }{10\Sigma_{i=1}^n r_i}\;.
\end{equation}
In the meantime, instance confidence according to the method by~\citet{ye2017probabilistic} is defined as
\begin{equation}
c = 1 - \prod_{i=1}^{n} {(1-r_i)} \;.
\end{equation}
Note that we divided the final VAD score by $10$ so that the data ranges between $0$ and $1$. Our final dataset to be used for further analysis retained only those instances with confidence higher than $0.95$.

Our website sets a default value for the start frame (0) and the end frame (total frame number of the clip) for each instance. Among the data collected, there were about a half annotations that have non-default values, which means a portion of the annotators either considered the whole clip as the basis for their annotations or did not finish the task. For each clip, we selected the time-interval entered by the participant with the highest reliability score as the final annotation for the clip. 



\subsection{Dataset Statistics}
\begin{figure*}[ph!]
  \centering
  \begin{tabular}{m{2.15in} m{2.15in} m{2.15in}}
  \subfloat[peace]{
    \includegraphics[height=1.4in]{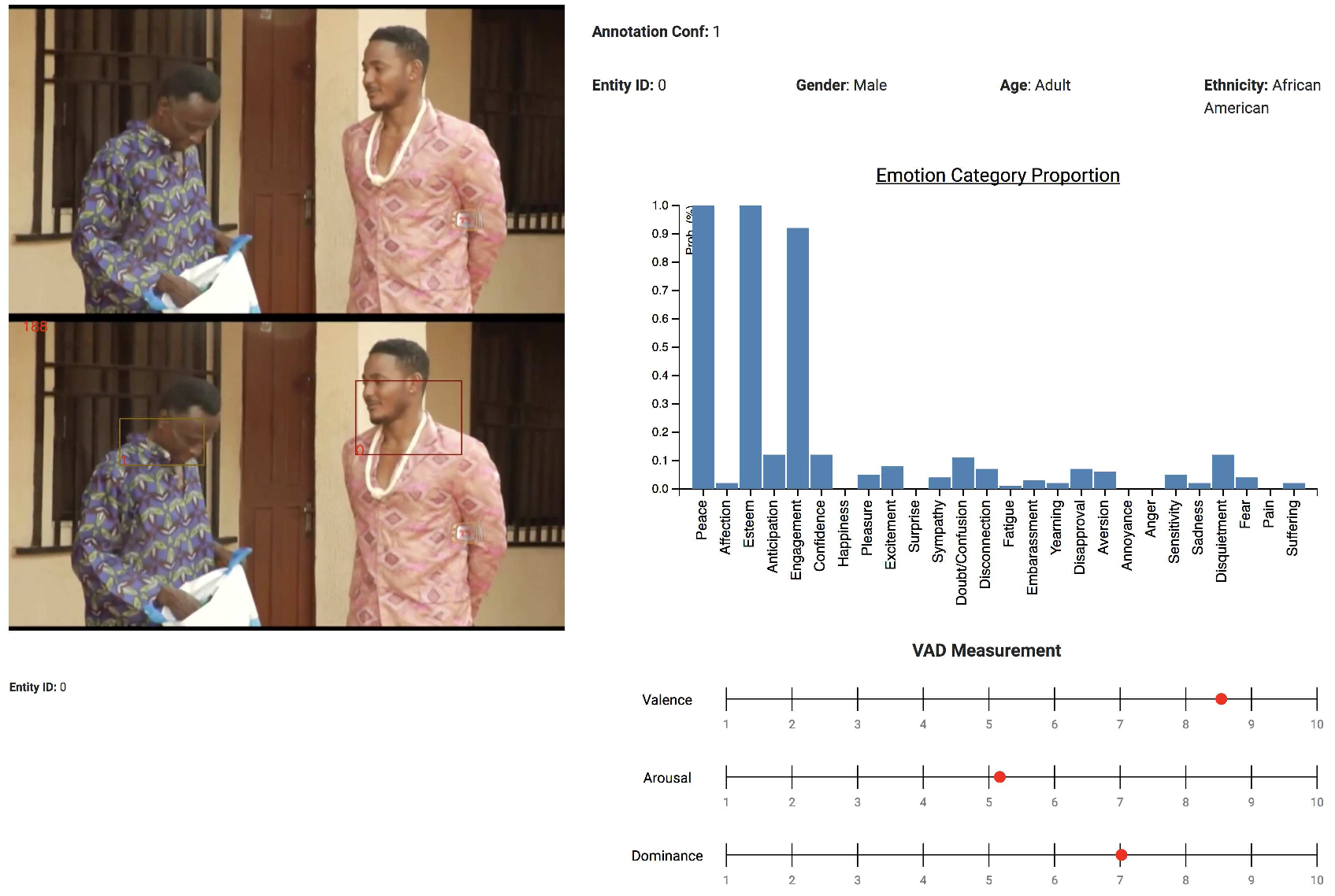}}&
  \subfloat[affection]{
\includegraphics[height=1.4in]{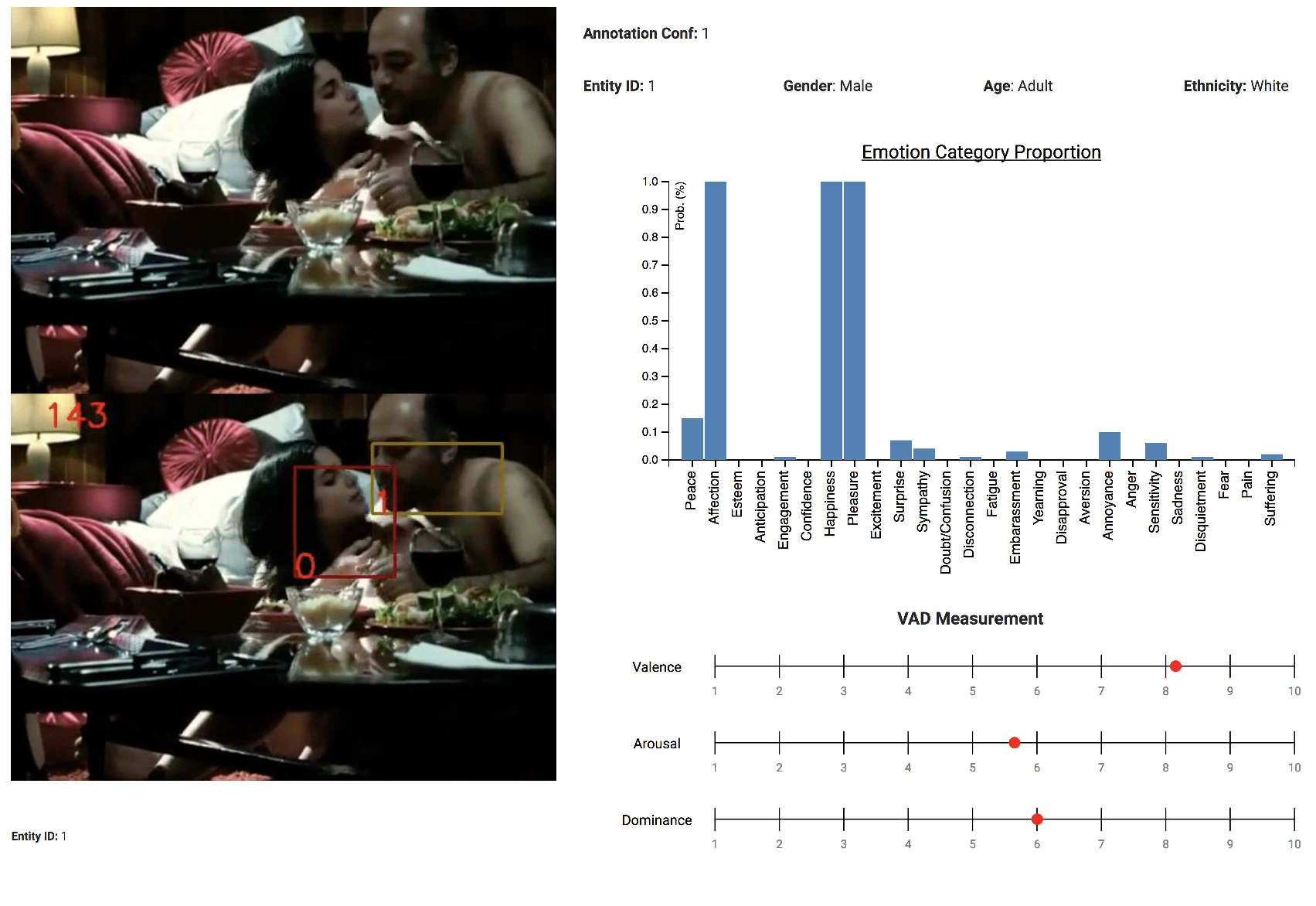}} &
  \subfloat[esteem]{
    \includegraphics[height=1.4in]{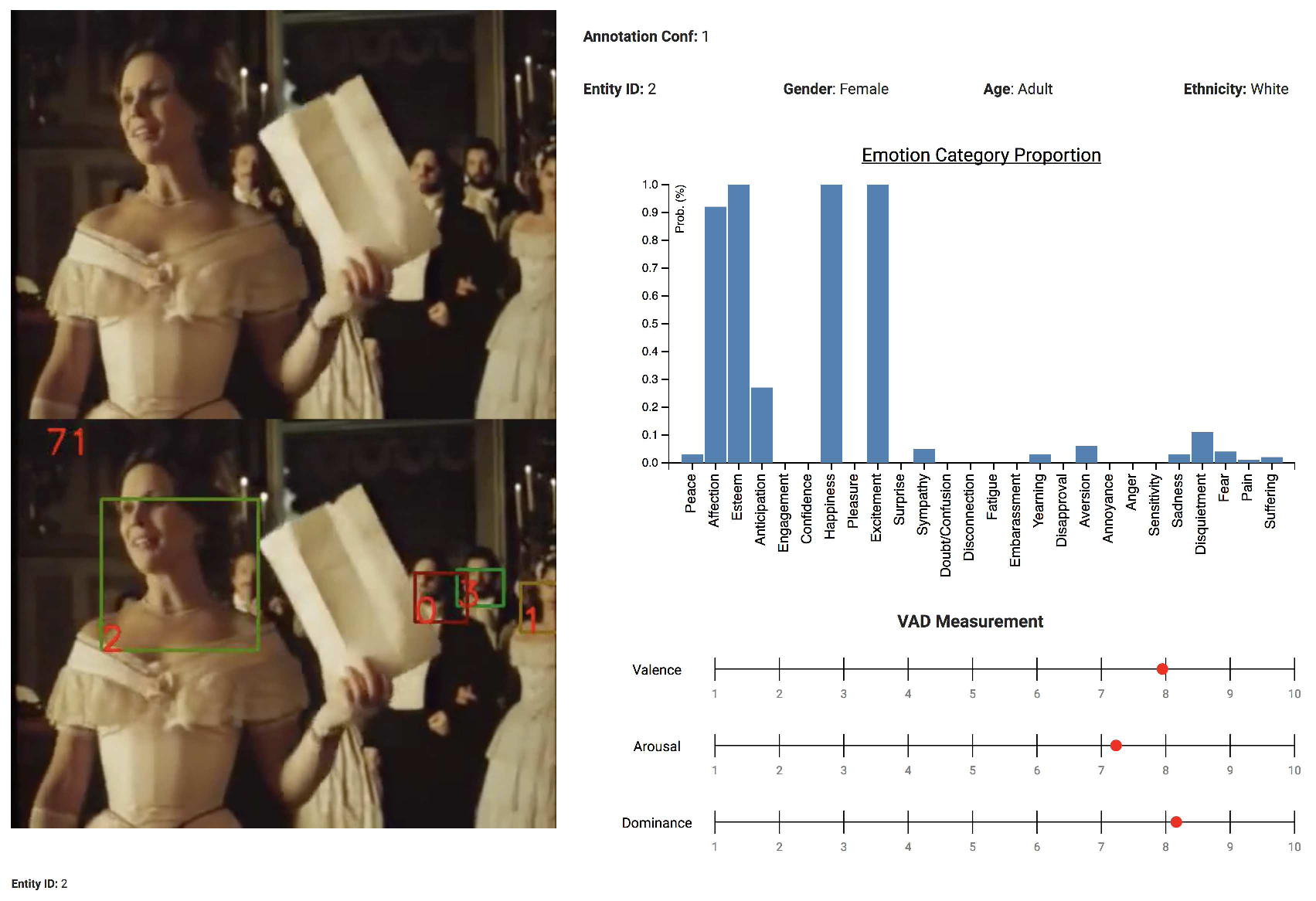}}\\
  \subfloat[anticipation]{
    \includegraphics[height=1.4in]{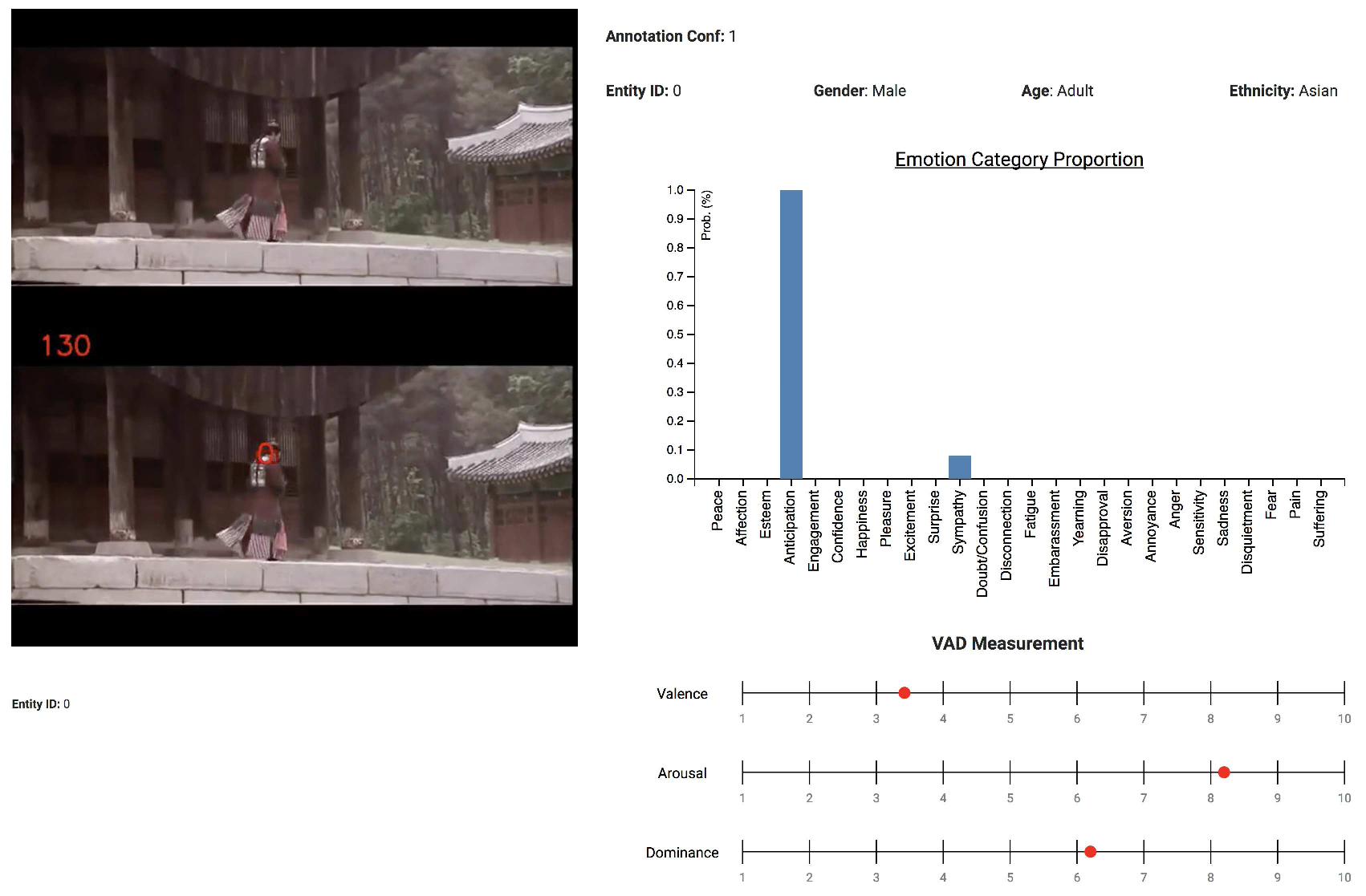}}&
     \subfloat[engagement]{
    \includegraphics[height=1.4in]{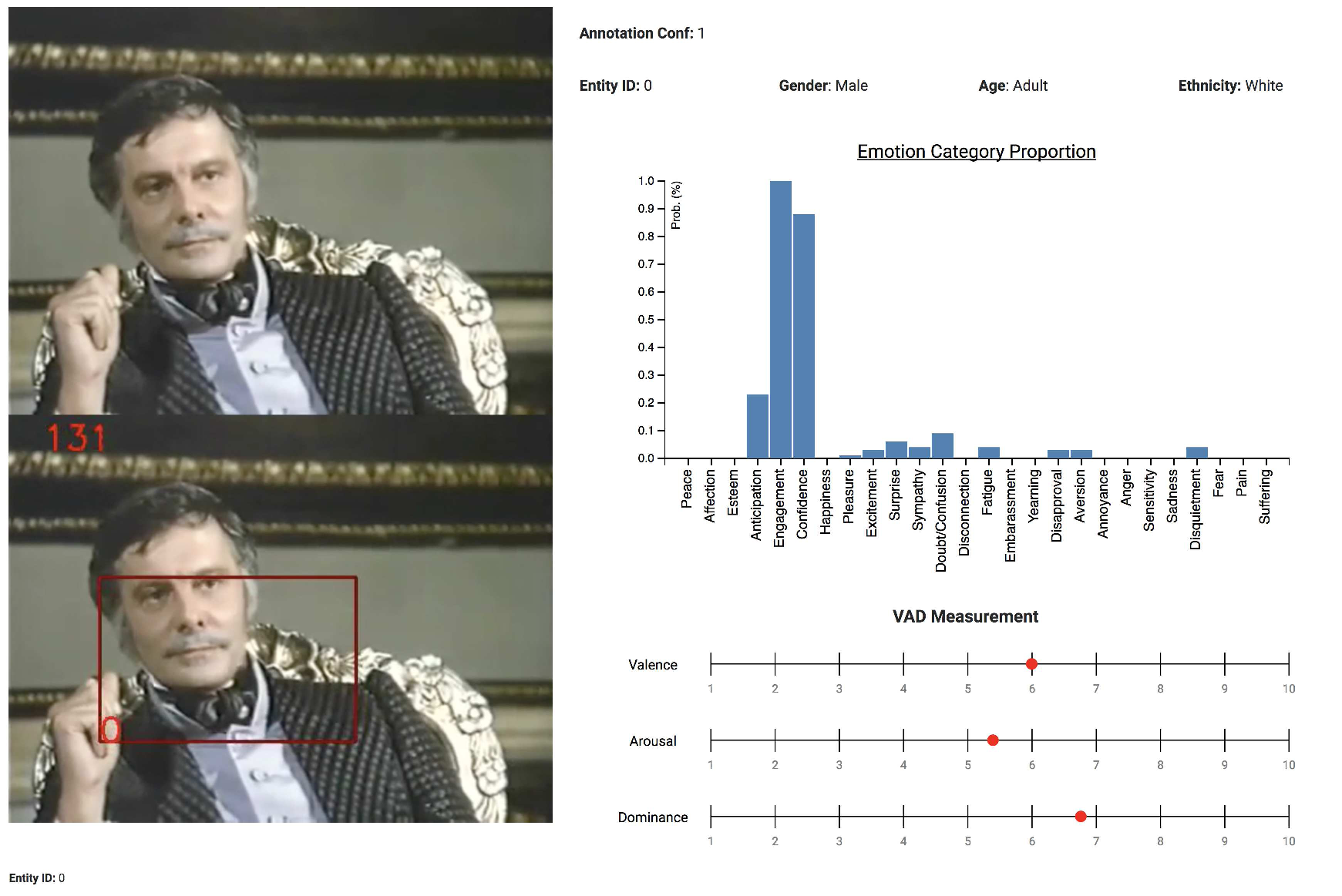}}&
  \subfloat[confidence]{
\includegraphics[height=1.4in]{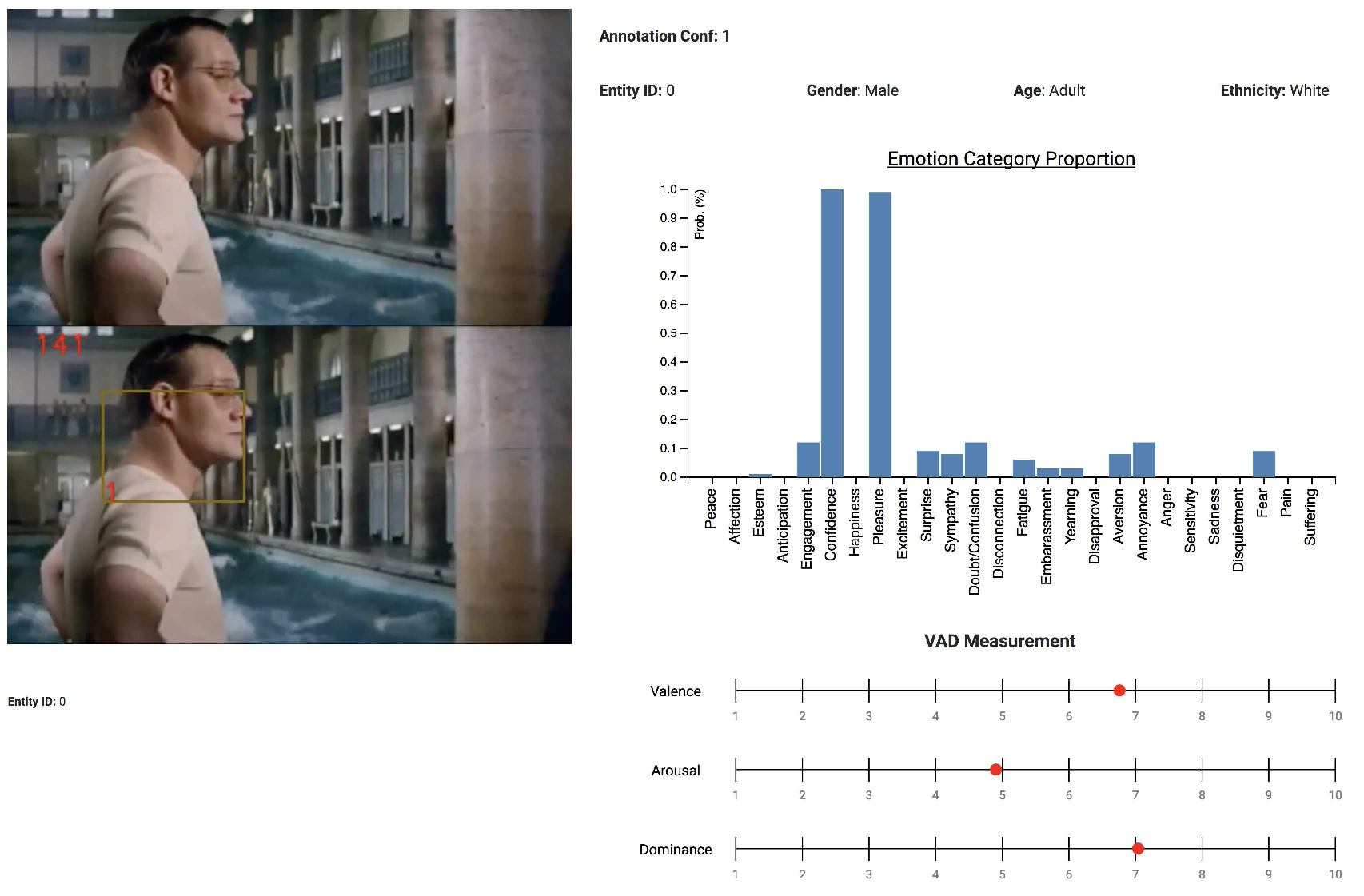}} \\
  \subfloat[happiness]{
    \includegraphics[height=1.4in]{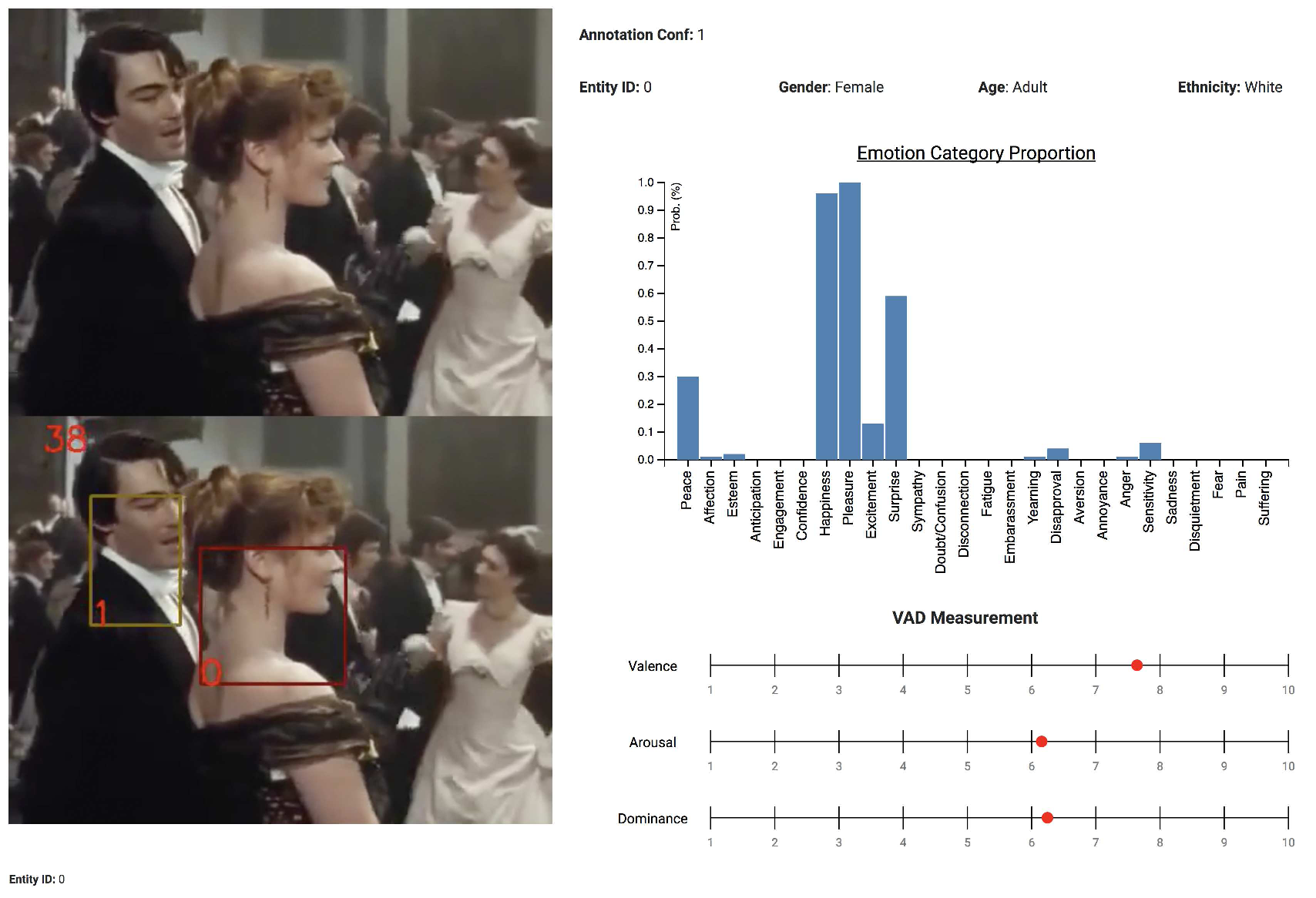}}&
  \subfloat[pleasure]{ 
    \includegraphics[height=1.4in]{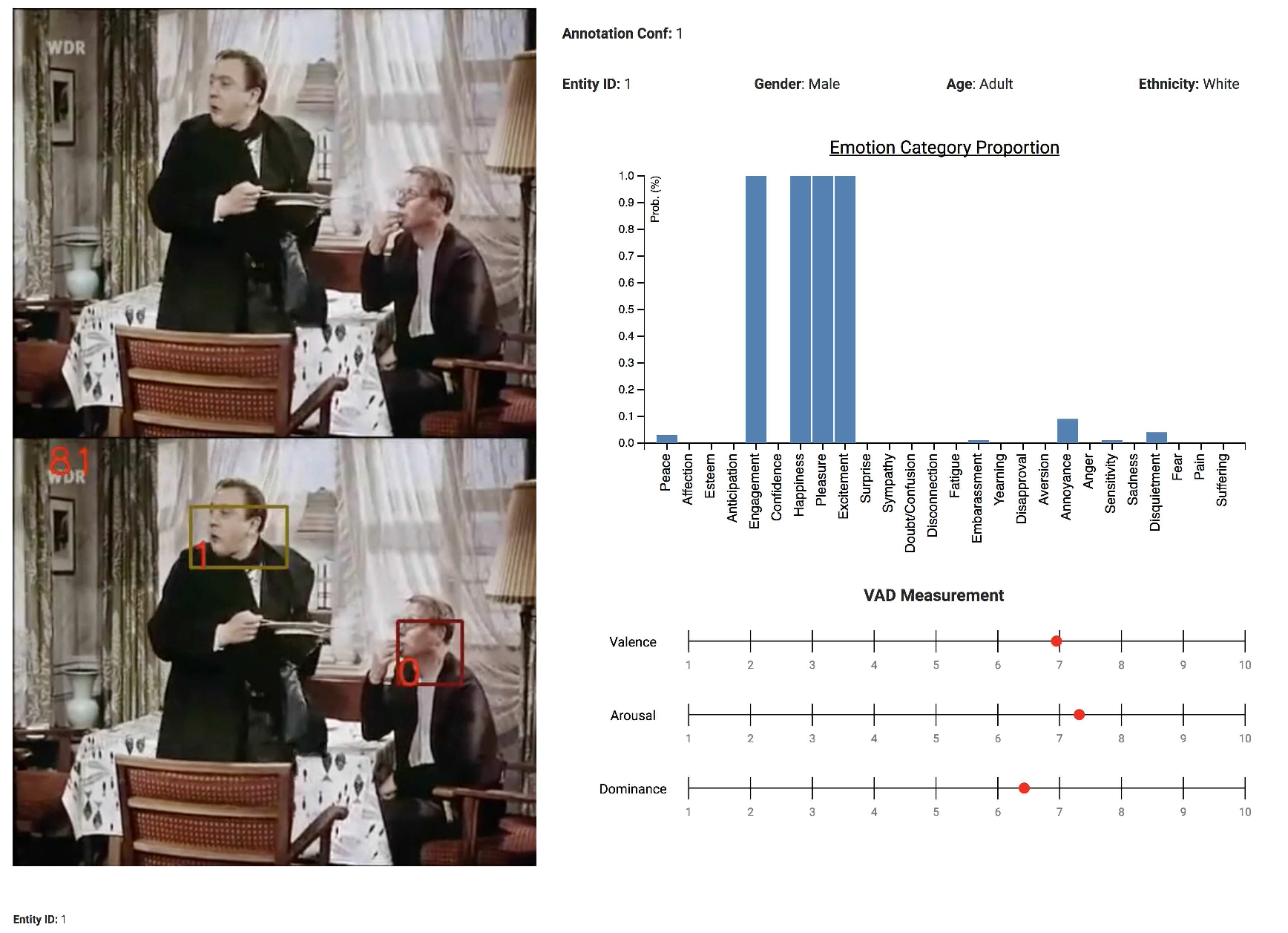}} &
     \subfloat[excitement]{
    \includegraphics[height=1.4in]{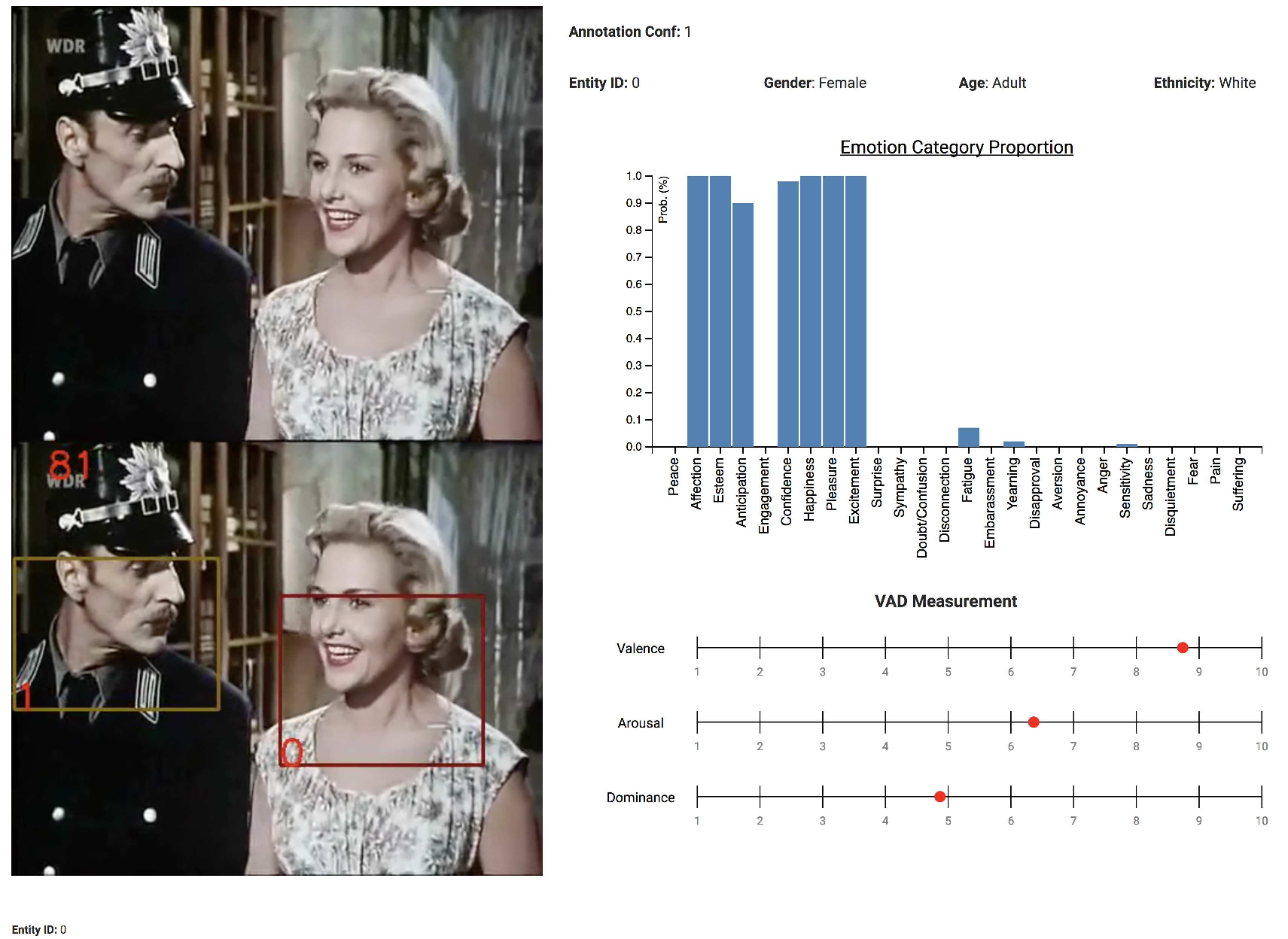}}\\
  \subfloat[surprise]{
\includegraphics[height=1.4in]{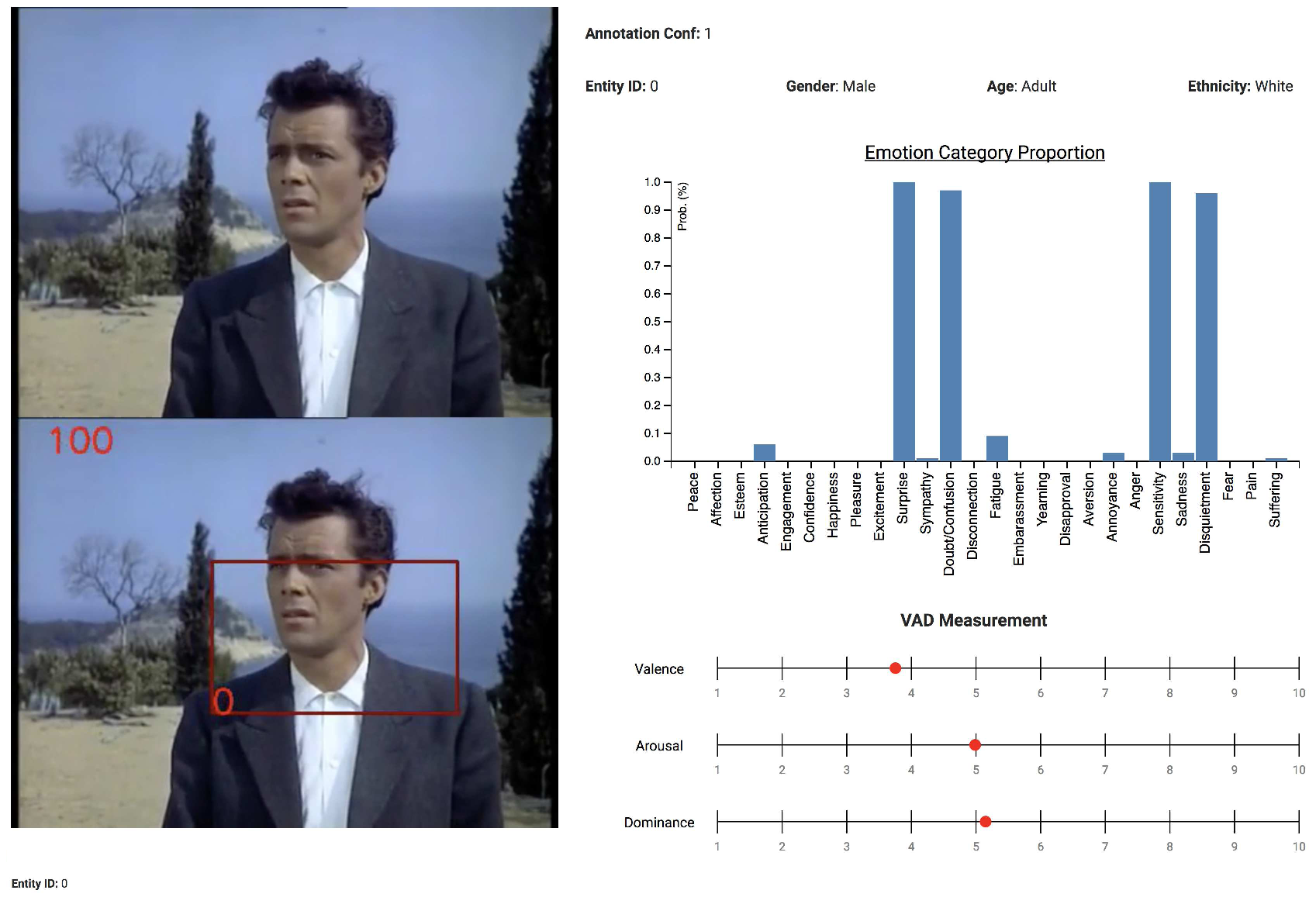}} &
  \subfloat[sympathy]{
    \includegraphics[height=1.4in]{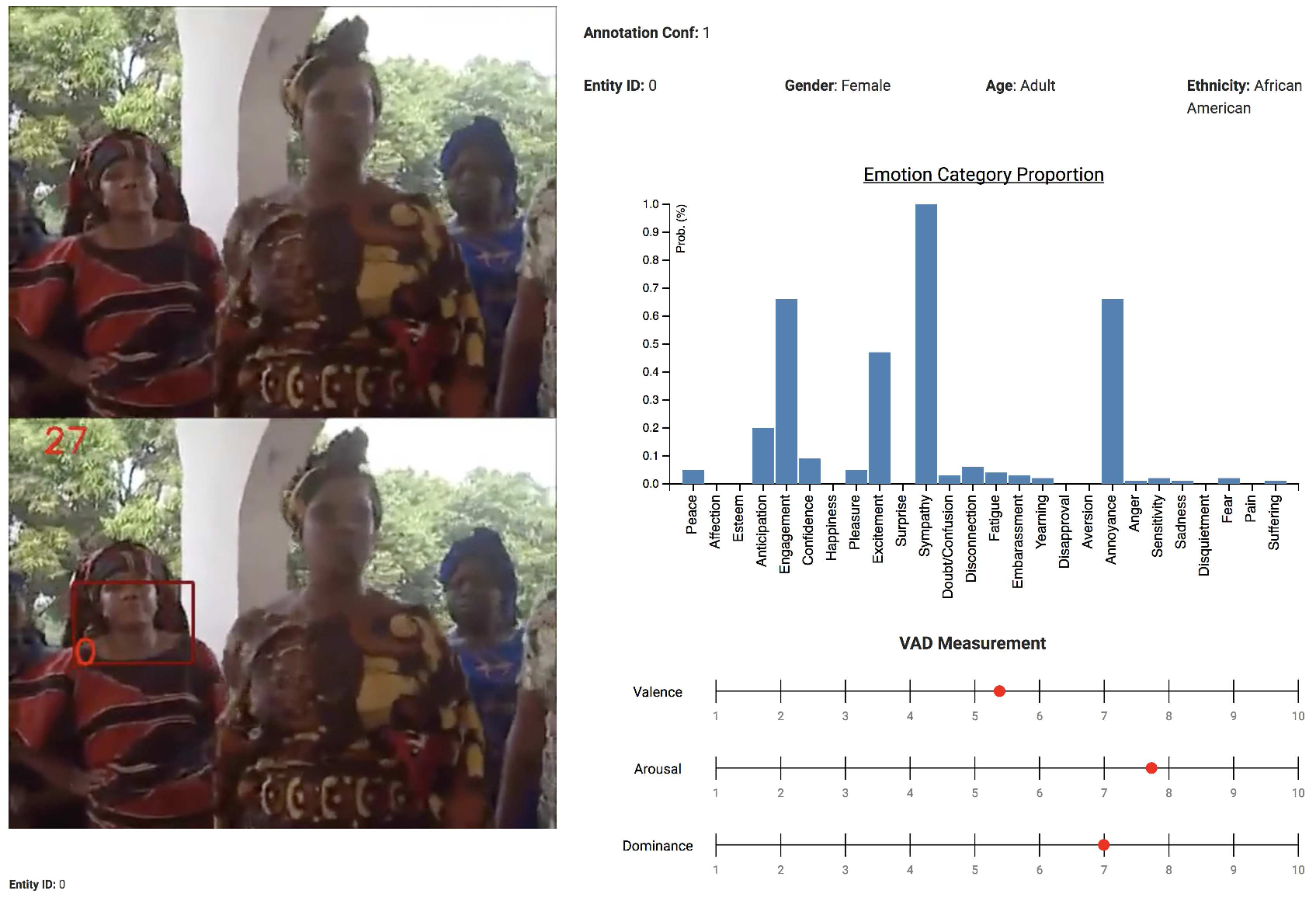}}&
  \subfloat[doubt/confusion]{
    \includegraphics[height=1.4in]{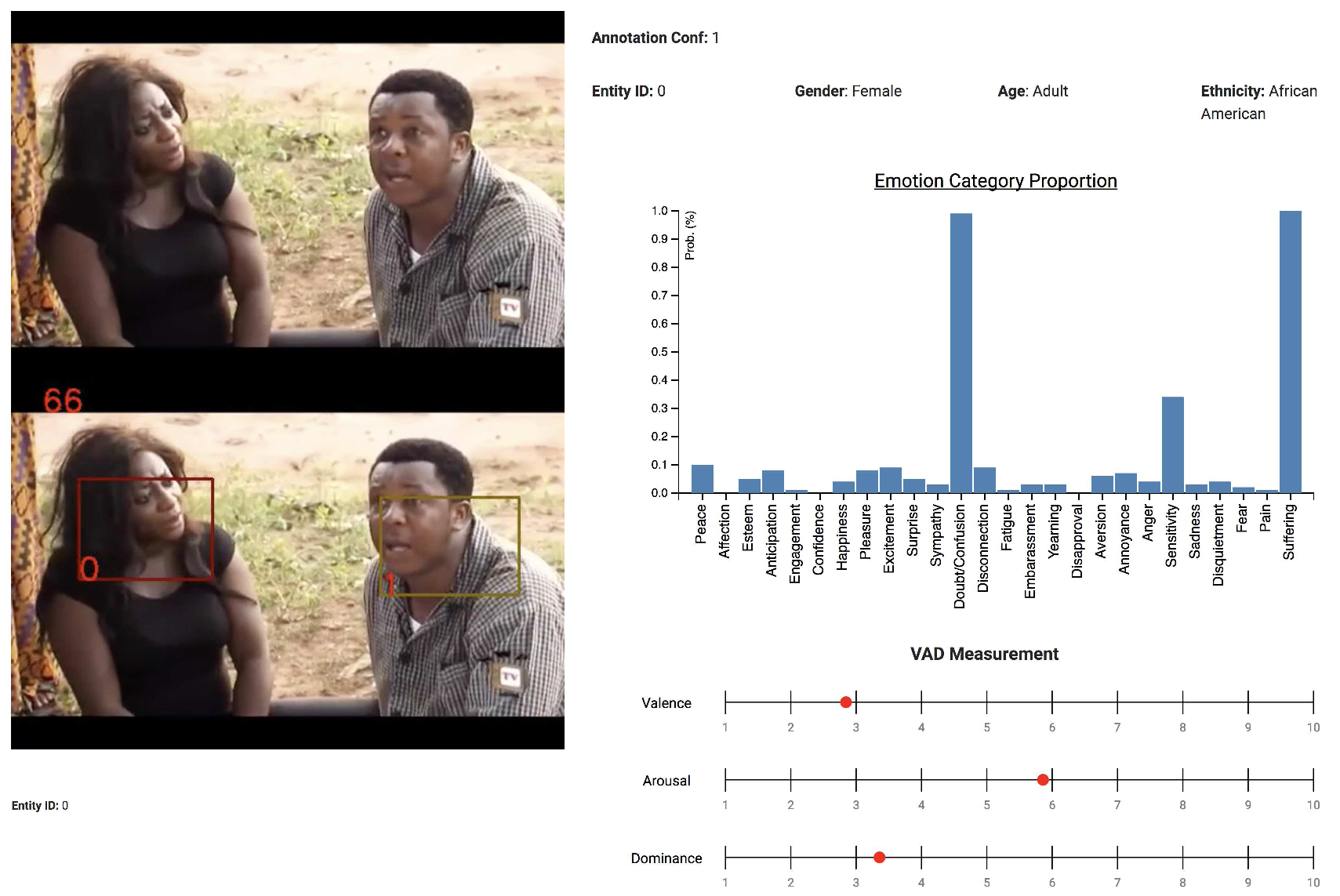}} \\
     \subfloat[disconnection]{
    \includegraphics[height=1.4in]{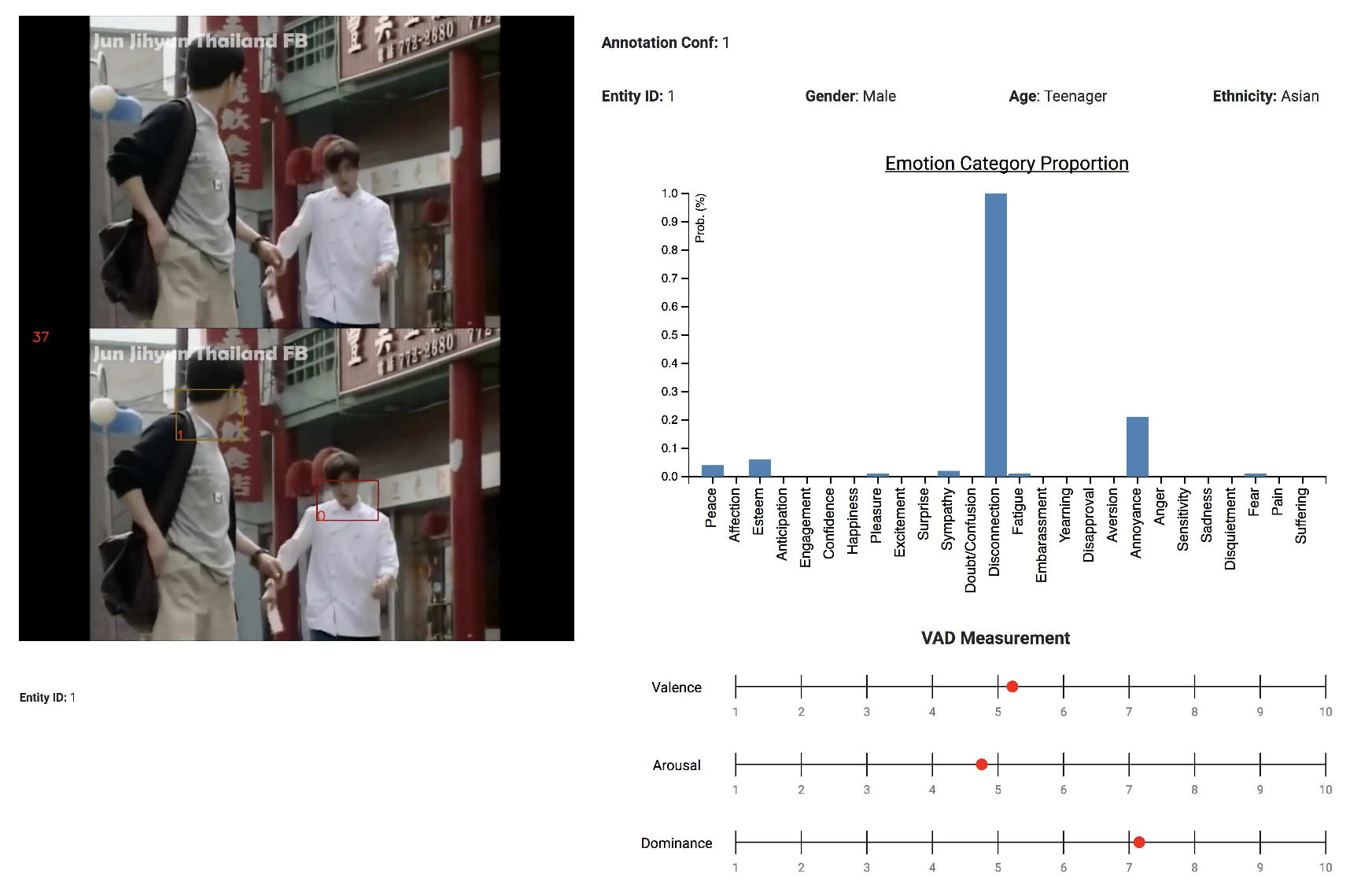}}&
  \subfloat[fatigue]{
\includegraphics[height=1.4in]{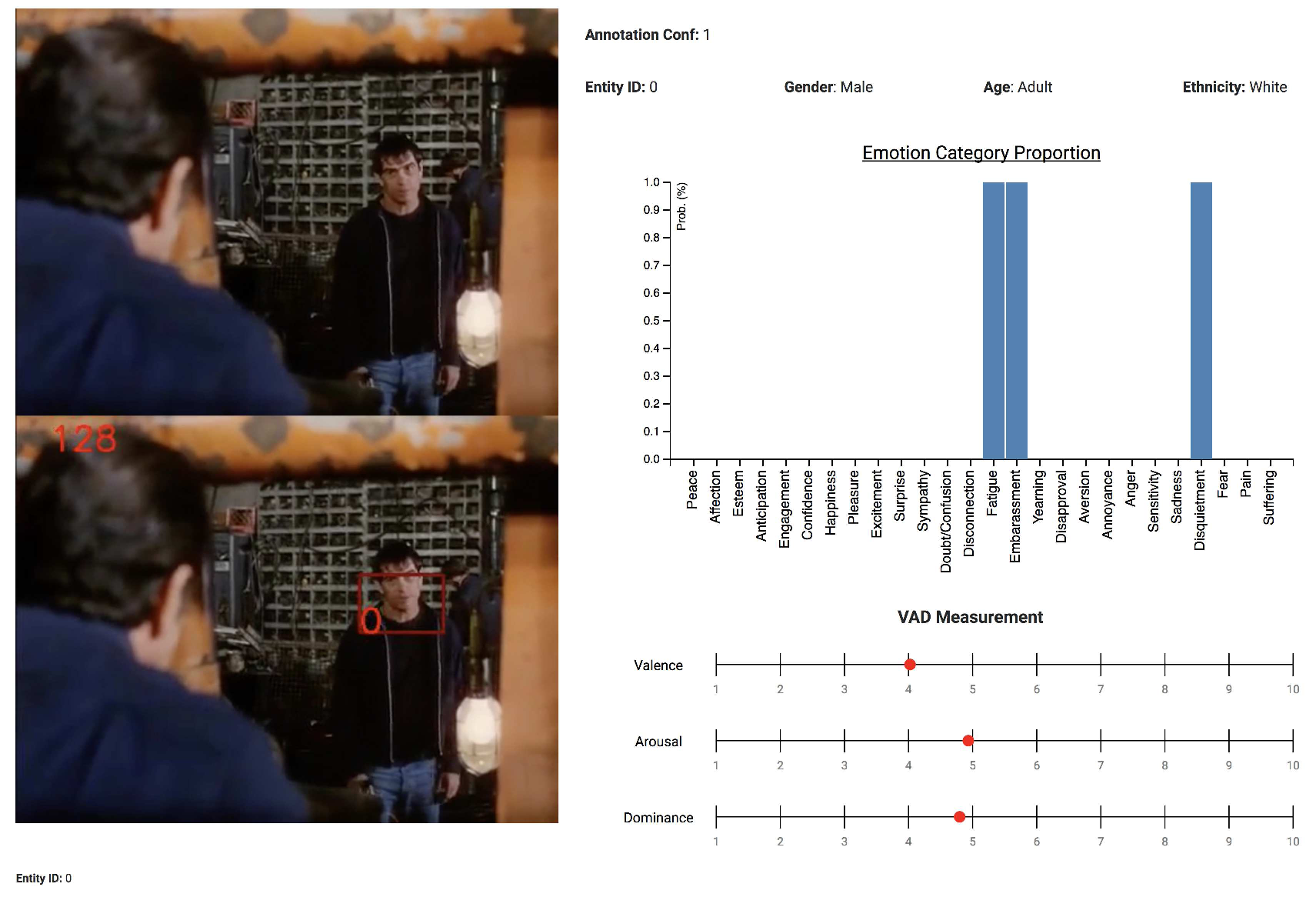}} &
  \subfloat[embarrassment]{
    \includegraphics[height=1.4in]{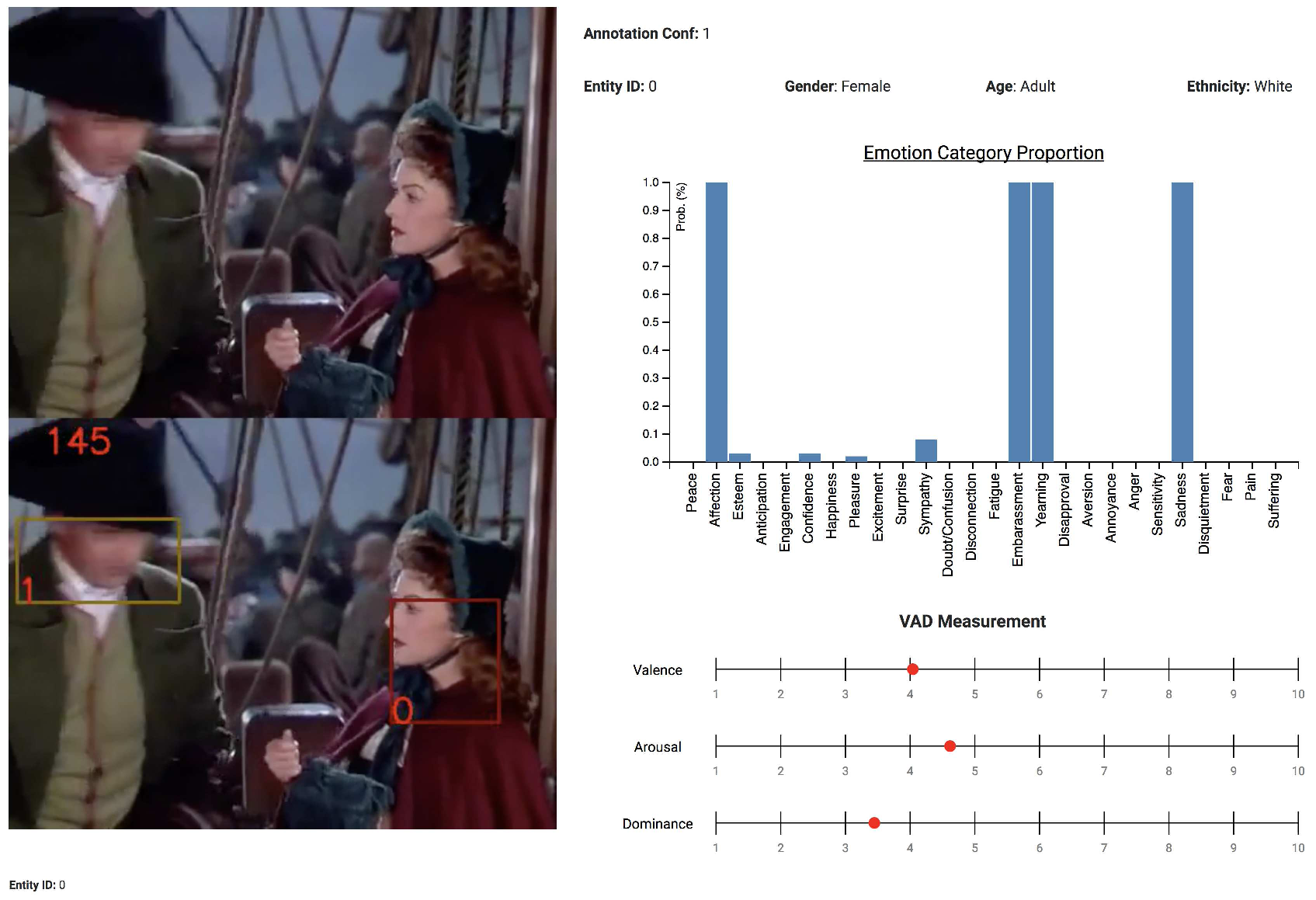}}\\
    \end{tabular}
   \caption{Examples of high-confidence instances in BoLD for the 26 categorical emotions and two instances that were used for quality control. \revise{For each subfigure, the left side is a frame from the video, along with another copy that has the character entity IDs marked in a bounding box. The right side shows the corresponding aggregated annotation, annotation confidence $c$, demographics of the character, and aggregated categorical and dimensional emotion.} To be continued on the next page. }
   \label{fig:data_examples}
\end{figure*}

\begin{figure*}[ph!]\ContinuedFloat
 
  \begin{tabular}{m{2.15in} m{2.15in} m{2.15in}}
  \subfloat[yearning]{
    \includegraphics[height=1.4in]{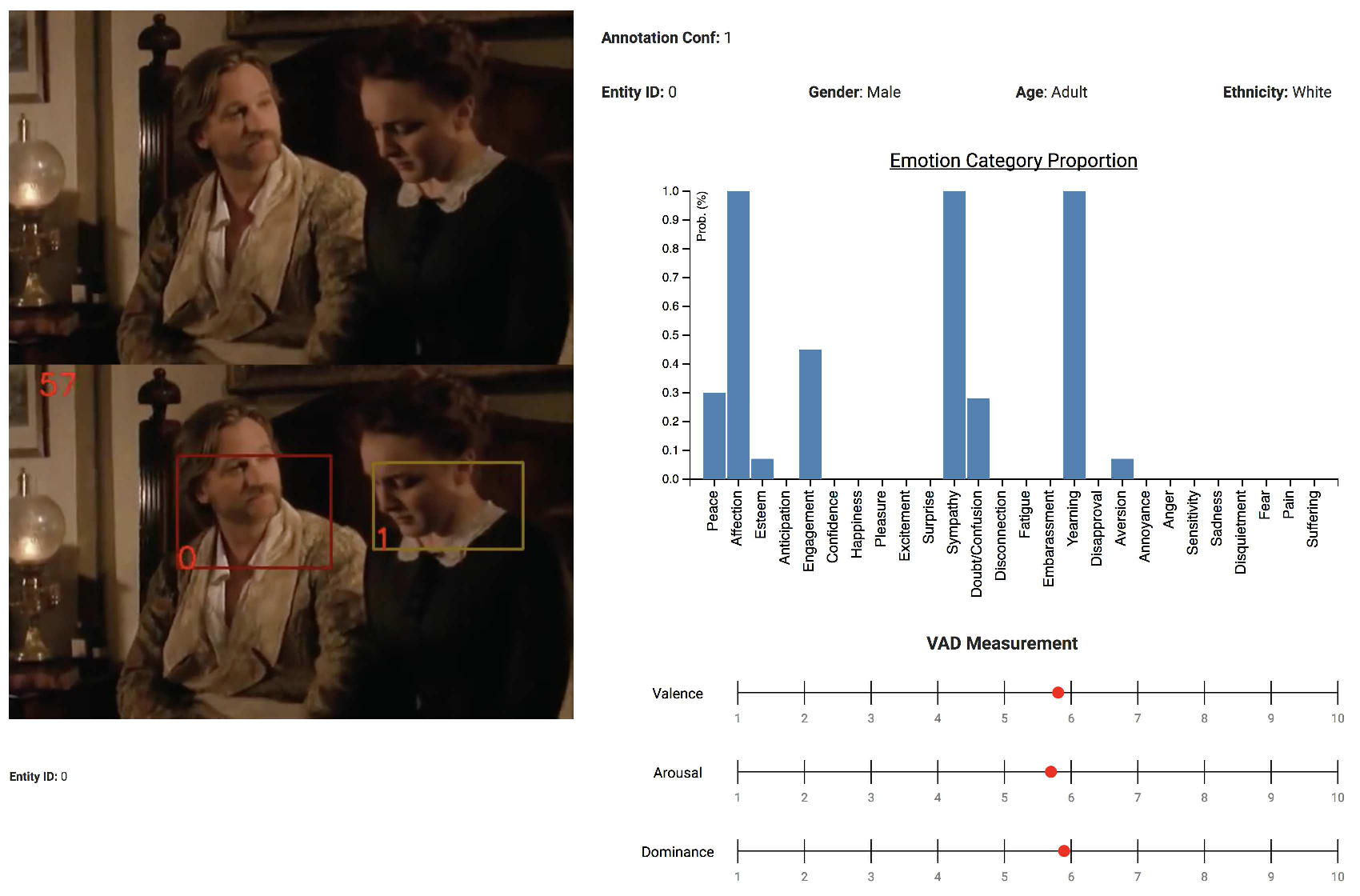}}&
     \subfloat[disapproval]{
    \includegraphics[height=1.4in]{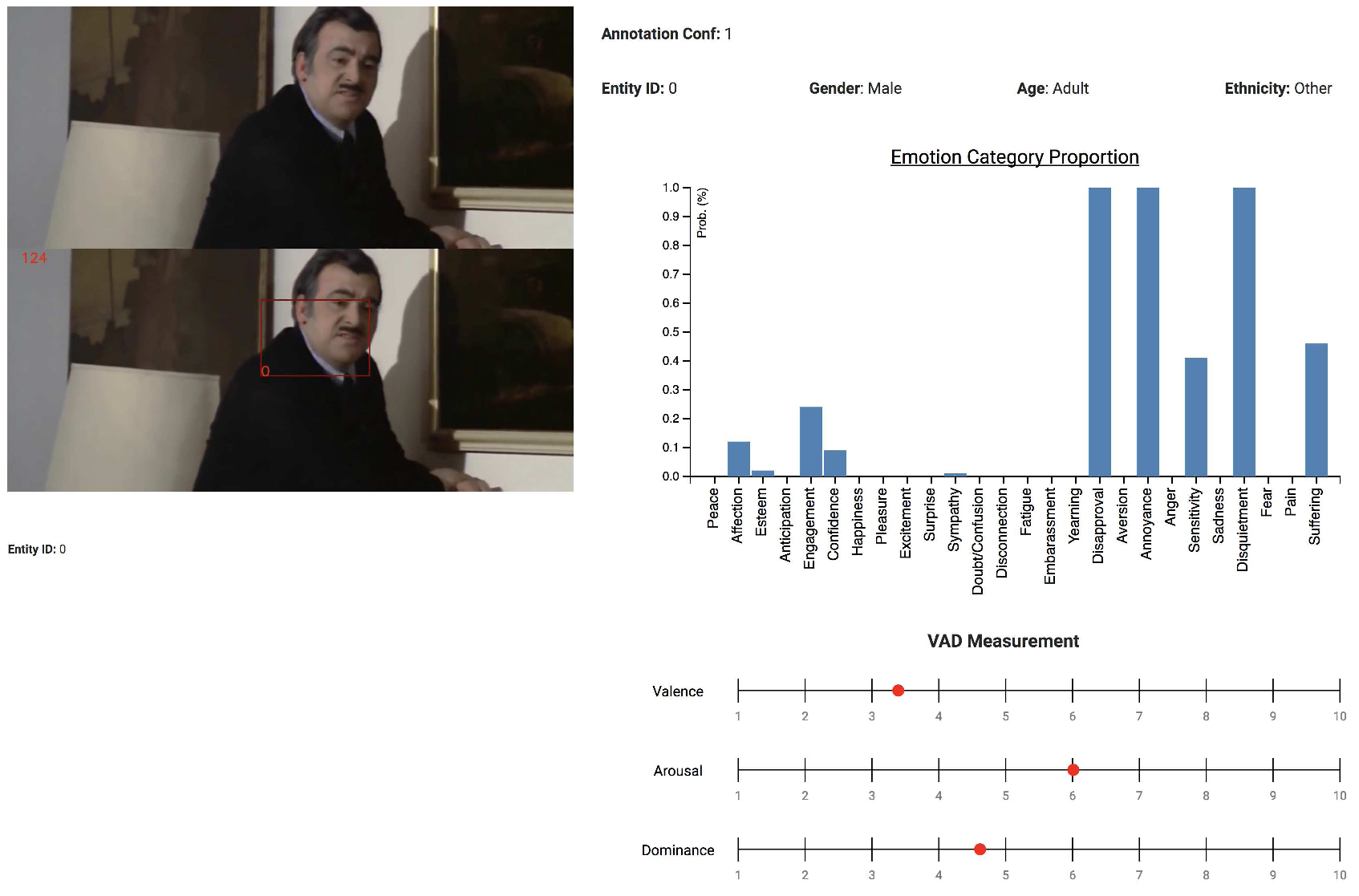}}&
  \subfloat[aversion]{
\includegraphics[height=1.4in]{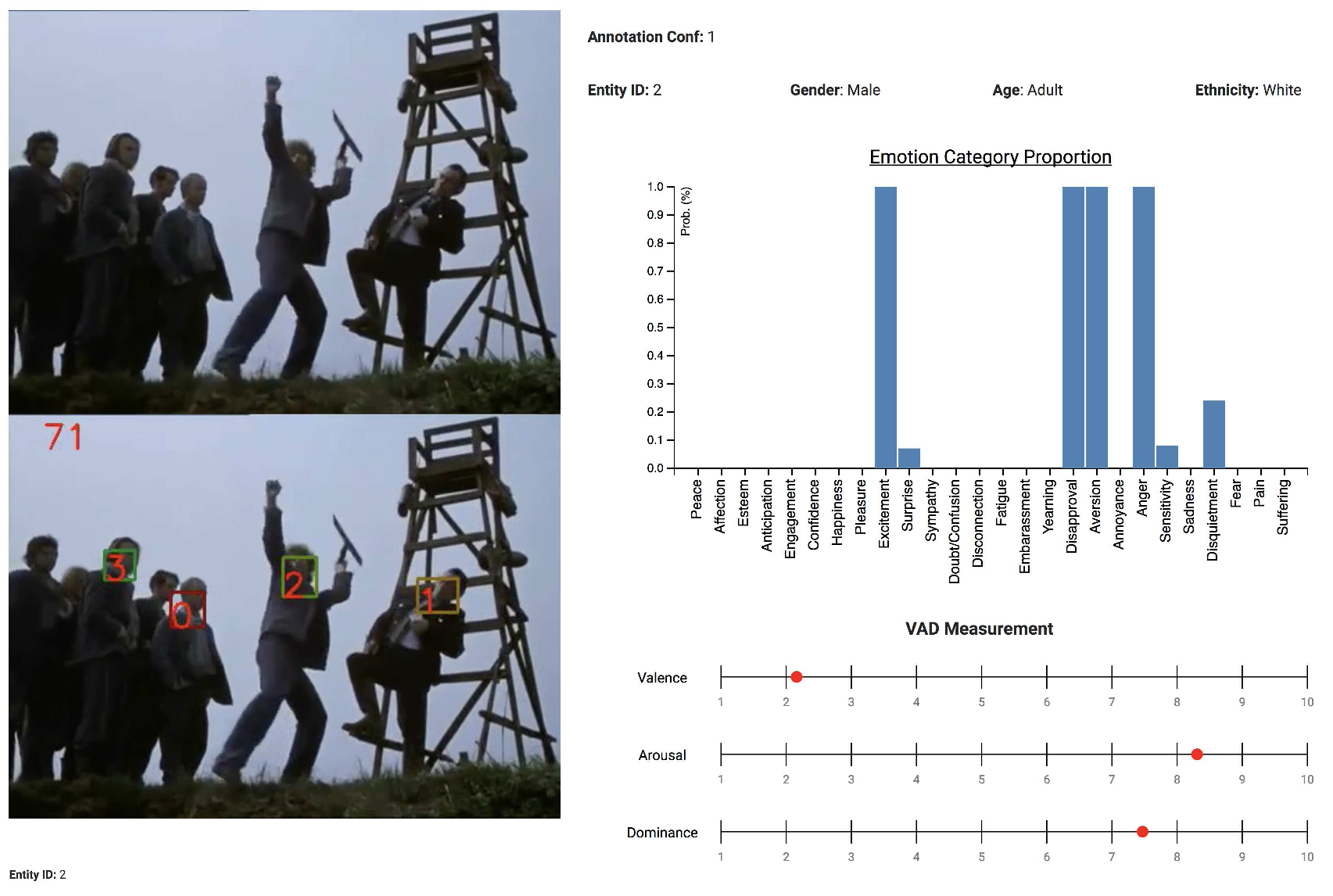}} \\
  \subfloat[annoyance]{
    \includegraphics[height=1.4in]{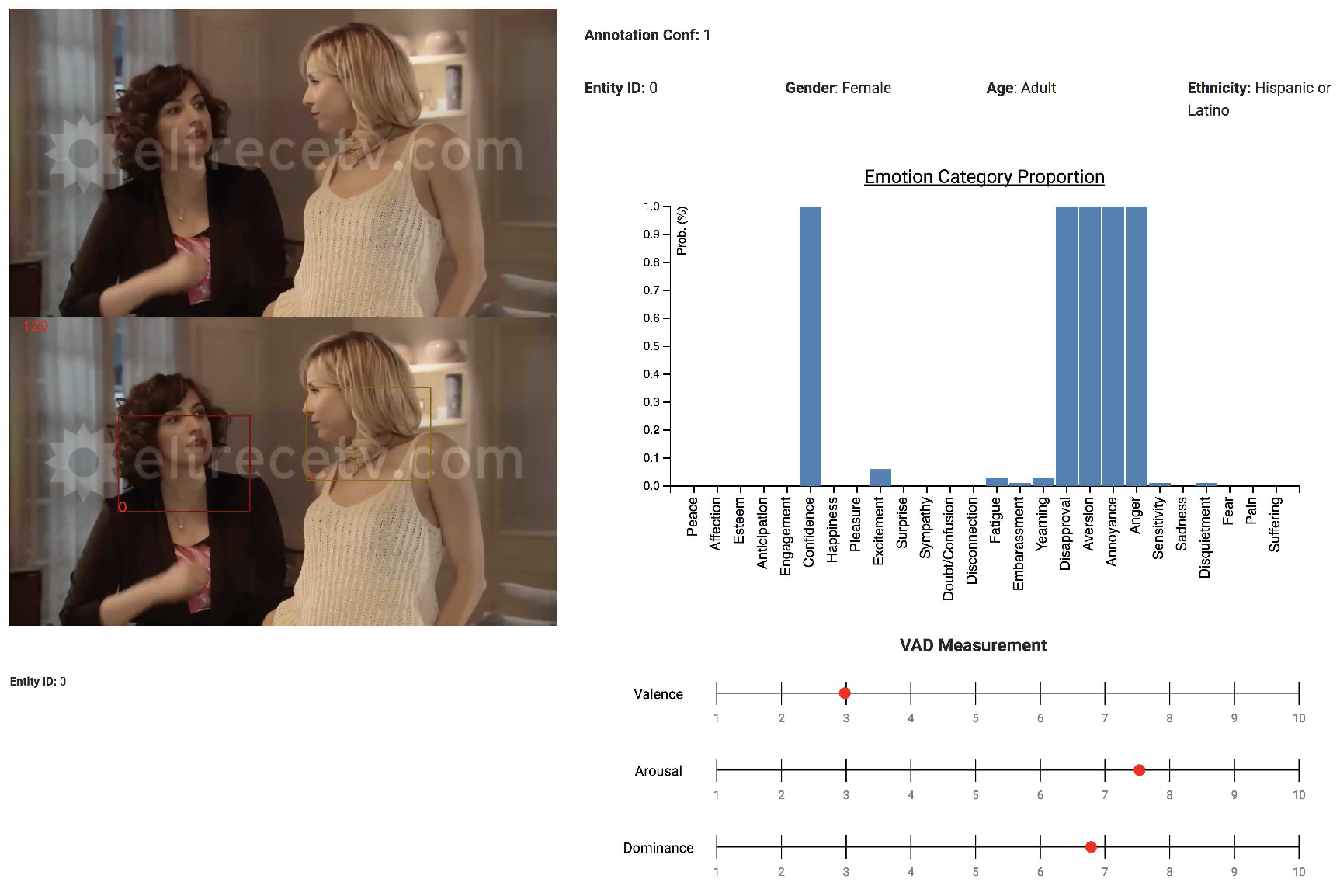}}&
  \subfloat[anger]{
    \includegraphics[height=1.4in]{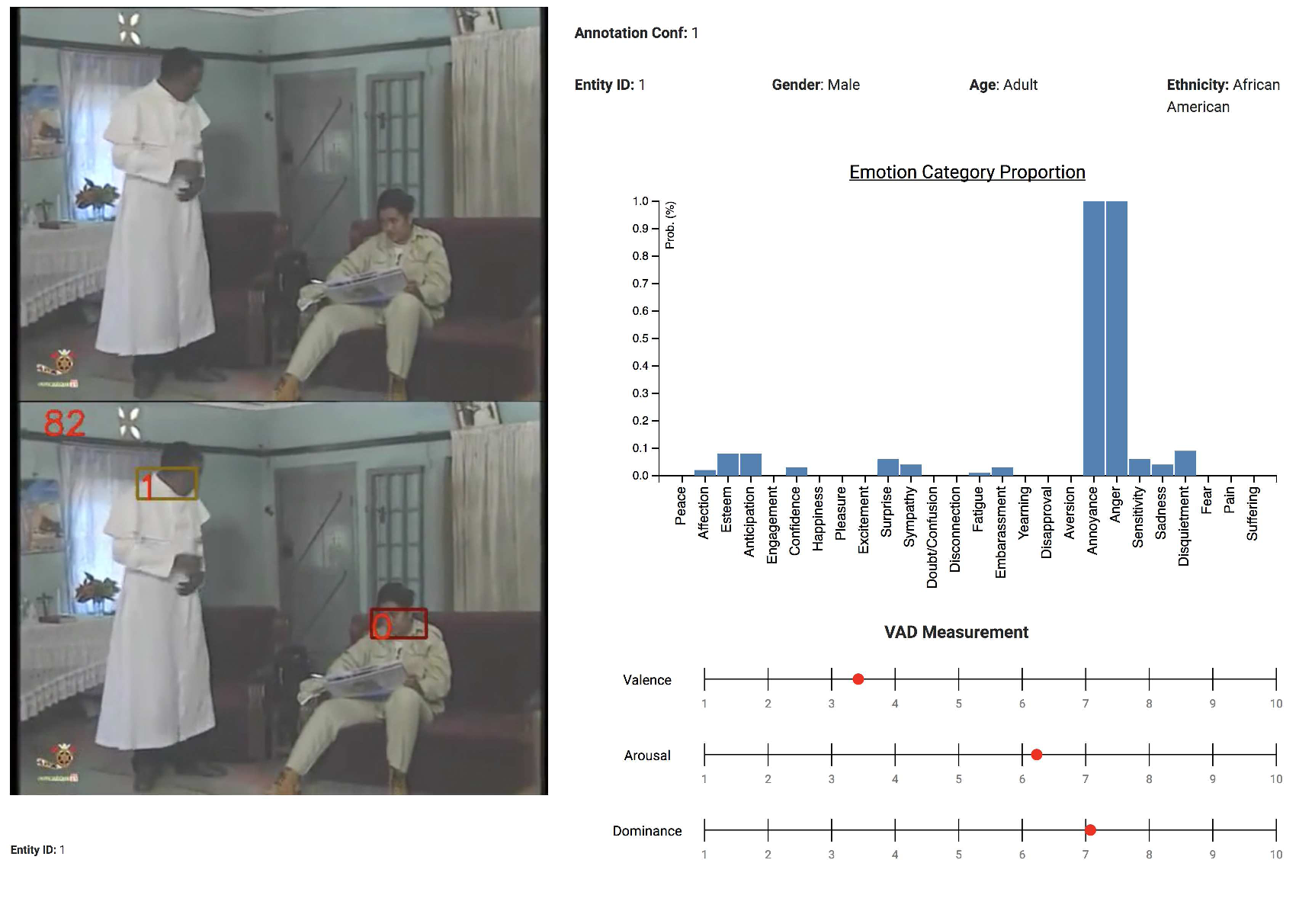}}&
     \subfloat[sensitivity]{
    \includegraphics[height=1.4in]{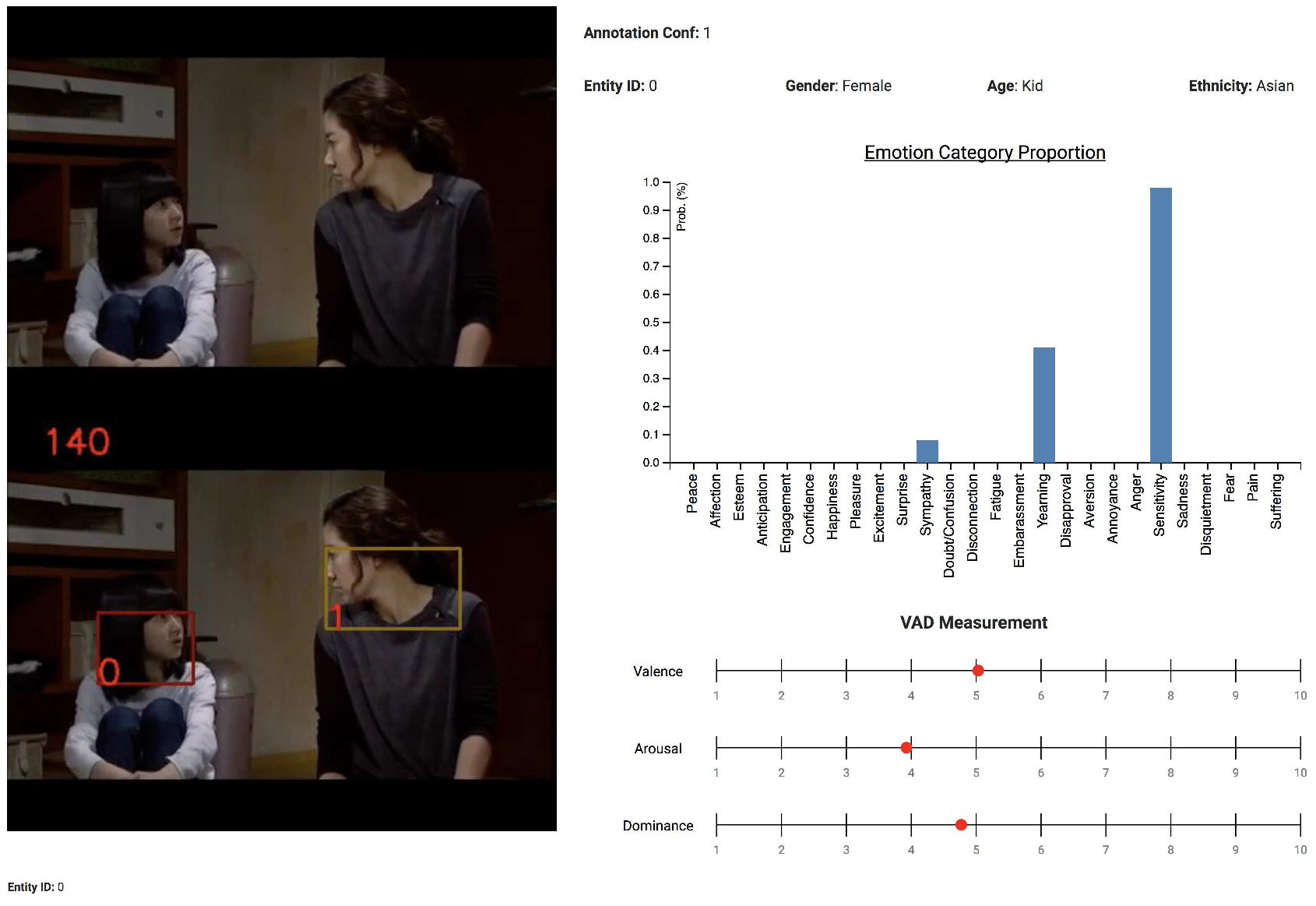}}\\
  \subfloat[sadness]{
\includegraphics[height=1.4in]{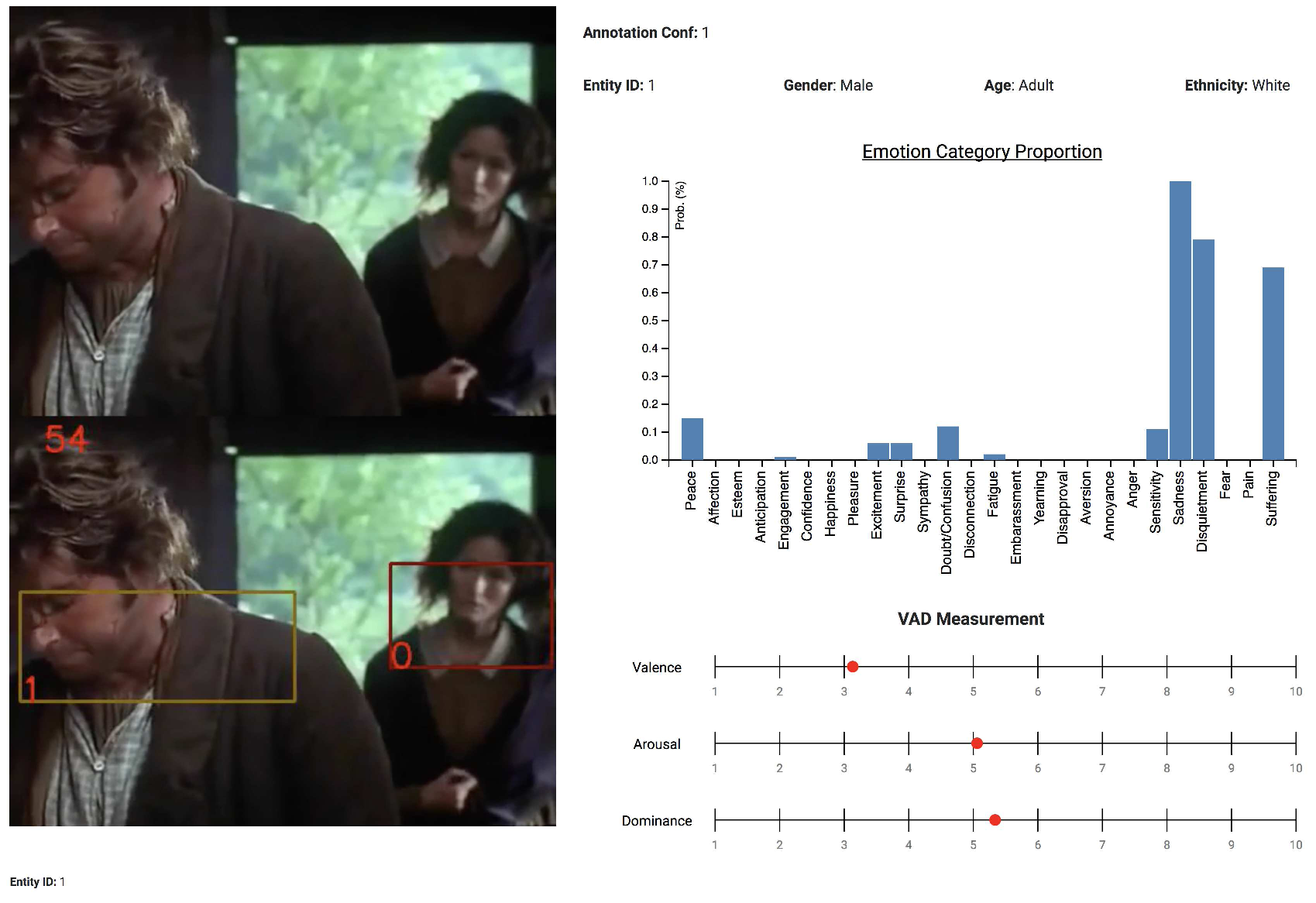}} &
  \subfloat[disquietment]{
    \includegraphics[height=1.4in]{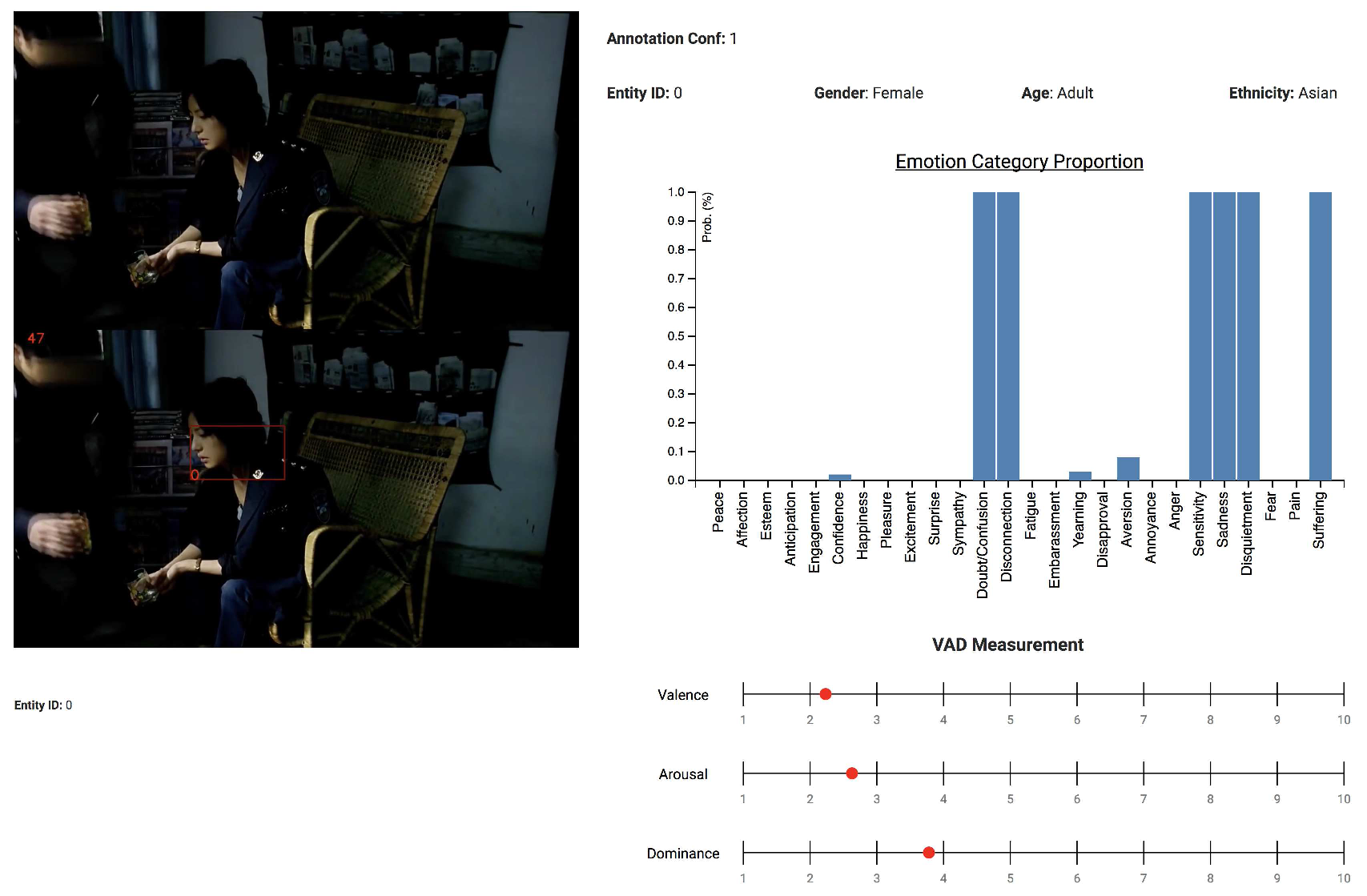}}&
  \subfloat[fear]{
    \includegraphics[height=1.4in]{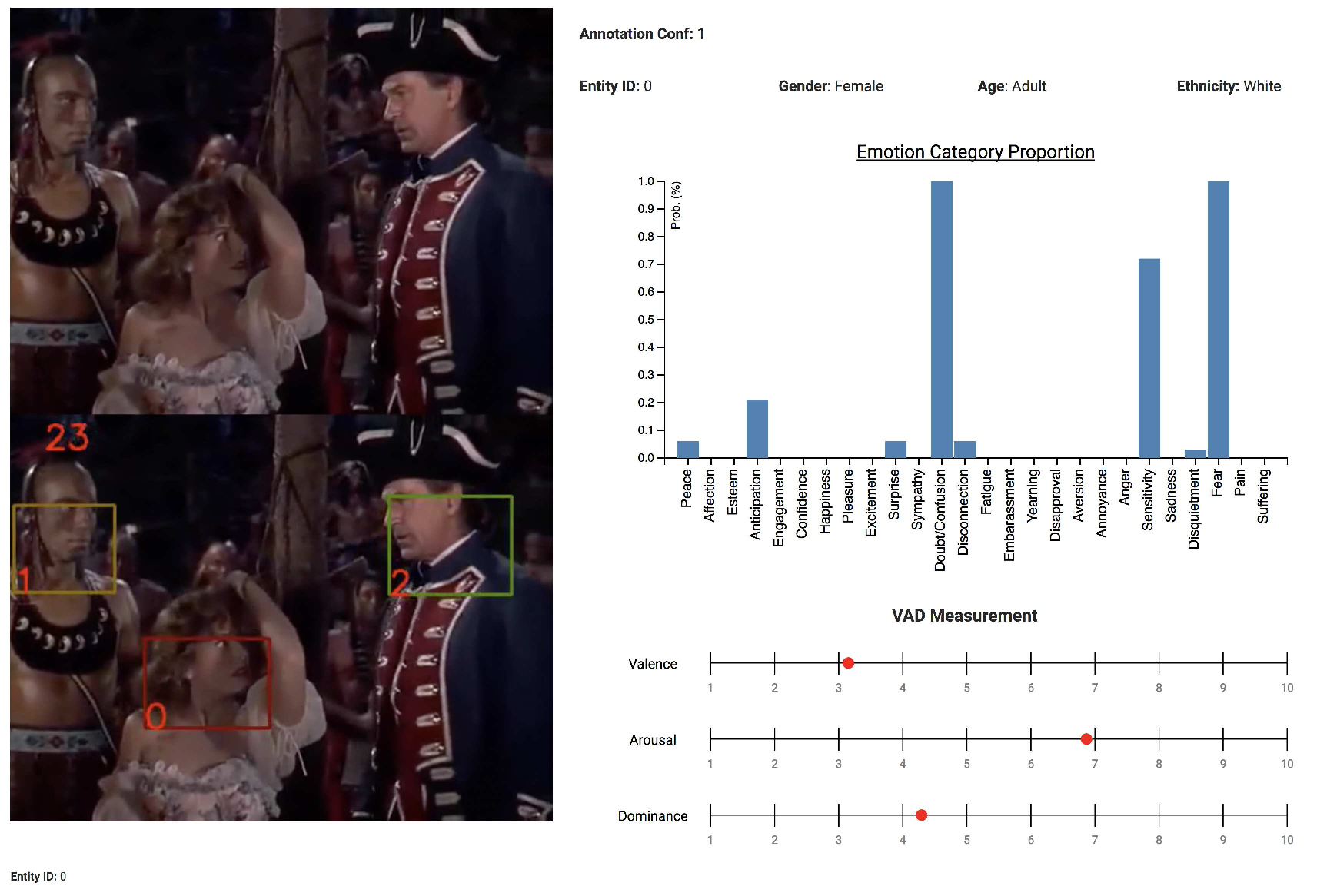}}\\
     \subfloat[pain]{
    \includegraphics[height=1.4in]{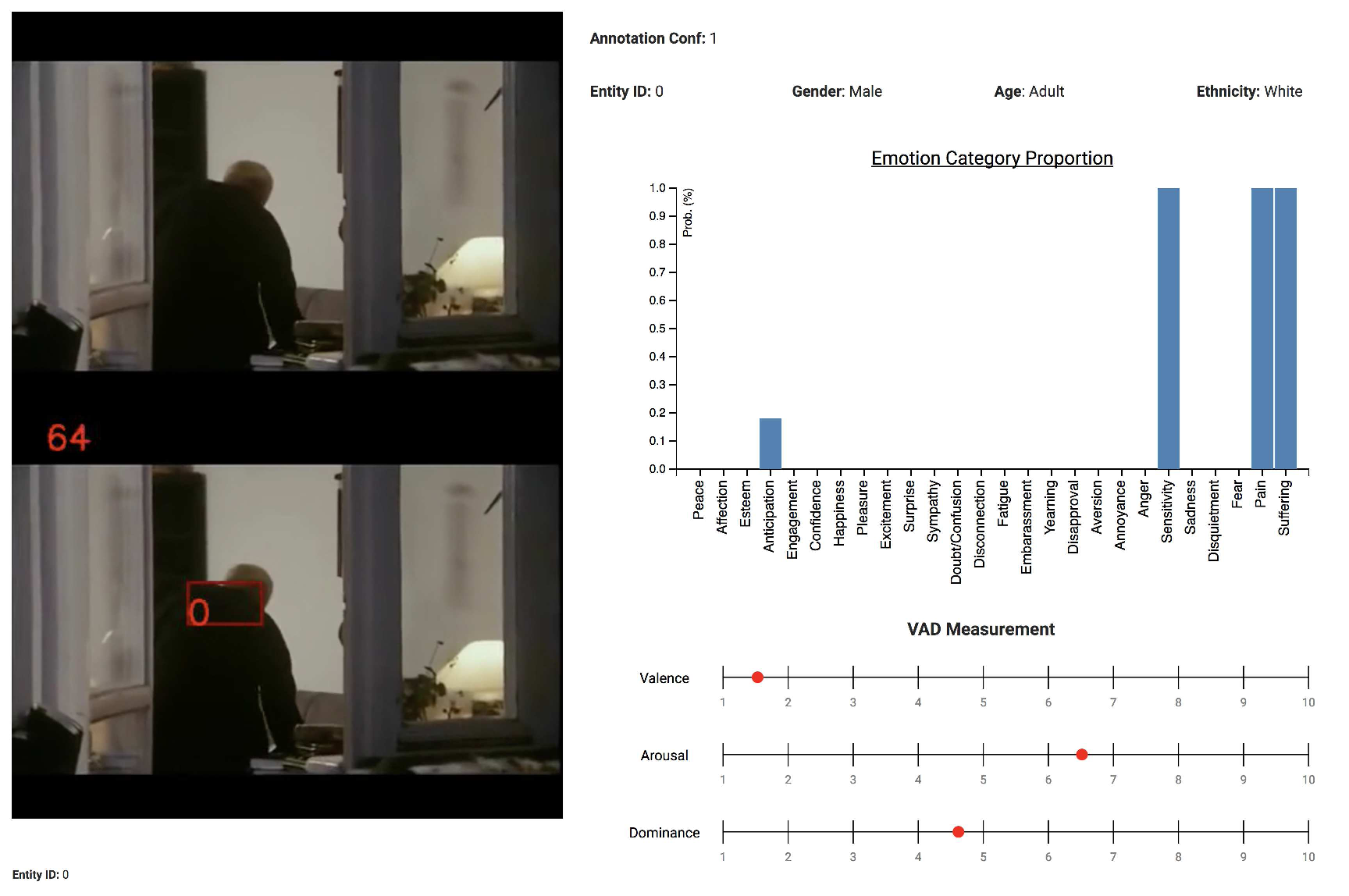}}&
  \subfloat[suffering]{
\includegraphics[height=1.4in]{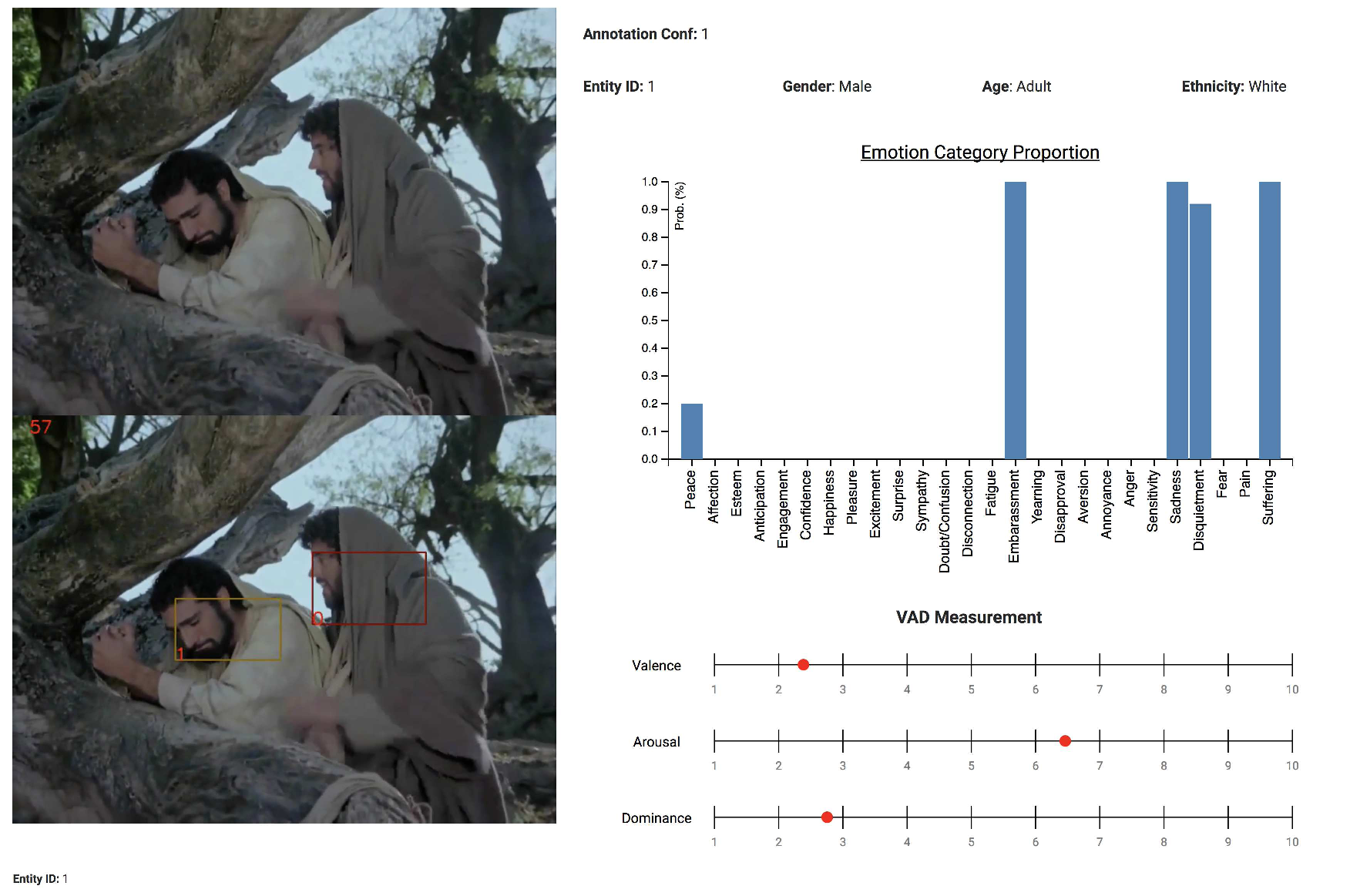}} \\
	  \subfloat[quality control]{
\includegraphics[height=1.4in]{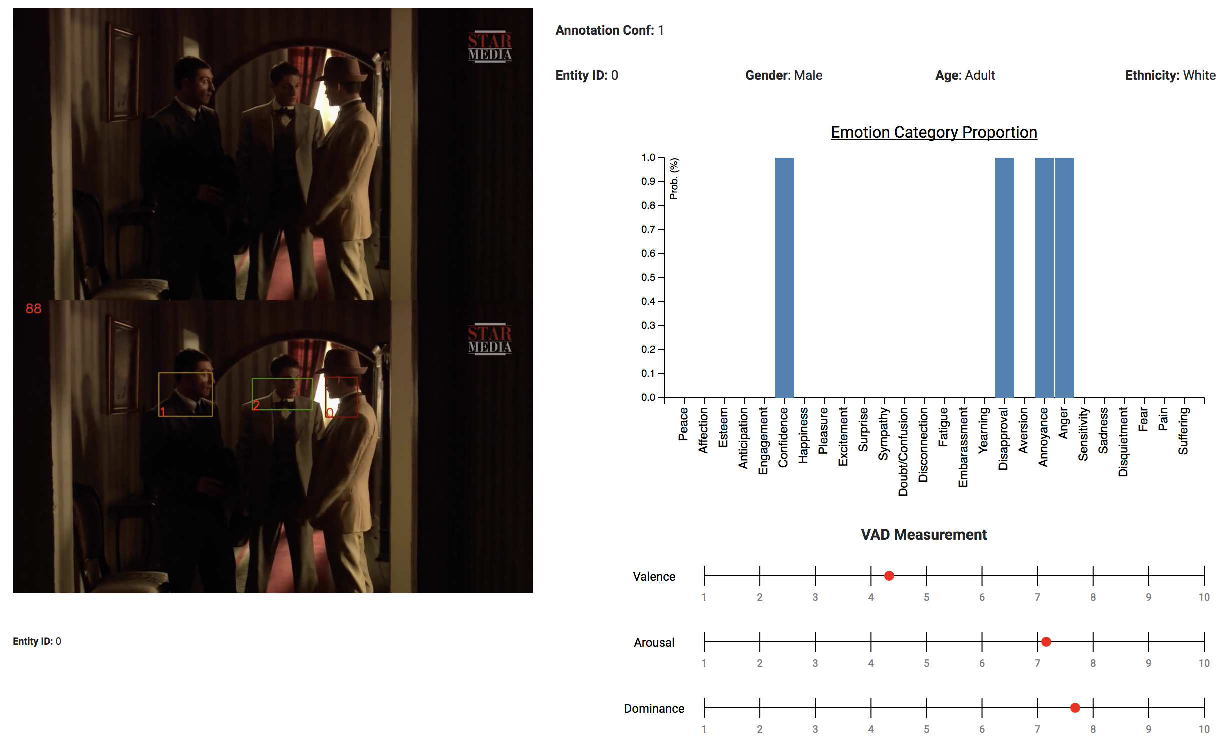}} &
	  \subfloat[quality control]{
\includegraphics[height=1.4in]{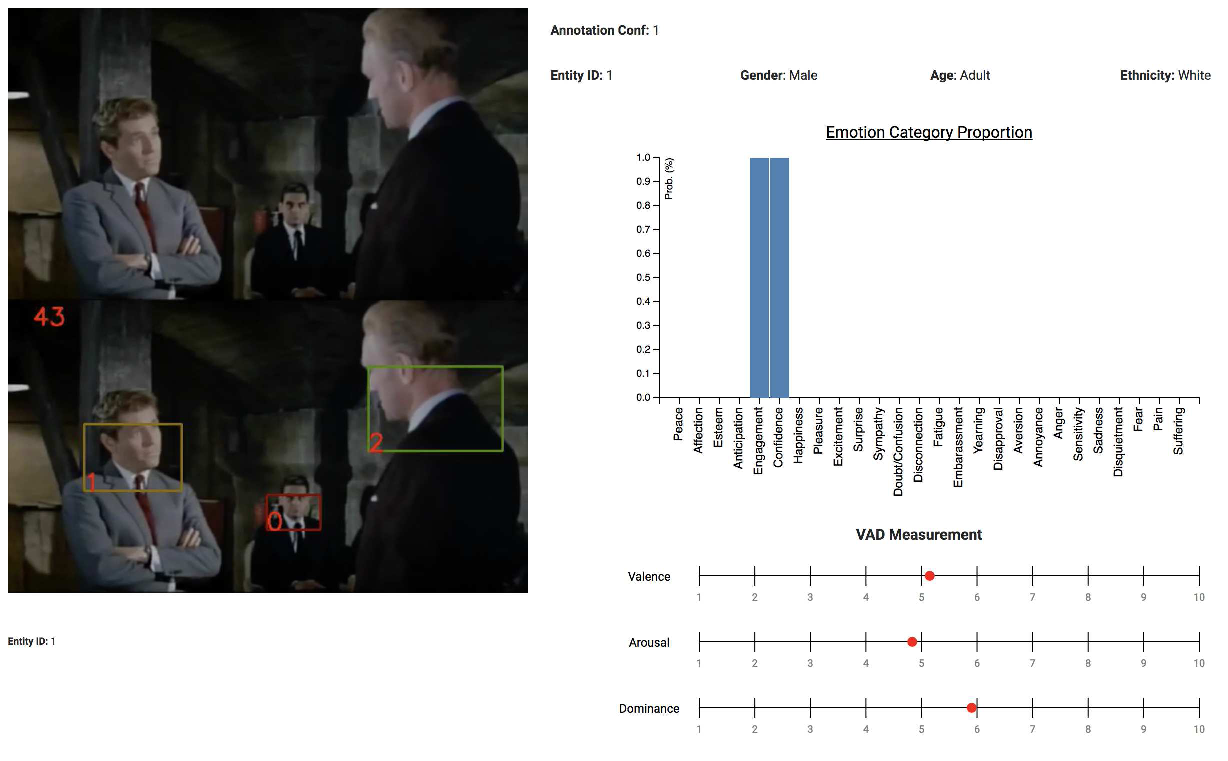}}&

    \end{tabular}
   \caption{(Continued from the previous page.) Examples of high-confidence instances in BoLD for the 26 categorical emotions and two instances (27 and 28) that were used for quality control. \revise{For each subfigure, the left side is a frame from the video, along with another copy that has the character entity IDs marked in a bounding box. The right side shows the corresponding aggregated annotation, annotation confidence $c$, demographics of the character, and aggregated categorical and dimensional emotion.}}
\end{figure*}

\begin{figure}[ht!]
  \centering
    \includegraphics[trim={0 13cm 0 13cm},clip,width=0.405\textwidth]{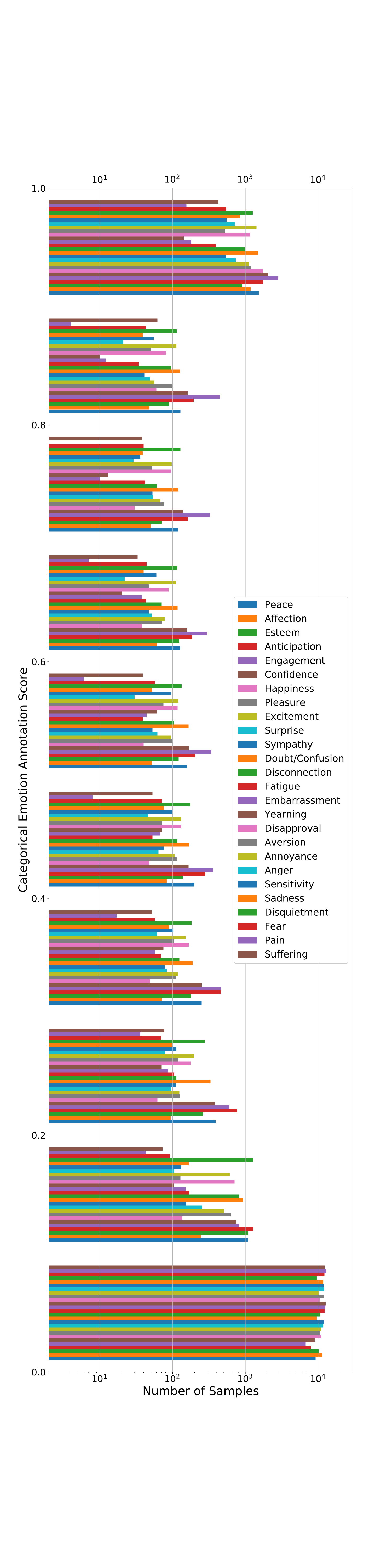}
    \caption{Distributions of the 26 different categorical emotions.}
   \label{fig:dataset_distrib}
\end{figure}

\begin{figure}[t!]
  \centering
\includegraphics[trim={0 0 1.0cm 1.0cm},clip,width=0.5\textwidth]{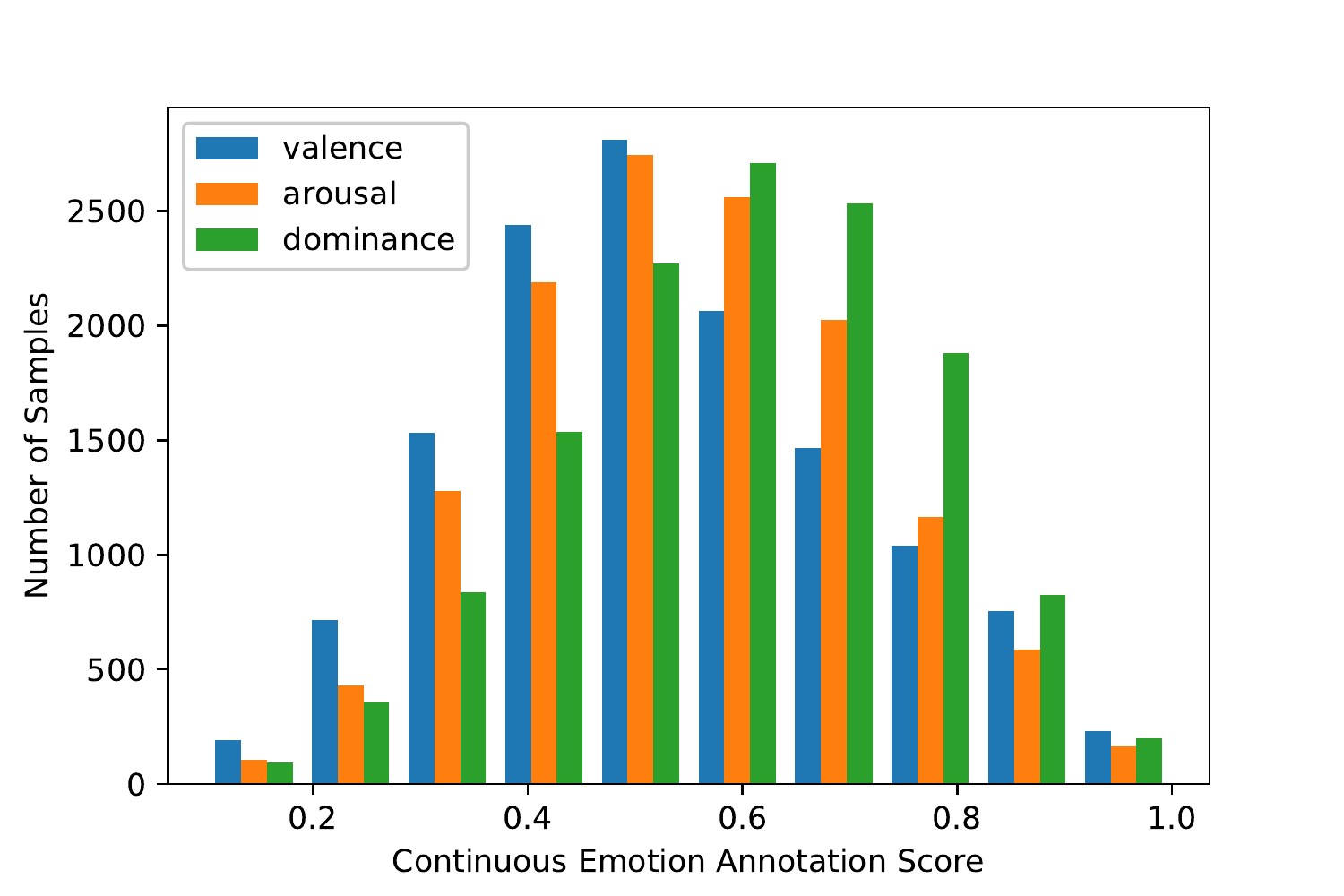}
    \caption{Distributions of the three dimensional emotion ratings: valence, arousal, and dominance.}
   \label{fig:dataset_distrib2}
\end{figure}

\begin{figure}[t!]
\includegraphics[width=0.48\textwidth]{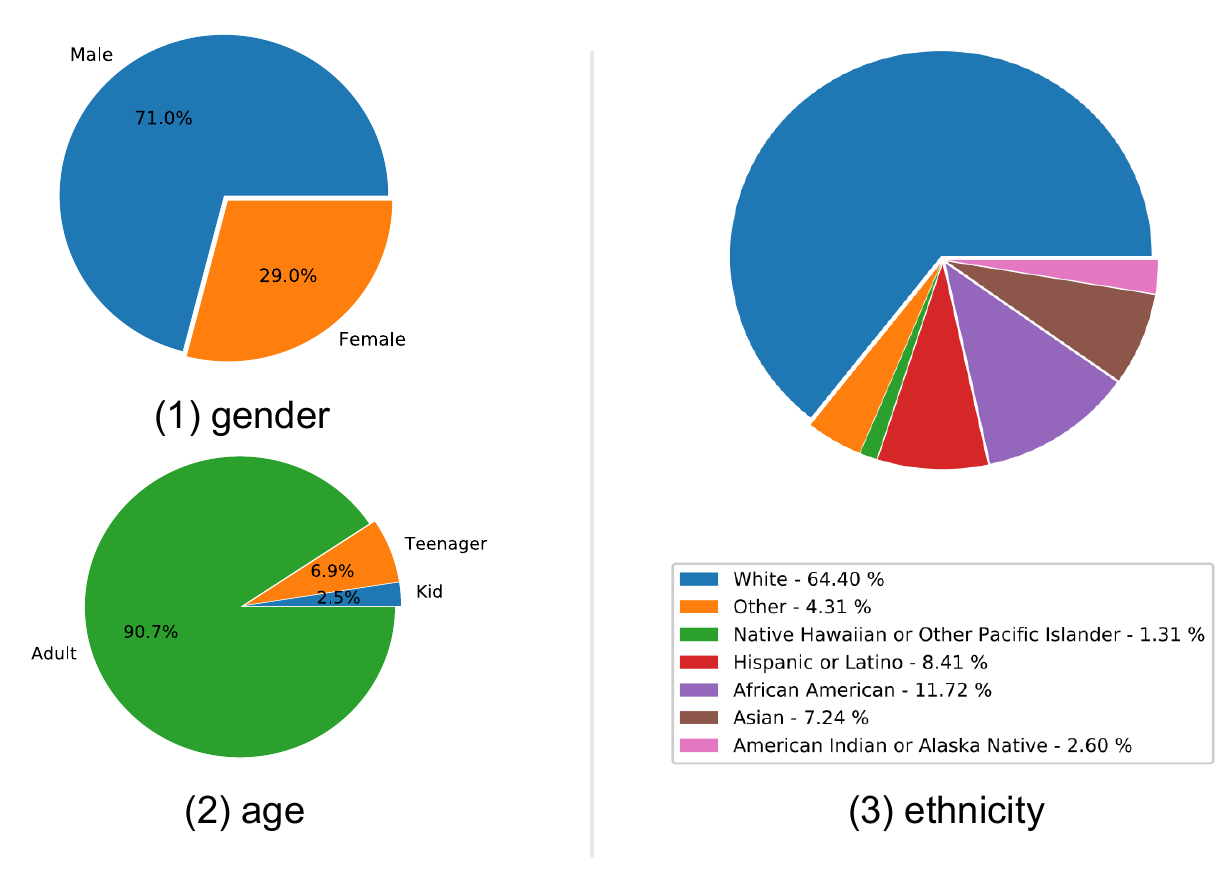}
\caption{Demographics of characters in our dataset.}
   \label{fig:dataset_distrib3}
\end{figure}


We report relevant dataset statistics. 
We used state-of-the-art statistical techniques to validate our quality control mechanisms and thoroughly understand the consensus level of 
our verified data labels. Because human perceptions of a character's emotions naturally varies across participants, we do not expect absolute consensus for collected labels. In fact, it is nontrivial to quantitatively understand and measure the quality of such affective data. 

\subsubsection{Annotation Distribution and Observations}

We have collected annotations for $13,239$ instances. The dataset continues to grow as more instances and annotations are added. Fig.~\ref{fig:data_examples} shows some high-confidence instances in our dataset. Figs.~\ref{fig:dataset_distrib}, \ref{fig:dataset_distrib2}, and \ref{fig:dataset_distrib3} show the distributions of categorical emotion, dimensional emotion, and demographic information, respectively. For each categorical emotion, the distribution is highly unbalanced. For dimensional emotion, the distributions of three dimensions are Gaussian-like, while valence is right-skewed and dominance is left-skewed. Character demographics is also unbalanced: most characters in our movie-based dataset are male, white, and adult. We partition all instances into three sets: the training set ($\sim$70\%, 9222), the validation set ($\sim$10\%, 1153), and the testing set (20\%, 2864). Our split protocol ensured that clips from the same raw movie video belong to the same set so that subsequent evaluations can be conducted faithfully.

\begin{figure*}[t!]
  \centering
    \includegraphics[trim={0cm 4cm 4.3cm 3cm},clip,width=0.95\textwidth]{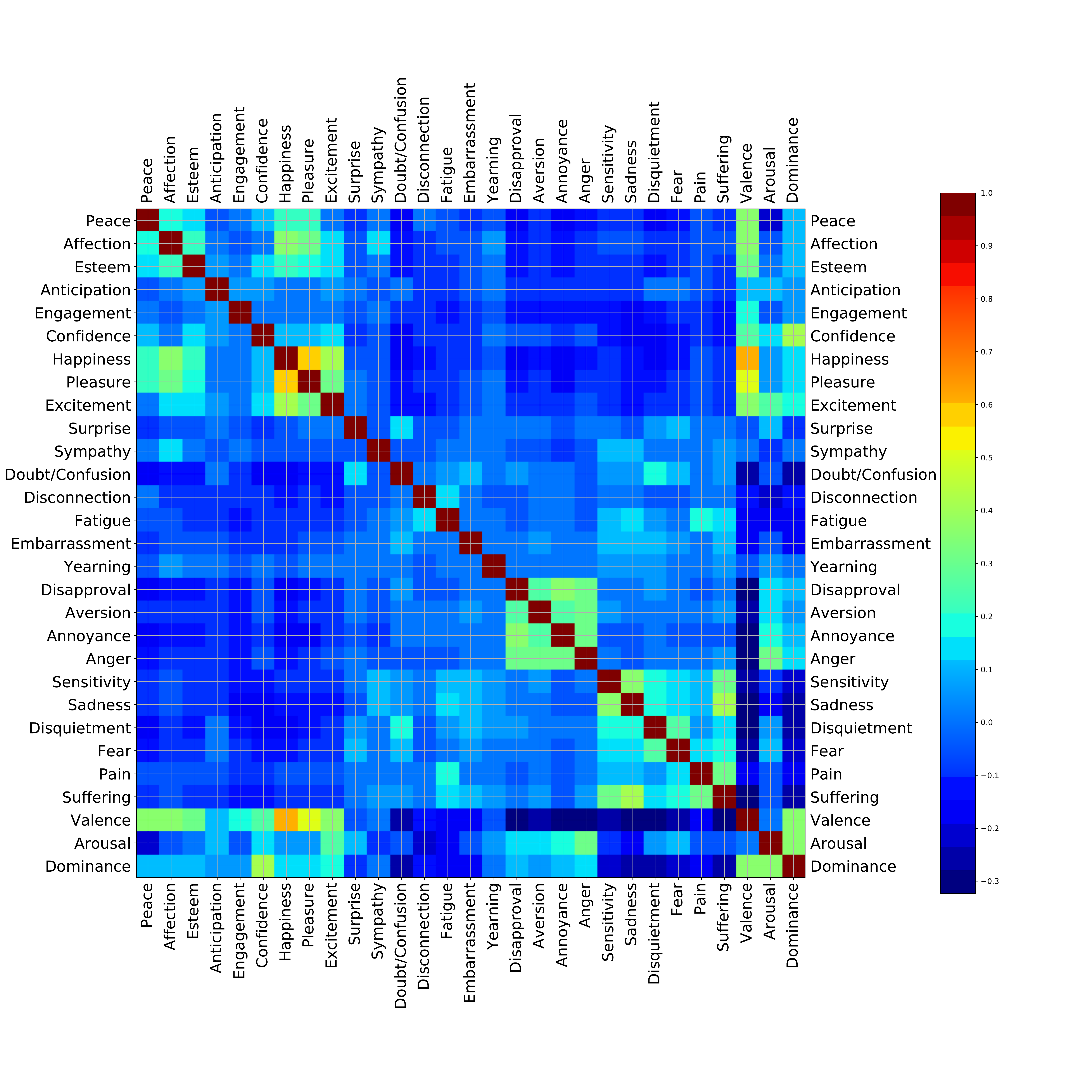}
    \caption{Correlations between pairs of categorical or dimensional emotions, calculated based on the BoLD dataset.}
   \label{fig:dataset_cm}
\end{figure*}

We observed interesting correlations between pairs of categorical emotions and pairs of dimensional emotions. Fig.~\ref{fig:dataset_cm} shows correlations between each pair of emotion categories. Categorical emotion pairs such as pleasure and happiness ($0.57$), happiness and excitement ($0.40$), sadness and suffering ($0.39$), annoyance and disapproval ($0.37$), sensitivity and sadness ($0.37$), and affection and happiness ($0.35$) show high correlations, matching our intuition. Correlations between dimensional emotions (valence and arousal) are weak ($0.007$). Because these two dimensions were designed to indicate independent characteristics of emotions, weak correlations among them confirm their validity. However, correlations between valence and dominance $0.359$, and between arousal and dominance ($0.356$) are high. This finding is evidence that dominance is not a strictly independent dimension in the VAD model.

We also observed sound correlations between dimensional and categorical emotions. Valence shows strong positive correlations with happiness ($0.61$) and pleasure ($0.51$), and strong negative correlations with disapproval ($-0.32$), sadness ($-0.32$), annoyance ($-0.31$), and disquitement ($-0.32$). Arousal shows positive correlations with excitement ($0.25$) and anger ($0.31$), and negative correlations with peace ($-0.20$), and disconnection ($-0.23$). Dominance shows strong correlation with confidence ($0.40$), and strong negative correlation with doubt/confusion ($-0.23$), sadness ($-0.28$), fear (-0.23), sensitivity ($-0.22$), disquitement ($-0.24$), and suffering ($-0.25$). All of these correlations match with our intuition about these emotions.

\subsubsection{Annotation Quality and Observations}

\begin{table*}[t!]
  \begin{center}
  \caption{Agreement among participants on categorical emotions and characters' demographic information.}
{\setlength{\tabcolsep}{0.7em}\renewcommand{\arraystretch}{1.35}
    \begin{tabular}{| c !{\color{light-gray}\vrule}
 c !{\color{light-gray}\vrule}
 c | c !{\color{light-gray}\vrule}
 c !{\color{light-gray}\vrule}
 c | c !{\color{light-gray}\vrule}
 c !{\color{light-gray}\vrule}
 c |} 
    \hline
    {\bf Category} & {\bf $\kappa$} & {\bf filtered $\kappa$} & {\bf Category} & {\bf $\kappa$} & {\bf filtered $\kappa$} & {\bf Category} & {\bf $\kappa$} & {\bf filtered $\kappa$} \\ \hline
 Peace & $0.132$ & $0.148$ & Affection & $0.262$ & $0.296$  &
 Esteem & $0.077$ & $0.094$ \\ \arrayrulecolor{light-gray}\hline\arrayrulecolor{black}
 Anticipation & $0.071$ & $0.078$  &
 Engagement & $0.110$ & $0.126$ & Confidence & $0.166$ & $0.183$ \\ \arrayrulecolor{light-gray}\hline\arrayrulecolor{black}
 Happiness & $0.385$ & $0.414$ &
 Pleasure & $0.171$ & $0.200$ &
 Excitement & $0.178$ & $0.208$ \\ \arrayrulecolor{light-gray}\hline\arrayrulecolor{black}
 Surprise & $0.137$ & $0.155$ &
 Sympathy & $0.114$ & $0.127$ & Doubt/Confusion & $0.127$ & $0.141$  \\ \arrayrulecolor{light-gray}\hline\arrayrulecolor{black}
 Disconnection & $0.125$ & $0.140$ & Fatigue & $0.113$ & $0.131$  &
 Embarrassment & $0.066$ & $0.085$ \\ \arrayrulecolor{light-gray}\hline\arrayrulecolor{black}
 Yearning & $0.030$ & $0.036$ &
 Disapproval & $0.140$ & $0.153$ & Aversion & $0.075$ & $0.087$  \\ \arrayrulecolor{light-gray}\hline\arrayrulecolor{black}
 Annoyance & $0.176$ & $0.197$ & Anger & $0.287$ & $0.307$  &
 Sensitivity & $0.082$ & $0.097$ \\ \arrayrulecolor{light-gray}\hline\arrayrulecolor{black}
  Sadness & $0.233$ & $0.267$ &
 Disquietment & $0.110$ & $0.125$ & Fear & $0.193$ & $0.214$  \\ \arrayrulecolor{light-gray}\hline\arrayrulecolor{black}
 Pain & $0.273$ & $0.312$ & Suffering & $0.161$ & $0.186$  & {\bf Average} & {\bf $0.154$} & {\bf $0.173$}\\ \arrayrulecolor{light-gray}\hline\arrayrulecolor{black}\hline 
 Gender  & $0.863$ & $0.884$ & Age & $0.462$ & $0.500$ &
 Ethnicity & $0.410$ & $0.466$  \\ \hline 
    \end{tabular}
}
\label{table:kappa}
  \end{center}      
\end{table*}


We computed Fleiss' Kappa score ($\kappa$) for each categorical emotion and categorical demographic information to understand the extent and reliability of agreement among participants. Perfect agreement leads to a score of one, while no agreement leads to a score less than or equal to zero. Table~\ref{table:kappa} shows Fleiss' Kappa \citep{gwet2014handbook} among participants on each categorical emotion and categorical demographic information. $\kappa$ is computed on all collected annotations for each category. For each category, we treated it as a two-category classification and constructed a subject-category table to compute Fleiss' Kappa. By filtering out those with low reliability scores, we also computed filtered $\kappa$. Note that some instances may have less than five annotations after removing annotations from low-reliability participants. We edited the way to compute $p_j$, defined as the proportion of all assignments which were to the $j$-th category. Originally, it should be
\begin{equation}
p_j = \frac{1}{N} \sum_{i=1}^N \frac{n_{ij}}{n}\;,
\label{eq1}
\end{equation}
where $N$ is the number of instances, $n_{ij}$ is the number of ratings annotators have assigned to the $j$-th category on the $i$-th instance, and $n$ is the number of annotators per instance. In our filtered $\kappa$ computation, $n$ varies for different instances and we denote the number of annotators for instance $i$ as $n_i$. Then Eq.~\eqref{eq1} is revised as:
\begin{equation}
p_j = \frac{1}{N} \sum_{i=1}^N \frac{n_{ij}}{n_i}\;.
\label{eq2}
\end{equation}
Filtered $\kappa$ is improved for each category, \revise{even for those objective category like gender,} which also suggests the validity of our offline quality control mechanism. Note that our reliability score is computed over dimensional emotions, and thus the offline quality control approach is complementary.
As shown in the table, affection, anger, sadness, fear, and pain have fair levels of agreement ($0.2< \kappa < 0.4$). Happiness has moderate level of agreement ($0.4 < \kappa < 0.6$), which is comparable to objective tasks such as age and ethnicity. This result indicates that humans are mostly consistent in their 
sense of happiness.
Other emotion categories fall into the level of slight agreement ($0< \kappa < 0.2$). Our $\kappa$ score of demographic annotation is close to previous studies reported in \citep{biel2013youtube}. 
Because the annotation is calculated from the same participant population, $\kappa$ also represents how difficult or subjective the task is. Evidently gender is the most consistent (hence the easiest) task among all categories. The data confirms that emotion recognition is both challenging 
and subjective even for human beings with sufficient level of EQ. 
Participants in our study passed an EQ test designed to measure one's ability
to sense others' feelings as well as response to others' feelings, and we suspect that individuals we excluded due to a failed EQ test would likely experience greater difficulty in recognizing emotions.


For dimensional emotions, we computed both across-annotation variances and within-instance annotation\break variances. The variances across all annotations are $5.87$, $6.66$, and $6.40$ for valence, arousal, and dominance, respectively. Within-instance variances (over different annotators) is computed for each instance and the means of these variances are $3.79$, $5.24$, and $4.96$, respectively. Notice that for the dimensions, the variances are reduced by $35\%$, $21\%$, and $23\%$, respectively, which illustrates human performance at reducing variance given concrete examples. Interestingly, participants are better at recognizing positive and negative emotions ({\it i.e.} valence) than in other dimensions.

\subsubsection{Human Performance} \label{sec:human_performance}

\begin{figure}[t!]
\centering
 \subfloat{
    \includegraphics[trim={0cm 0 0 0.8cm},clip,width=2.5in]
    {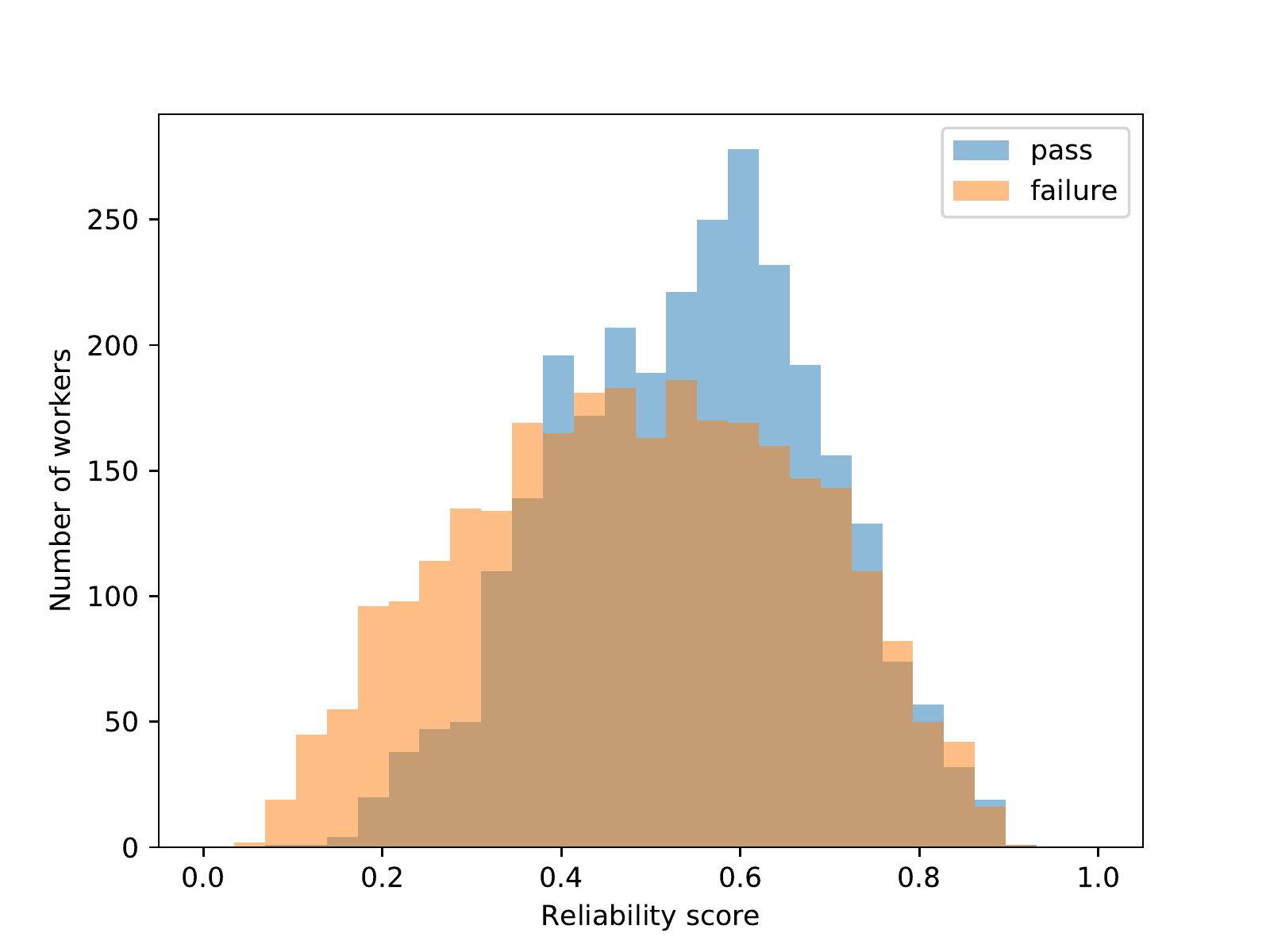}}
    \caption{Reliability score distribution among low-performance participants (failure) and non low-performance participants (pass). }
    \label{fig:rule_vs_glba}
\end{figure}
We explored the difference between low-performance participants and low reliability-score participants. As\break shown in Fig. \ref{fig:rule_vs_glba}, low-performance participants shows lower reliability score by average. While a significantly large number of low-performance participants have\break rather high reliability scores, most non-low-performance participants have reliability scores larger than $0.33$.\break These distributions suggests that participants who pass annotation sanity checks and relaxed gold standard tests are more likely to be reliable. However, participants who fail at those tests may still be reliable. Therefore, conventional quality control mechanism like the gold standard is insufficient when it comes to affect data.

\begin{figure*}[t!]
  \centering
   \begin{tabular}{ccc}
   \subfloat[$F1$ score]{\label{fig:cat_worker_eval1}
    \includegraphics[trim={1cm 0 2.4cm 0.8cm},clip,width=2.1in]{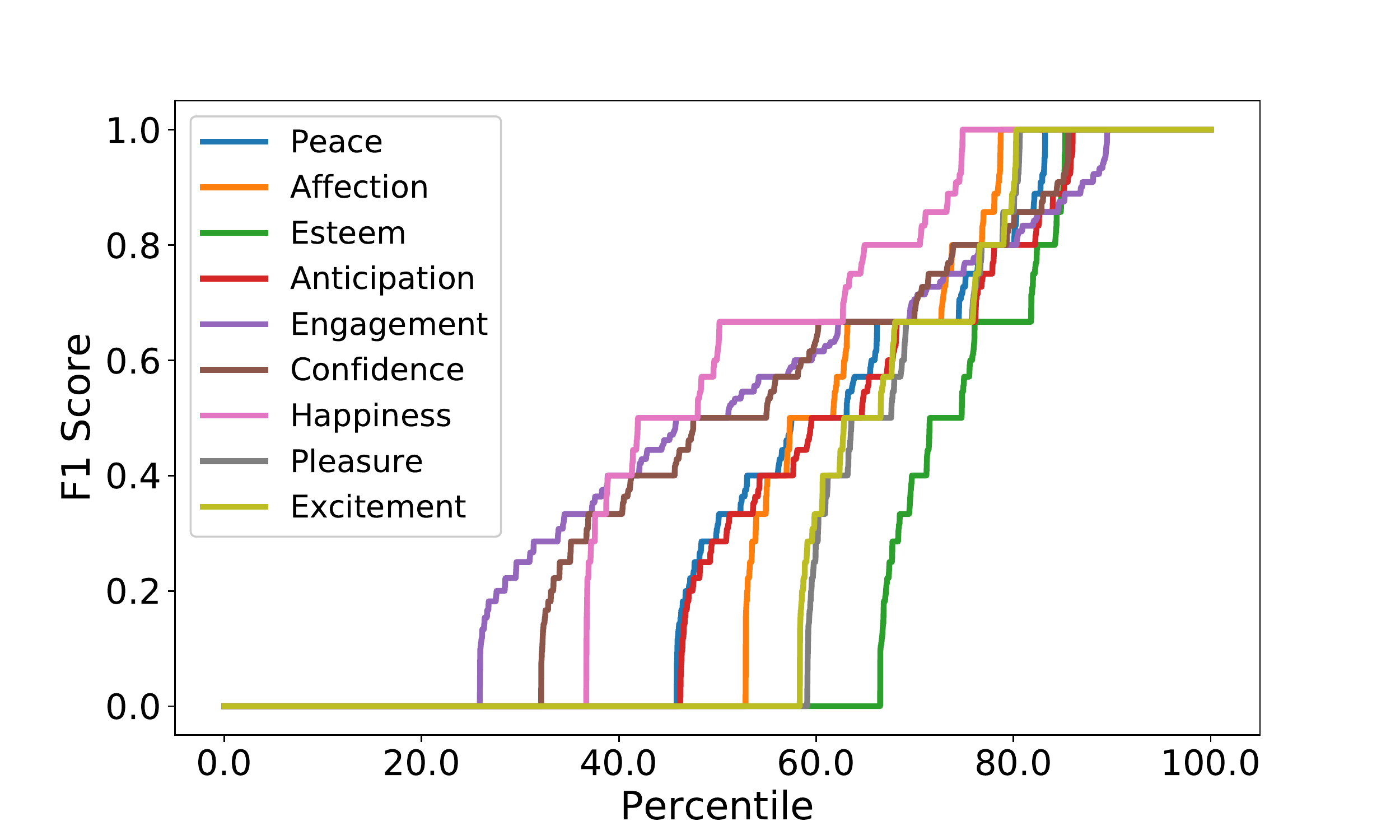}}&
    \subfloat[$F1$ score]{\label{fig:cat_worker_eval2}
    \includegraphics[trim={1cm 0 2.4cm 0.8cm},clip,width=2.1in]{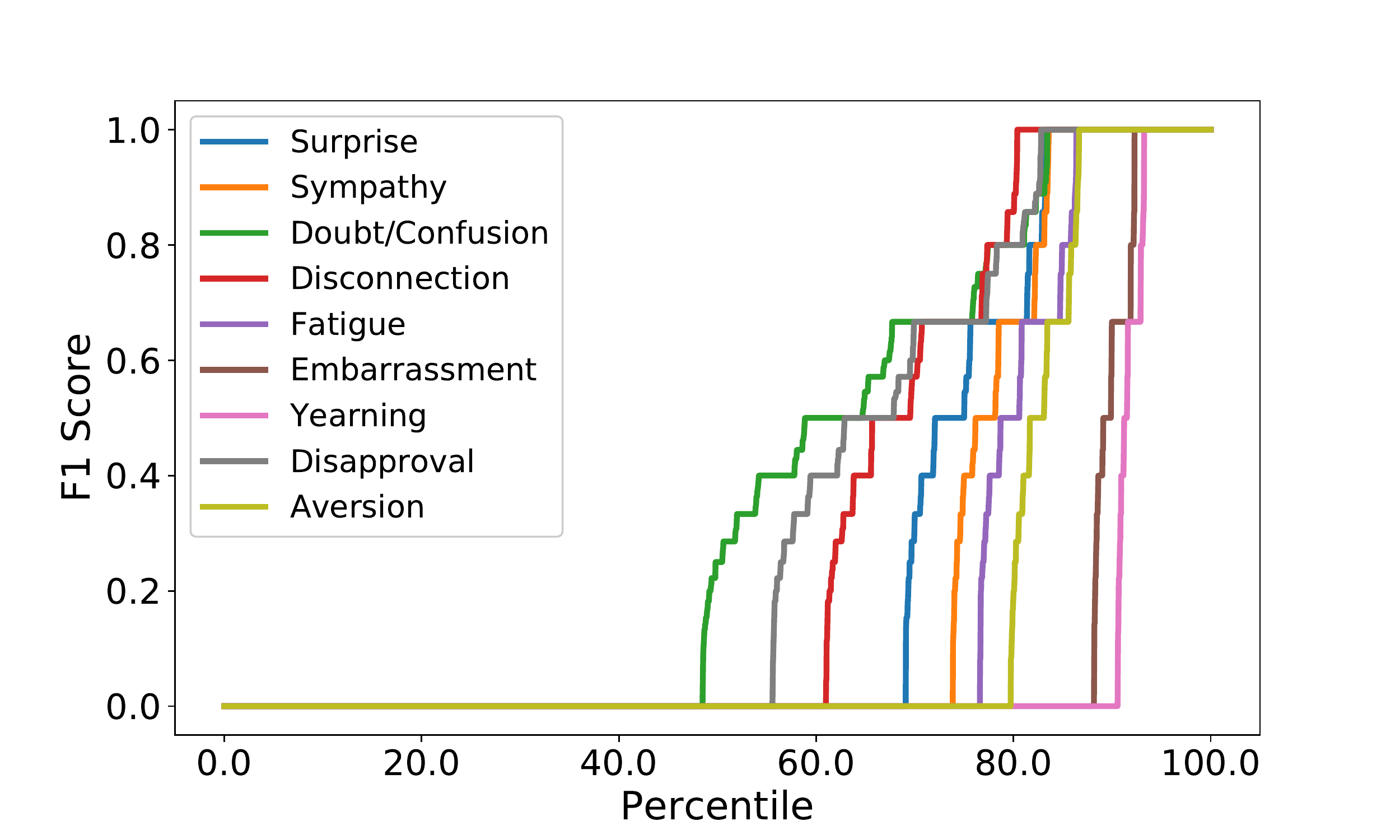}}&
    \subfloat[$F1$ score]{\label{fig:cat_worker_eval3}
    \includegraphics[trim={1cm 0 2.4cm 0.8cm},clip,width=2.1in]{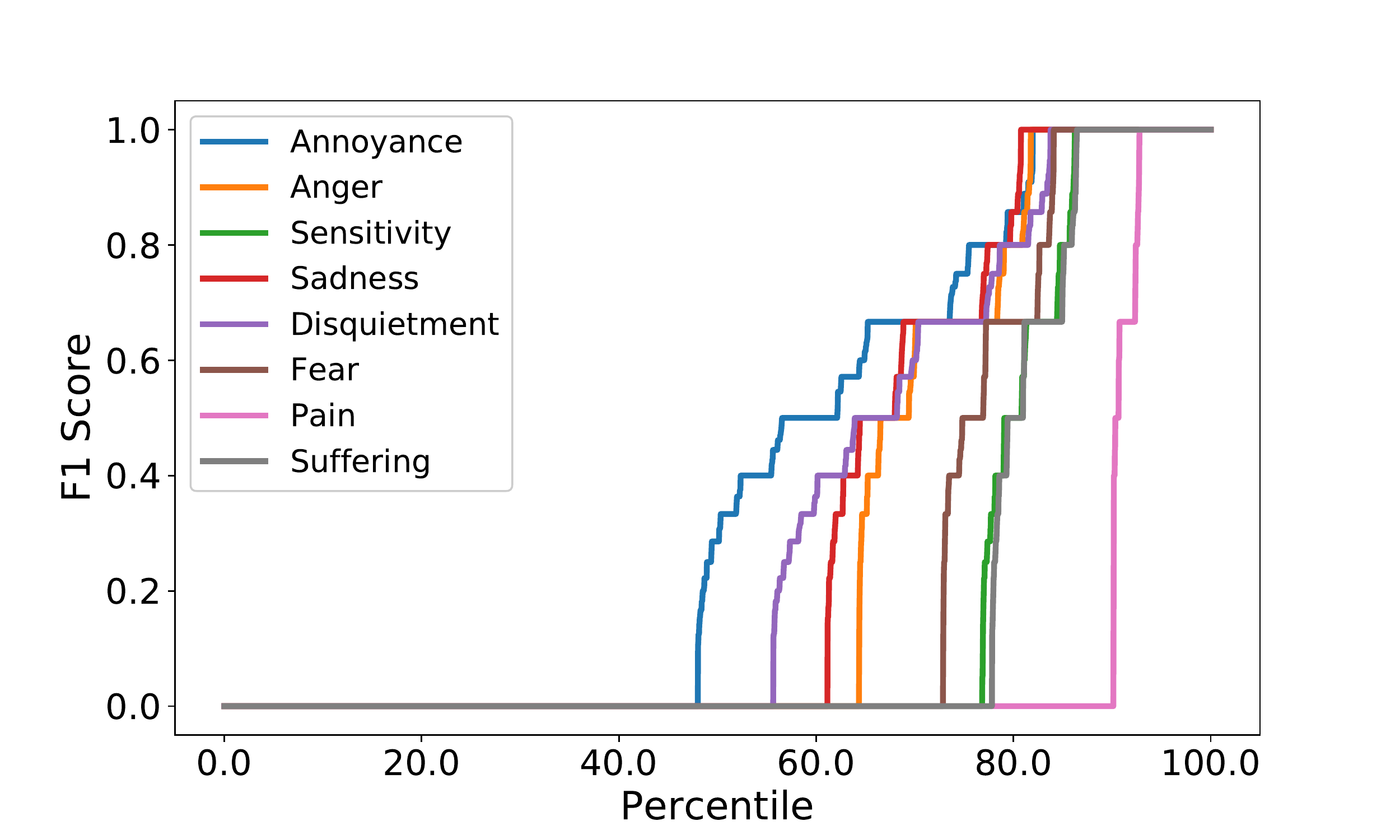}}
  \end{tabular}
  \begin{tabular}{cc}
  \subfloat[$R^2 score$]{\label{fig:vad_worker_eval_r2}
    \includegraphics[trim={0 0 0 0.8cm},clip,width=3.1in]{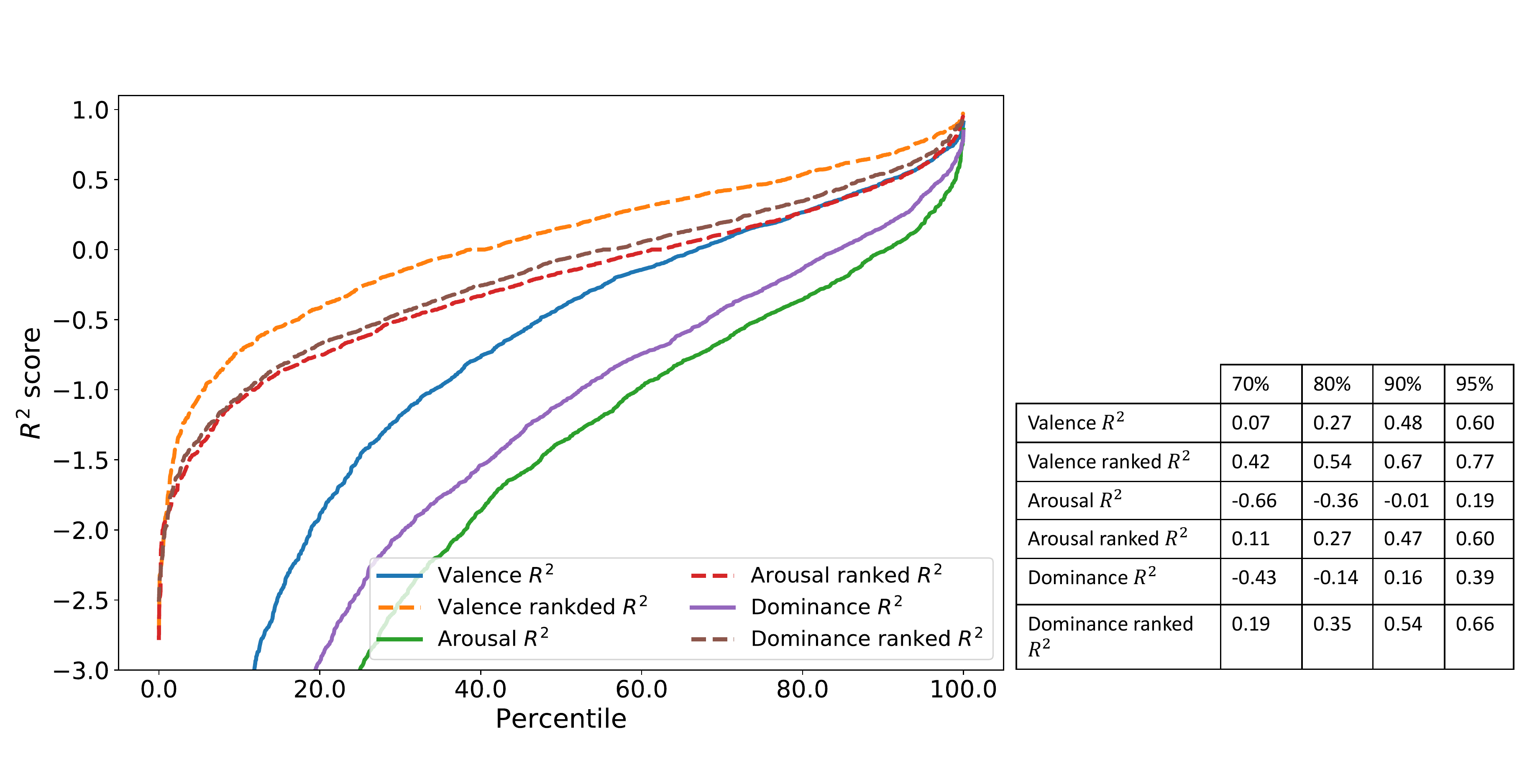}}&
    \subfloat[MSE]{\label{fig:vad_worker_eval_mse}
    \includegraphics[trim={0 0 0 1.0cm},clip,width=3.1in]{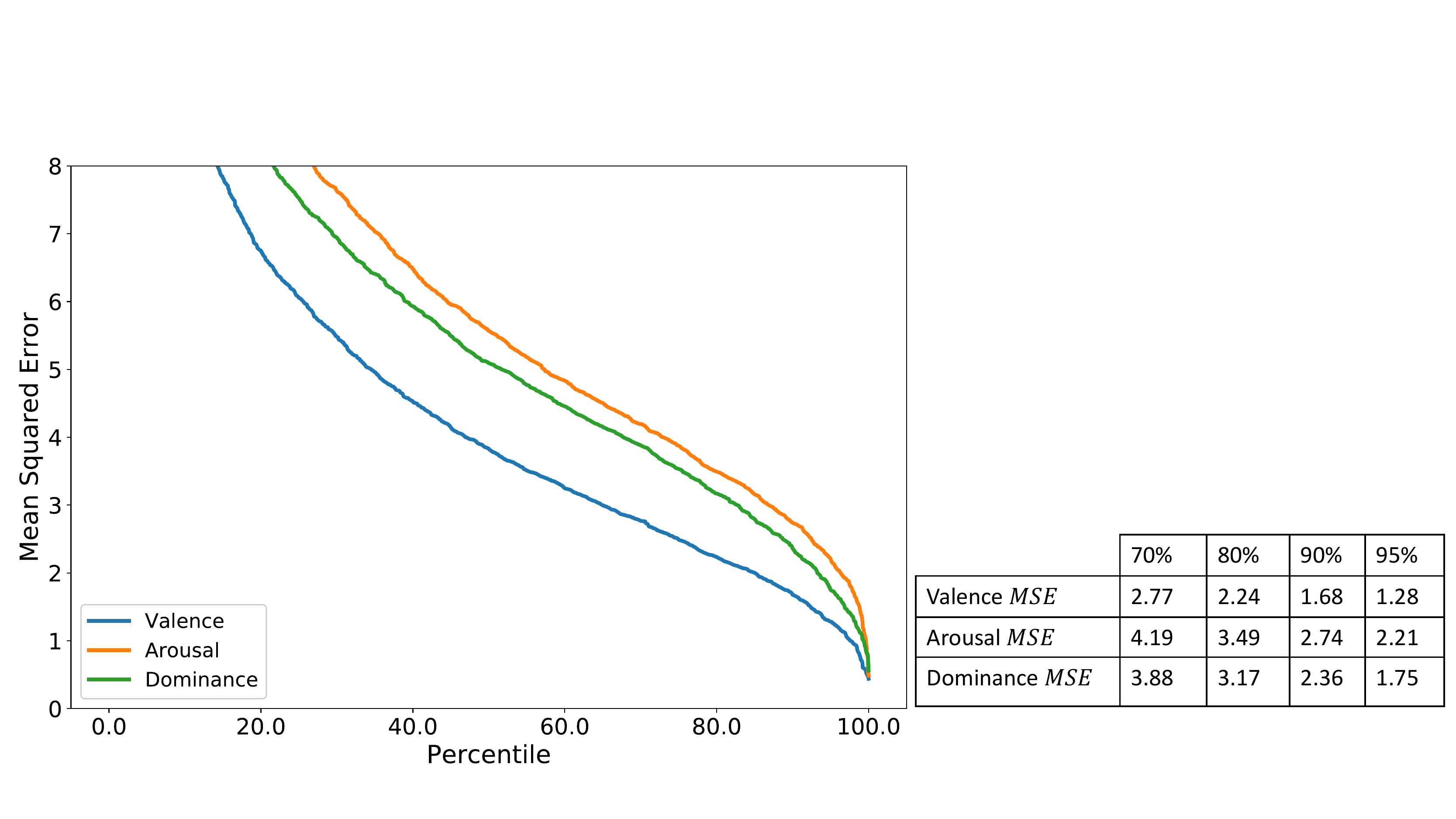}}
   \end{tabular}
    \caption{Human regression performance on dimensional emotions. X-axis: participant population percentile. Y-axis: $F1$, $R^2$ and MSE score. 
    Tables inside each plot in the second row summarize top 30\%, $20\%$, $10\%$, and $5\%$ participant regression scores.}
   \label{fig:worker_score}
\end{figure*}

We further investigated how well humans can\break achieve on emotion recognition tasks. There are $5,650$ AMT participants contributing to our dataset annotation.\break 
They represent over 100 countries (including 3,421 from the USA and 1,119 from India), with 48.4\% male and 51.6\% female, and an average age of 32. In terms of ethnicity,  57.3\% self-reported as White, 21.2\% Asian, 7.8\% African American, 7.1\% Hispanic or Latino, 1.6\% American Indian or Alaskan Native,  0.4\% Native Hawaiian or Other Pacific Islander, and 4.5\% Other.
For each participant, we used annotations from other participants and aggregated final dataset annotation to evaluate the performance. We treated this participant's annotation as prediction from an oracle model and calculate $F1$ score for categorical emotion, and coefficient of determination ($R^2$) and mean squared error (MSE) for dimensional emotion to evaluate the participant's performance. Similar to our standard annotation aggregation procedure, we ignored instances with a confidence score less than $0.95$ when dealing with dimensional emotions. Fig.~\ref{fig:worker_score} shows the cumulative distribution of participants' $F1$ scores of categorical emotions, the $R^2$ score, and the MSE score of dimensional emotion, respectively. We calculated vanilla $R^2$ score and rank percentile-based $R^2$ score. For the latter, we used rank percentile for both prediction and the ground truth. The areas under the curves (excluding Fig.~\ref{fig:worker_score}(5)) can be interpreted as how difficult it is for humans to recognize the emotion. For example, humans are effective at recognizing happiness while ineffective at recognizing yearning. Similarly, humans are better at recognizing the level of valence than that of arousal or dominance. These results reflect the challenge of achieving high classification and regression performance for emotion recognition even for human beings.

\subsubsection{Demographic Factors}
Culture, gender, and age could be important factors of emotion understanding. As mentioned in Section~\ref{sec:qc}, we have nine quality control videos in our crowdsourcing process that have been annotated for emotion more than 300 times. We used these quality control videos to test whether the annotations are independent of annotators' culture, gender, and age.

For categorical annotations (including both categorical emotions and categorical character demographics), we conducted $\chi^2$ test on each video. For each control instance, we calculated the p-value of the $\chi^2$ test over annotations (26 categorical emotions and 3 character demographic factors) from different groups resulting from annotators' three demographic factors. This process results in $29\times 3=87$ p-value scores for each control instance. For each test among 87 pairs, we further counted the total number of videos with significant p-value ($p<0.01$ or $p<0.001$). 
Interestingly, there is significant dependence over characters' ethnicity and annotators' ethnicity (9 out of 9, $p<0.001$). It is possible that humans are good at recognizing the ethnicity of others in the same ethnic group.
Additionally, there is intermediate dependence between annotators' ethnicity and categorical emotions (17 out of $26 \times 9 = 234$, $p<0.001$). We did not find strong dependence over other tested pairs (less than 3 out of 9, $p< 0.001$). 
This lack of dependence seems to suggest that a person's understanding of emotions  depends more on their own ethnicity than on their age or gender.

For VAD annotation, we conducted one-way ANOVA tests on each instance. 
For each control instance, we calculated p-value of one-way ANOVA test over VAD (3) annotations from different groups resulting from annotators' demographic factors (3).
This results in $3\times 3 = 9$ p-value scores for each control instance.
We also conducted Kruskal-Wallis H-test and found similar results. We report p-value of one-way ANOVA tests. Our results show that gender and age have little effect (less than 8 out of $9\times (3 + 3) = 54$, $p < 0.001$) on emotion understanding, while ethnicity has a strong effect (13 out of $9\times 3 = 27$, $p < 0.001$) on emotion understanding. Specifically, participants with different ethnicities have different understandings regarding valence for almost all control clips (7 out of 9, $p < 0.001$). Fig.~\ref{fig:data_examples}(27-28) shows two control clips. For Fig.~\ref{fig:data_examples}(27), valence average of person $0$ among Asians is $5.56$, yet $4.12$ among African Americans and $4.41$ among Whites. However, arousal average among Asians is $7.20$, yet $8.27$ among African Americans and $8.21$ among Whites. For Fig.~\ref{fig:data_examples}(28), valence average of person $1$ among Asians is $6.30$, yet $5.09$ among African Americans and $4.97$ among Whites. However, arousal average among Asians is $7.20$, yet $8.27$ among African Americans and $8.21$ among Whites. Among all of our control instances, the average valence among Asians is consistently higher than among Whites and African Americans. This repeated finding seems to suggest that Asians tend to assume more positively when interpreting others' emotions.

\subsubsection{Discussion}
\revise{Our data collection efforts offer important lessons. The efforts confirmed that reliability analysis is useful for collecting subjective annotations such as emotion labels when no gold standard ground truth is available. As shown in Table~\ref{table:kappa}, consensus (filtered $\kappa$ value) over high-reliable participants is higher than that of all participants ($\kappa$ value). This finding holds for both subjective questions (categorical emotion) and objective questions (character demographics), even though the reliability score is calculated with the different VAD annotations --- an evidence that the score does not overfit. As an offline quality control component, the method we developed and used to generate reliability scores~\citep{ye2017probabilistic} is suitable for analyzing such affective data. For example, one can also apply our proposed data collection pipeline to collect data for the task of image aesthetics modeling~\citep{datta2006studying}. In addition to their effectiveness in quality control, reliability scores are very useful for resource allocation. With a limited annotation budget, it is more reasonable to reward highly-reliable participants rather than less reliable ones.}



\section{Bodily Expression Recognition}\label{sec:method}
\revise{In this section, we investigate two pipelines for automated recognition of bodily expression and present quantitative results for some baseline methods.}
Unlike AMT participants, who were provided with all the information regardless of whether they use all in their annotation process, the first computerized pipeline relied solely on body movements, but {\it not} on facial expressions, audio, or context. The second pipeline took a sequence of cropped images of the human body as input, without explicitly modeling facial expressions.

\subsection{Learning from Skeleton}
\subsubsection{Laban Movement Analysis}

\begin{table}[t!]
\caption{Laban Movement Analysis (LMA) features.\break ($f_i$: categories; $m$: number of measurements; dist.: distance;\break 
accel.: acceleration)}
	\begin{center}
{\setlength{\tabcolsep}{0.1em}\renewcommand{\arraystretch}{1.3}	
\begin{tabular}{c!{\color{light-gray}\vrule}cc|c!{\color{light-gray}\vrule}cc}
\hline
 $f_i$ & Description & $m$ &
$f_i$ & Description & $m$ \\  
\hline
$f_1$ & Feet-hip dist. & 4 &
$f_2$ & Hands-shoulder dist.& 4\\
$f_3$ & Hands dist.& 4 &
$f_4$ & Hands-head dist.& 4\\
$f_8$ & Centroid-pelvis dist.& 4 &
$f_9$ & Gait size (foot dist.)& 4\\
\arrayrulecolor{light-gray}\hline\arrayrulecolor{black}

$f_{29}$ & Shoulders velocity& 4 &
$f_{32}$ & Elbow velocity& 4\\
$f_{13}$ & Hands velocity&4 &
$f_{12}$ & Hip velocity& 4\\
$f_{35}$ & Knee velocity& 4&
$f_{14}$ & Feet velocity& 4\\
$f_{38}$ & Angular velocity & $4 C_{23}^{2}$ & 
& & \\
\arrayrulecolor{light-gray}\hline\arrayrulecolor{black}
$f_{30}$ & Shoulders accel.&4 &
$f_{33}$ & Elbow accel.& 4\\
$f_{16}$ & Hands accel.& 4 &
$f_{15}$ & Hip accel.& 4\\
$f_{36}$ & Knee accel.& 4 &
$f_{17}$ & Feet accel.& 4\\
$f_{39}$ & Angular accel. & $4 C_{23}^{2}$ &
& & \\
\arrayrulecolor{light-gray}\hline\arrayrulecolor{black}
$f_{31}$ & Shoulders jerk& 4 &
$f_{34}$ & Elbow jerk& 4\\
$f_{40}$ & Hands jerk& 4 &
$f_{18}$ & Hip jerk& 4\\
$f_{37}$ & Knee jerk& 4 &
$f_{41}$ & Feet jerk& 4\\
\arrayrulecolor{light-gray}\hline\arrayrulecolor{black}
$f_{19}$ & Volume& 4 &
$f_{20}$ & Volume (upper body)& 4\\
$f_{21}$ & Volume (lower body)& 4 &
$f_{22}$ & Volume (left side)& 4\\
$f_{23}$ & Volume (right side)& 4 &
$f_{24}$ & Torso height& 4\\
\hline
\end{tabular}
}
\label{table:feature}
	\end{center}
\end{table}

Laban notation, originally proposed by Rudolf Laban (\citeyear{laban1971mastery}), is used for documenting body movement of dancing such as ballet. Laban movement analysis (LMA) uses four components to record human body movements: body, effort, shape, and space. 
Body category represents structural and physical characteristics of the human body movements. It describes which body parts are moving, which parts are connected, which parts are influenced by others, and general statements about body organization. Effort category describes inherent intention of a movement. Shape describes static body shapes, the way the body interacts with something, the way the body changes toward some point in space, and the way the torso changes in shape to support movements in the rest of the body.
LMA or its equivalent notation systems are widely used in psychology for emotion analysis \citep{wallbott1998bodily,kleinsmith2006cross} and human computer interaction for emotion generation and classification \citep{aristidou2017emotion,aristidou2015emotion}. In our experiments, we use features listed in Table~\ref{table:feature}.

LMA is conventionally conducted for 3D motion capture data that have 3D coordinates of body landmarks. In our case, we estimated 2D pose on images using~\citep{cao2017realtime}. In particular, we denote $p_i^t \in R^2$ as the coordinate of the $i$-th joint at the $t$-th frame. As the nature of the data, our 2D pose estimation usually has missing values of joint locations and varies in scale. In our implementation, we ignored an instance if the dependencies to compute the feature are missing. To address the scaling issue, we normalized each pose by the average length of all visible limbs, such as shoulder-elbow and elbow-wrist. Let $\nu = \{(i,j) | \text{ joint } i \text{ and joint }\break j \text{ are visible}\}$ be the visible set of the instance. We computed normalized pose $\hat{p}_i^t$ by
\begin{equation}
\begin{split}
s =\frac{1}{T|\nu|}\sum_{(i,j) \in \nu } \sum_t^T \norm{p_i^t - p_j^t},\; 
\hat{p}_i^t = \frac{p_i^t }{s}\;.
\end{split}
\end{equation}

The first part of features in LMA, {\it body component}, captures the pose configuration. For $f_1$, $f_2$, $f_3$, $f_8$, and $f_9$, we computed the distance between the specified joints frame by frame. For symmetric joints like feet-hip distance, we used the mean of left-feat-hip and right-feat-hip distance in each frame. The same protocol was applied to other features that contains symmetric joints like hands velocity. For $f_4$, the centroid was averaged over all visible joints and pelvis is the midpoint between left hip and right hip. This feature is designed to represent barycenter deviation of the body.

\begin{figure}
\centering
\begin{tabular}{cc}
 \subfloat[natural human skeleton]{
    \includegraphics[trim={0cm 0 0 0cm},clip,width=1.5in]
    {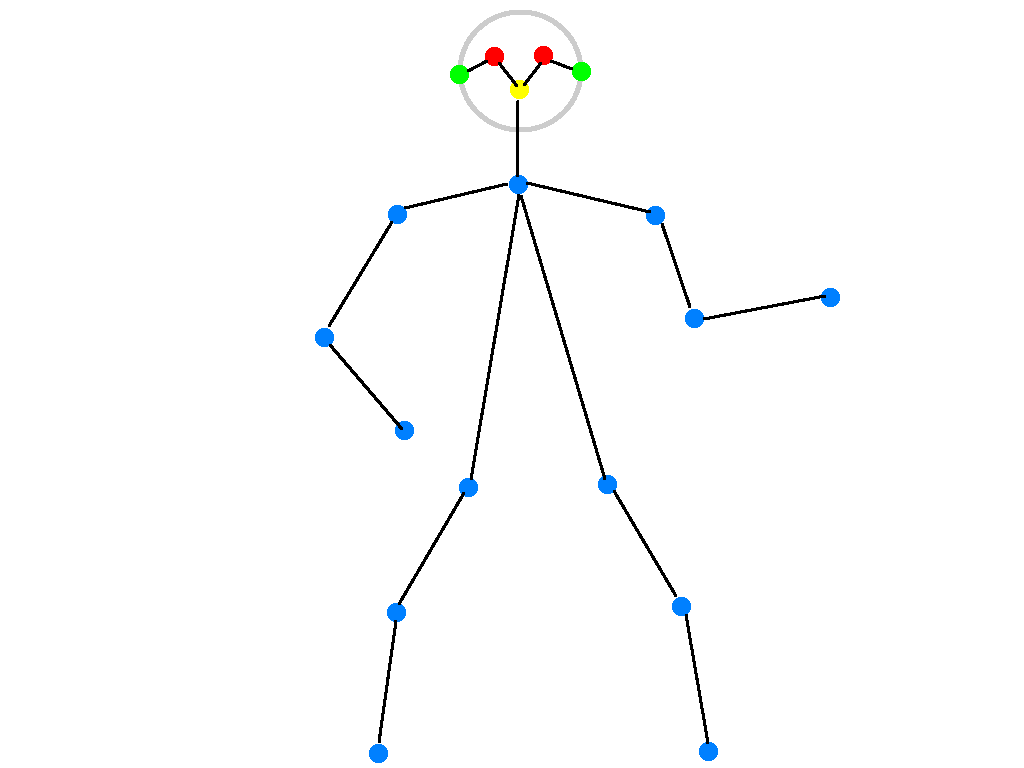}} &
  \subfloat[limbs that are used in feature extraction]{
    \includegraphics[trim={0cm 0 0 0cm},clip,width=1.5in]
    {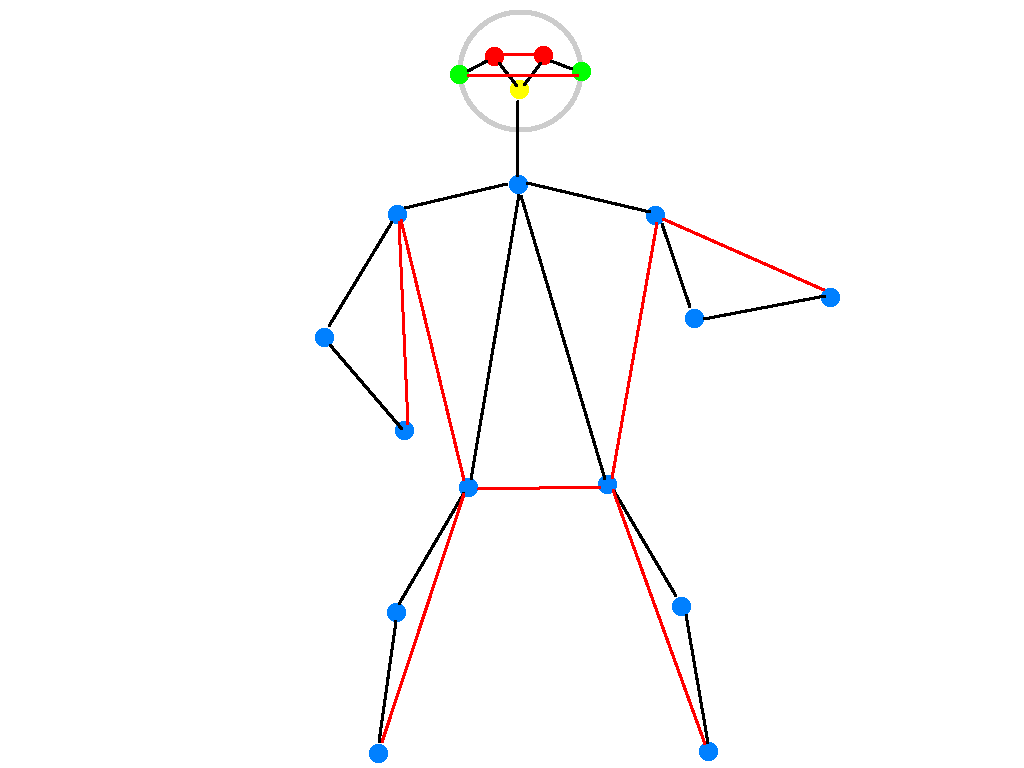}}
 \end{tabular}
    \caption{Illustration of the human skeleton. Both red lines and black lines are considered limbs in our context.}
    \label{fig:limb_angle}
\end{figure}

The second part of features in LMA, {\it effort component}, captures body motion characteristics. Based on the normalized pose, joints velocity $\hat{v}_i^t$, acceleration $\hat{a}_i^t$, and jerk $\hat{j}_i^t$ were computed as:
\begin{equation}
\begin{split}
&v_i^t = \frac{\hat{p}_i^{t+\tau} - \hat{p}_i^{t}}{\tau}\;,\;
a_i^t = \frac{{v}_i^{t+\tau} - {v}_i^{t}}{\tau}\;,\;
j_i^t = \frac{{a}_i^{t+\tau} - {a}_i^{t}}{\tau} \;,\\
&\hat{v}_i^t = \norm{v_i^t},\;
\hat{a}_i^t = \norm{a_i^t},\;
\hat{j}_i^t = \norm{j_i^t}\;.
\end{split}
\end{equation}
Angles, angular velocity, and angular acceleration between each pair of limbs (Fig.~\ref{fig:limb_angle}) were calculated for each pose:
\begin{equation}
\begin{split}
\theta^t(i,j,m,n) &= \arccos \left( \frac{(\hat{p}_i^t - \hat{p}_j^t) \cdot (\hat{p}_m^t - \hat{p}_n^t)}{\lVert \hat{p}_i^t - \hat{p}_j^t\lVert  \lVert{\hat{p}_m^t - \hat{p}_n^t}\lVert 
}\right) \;,\\
\omega_k^t(i,j,m,n) &= \frac{\theta^{t+\tau}(i,j,m,n) - \theta^t(i,j,m,n)}{\tau} \;,\\ 
\alpha_k^t(i,j,m,n) &= \frac{\omega^{t+\tau}(i,j,m,n) - \omega^t(i,j,m,n)}{\tau} \;.
\end{split}
\end{equation}
We computed velocity, acceleration, jerk, angular velocity, and angular acceleration of joints with $\tau = 15$. Empirically, features become less effective when $\tau$ is too small ($1\sim2$) or too large ($>$ 30).

The third part of features in LMA, {\it shape component}, captures body shape. For $f_{19}$, $f_{20}$, $f_{21}$, $f_{22}$, and $f_{23}$, the area of bounding box that contains corresponding joints is used to approximate volume.

Finally, all features are summarized by their basic statistics (maximum, minimum, mean, and standard deviation, denoted as $f_i^{\text{max}}$, $f_i^{\text{min}}$, $f_i^{\text{mean}}$, and $f_i^{\text{std}}$, respectively) over time.

\revise{With all LMA features combined, each skeleton sequence can be represented by a $2,216$-D feature vector. We further build classification and regression models for bodily expression recognition tasks. Because some measurements in our feature set can be linearly correlated and features can be missing, we choose the random forest for our classification and regression task. Specifically, we impute missing feature values with a large number ($1,000$ in our case). We then search model parameters with cross validation on the combined set of training and validation. Finally, we use the selected best parameter to retrain a model on the combined set.}

\subsubsection{Spatial Temporal Graph Convolutional Network}
\revise{
Besides handcrafted LMA features, we experimented with an end-to-end feature learning method. Following~\citep{yan2018spatial}, human body landmarks can be constructed as a graph with their natural connectivity. Considering the time dimension, a skeleton sequence could be represented with a spatiotemporal graph.\break Graph convolution in~\citep{kipf2016semi} is used as building blocks in ST-GCN. ST-GCN was originally proposed for skeleton action recognition. In our task, each skeleton sequence is first normalized between $0$ and $1$ with the largest bounding box of skeleton sequence. Missing joints are filled with zeros. We used the same architecture as in~\citep{yan2018spatial} and trained on our task with binary cross-entropy loss and mean-squared-error loss. Our learning objective $\mathcal{L}$ can be written as:}
\begin{equation}
\begin{aligned}
    \mathcal{L}_{\text{cat}} &= \sum_{i=1}^{26} y_i^{\text{cat}} \log x_i + (1-y_i^{\text{cat}}) \log(1-x_i^{\text{cat}}) \;,\\
    \mathcal{L}_{\text{cont}} &= \sum_{i=1}^{3} (y_i^{\text{cont}} - x_i^{\text{cont}})^2 \;,\\
    \mathcal{L} &= \mathcal{L}_{\text{cat}} + \mathcal{L}_{\text{cont}}\;,
\end{aligned}
\label{eq:loss}
\end{equation}
\revise{where $x_i^{\text{cat}}$ and $y_i^{\text{cat}}$ are predicted probability and ground truth, respectively, for the $i$-th categorical emotion, and $x_i^{\text{cont}}$ and $y_i^{\text{cont}}$ are model prediction and ground truth, respectively, for the $i$-th dimensional emotion.}

\subsection{Learning from Pixels}
Essentially, bodily expression is expressed through body activities. Activity recognition is a popular task in computer vision. The goal is to classify human activities, like sports and housework, from videos. In this subsection, we use four classical human activity recognition methods to extract features~\citep{kantorov2014efficient,simonyan2014two,wang2016temporal,carreira2017quo}. Current state-of-the-art results of activity recognition are achieved by two-stream network-based deep-learning methods~\citep{simonyan2014two}. Prior to that, trajectory-based handcrafted features are shown to be efficient and robust~\citep{wang2011action,wang2013action}. 

\subsubsection{Trajectory based Handcrafted Features}
The main idea of trajectory-based feature extraction is selecting extended image features along point trajectories. Motion-based descriptors, such as histogram of flow (HOF) and motion boundary histograms (MBH) \citep{dalal2006human}, are widely used in activity recognition for their good performance~\citep{wang2011action,wang2013action}. Common trajectory-based activity recognition has the following steps: 1) computing the dense trajectories based on optical flow; 2) extracting descriptors along those dense trajectories; 3) encoding dense descriptors by Fisher vector~\citep{perronnin2007fisher}; and 4) training a classifier with the encoded histogram-based features. 

In this work, we cropped each instance from raw clips with a fixed bounding box that bounds the character over time. We used the implementation in~\citet{kantorov2014efficient} to extract trajectory-based activity features\footnote{\url{https://github.com/vadimkantorov/fastvideofeat}}. We trained 26 SVM classifiers for the binary categorical emotion classification and three SVM regressors for the dimensional emotion regression. We selected the penalty parameter based on the validation set and report results on the test set.

\subsubsection{Deep Activity Features}
Two-stream network-based deep-learning methods learn to extract features in an end-to-end fashion~\cite{simonyan2014two}. A typical model of this type contains two convolutional neural networks (CNN). One takes static images as input and the other takes stacked optical flow as input. The final prediction is an averaged ensemble of the two networks. \revise{In our task, we used the same learning objective of $\mathcal{L}$ as defined in Eq.~\ref{eq:loss}.}

\begin{figure*}[ht!]
\centering
  \begin{tabular}{ccc}
  \subfloat[$f_{13}^{\text{mean}}$]{
    \includegraphics[height=1.5in]{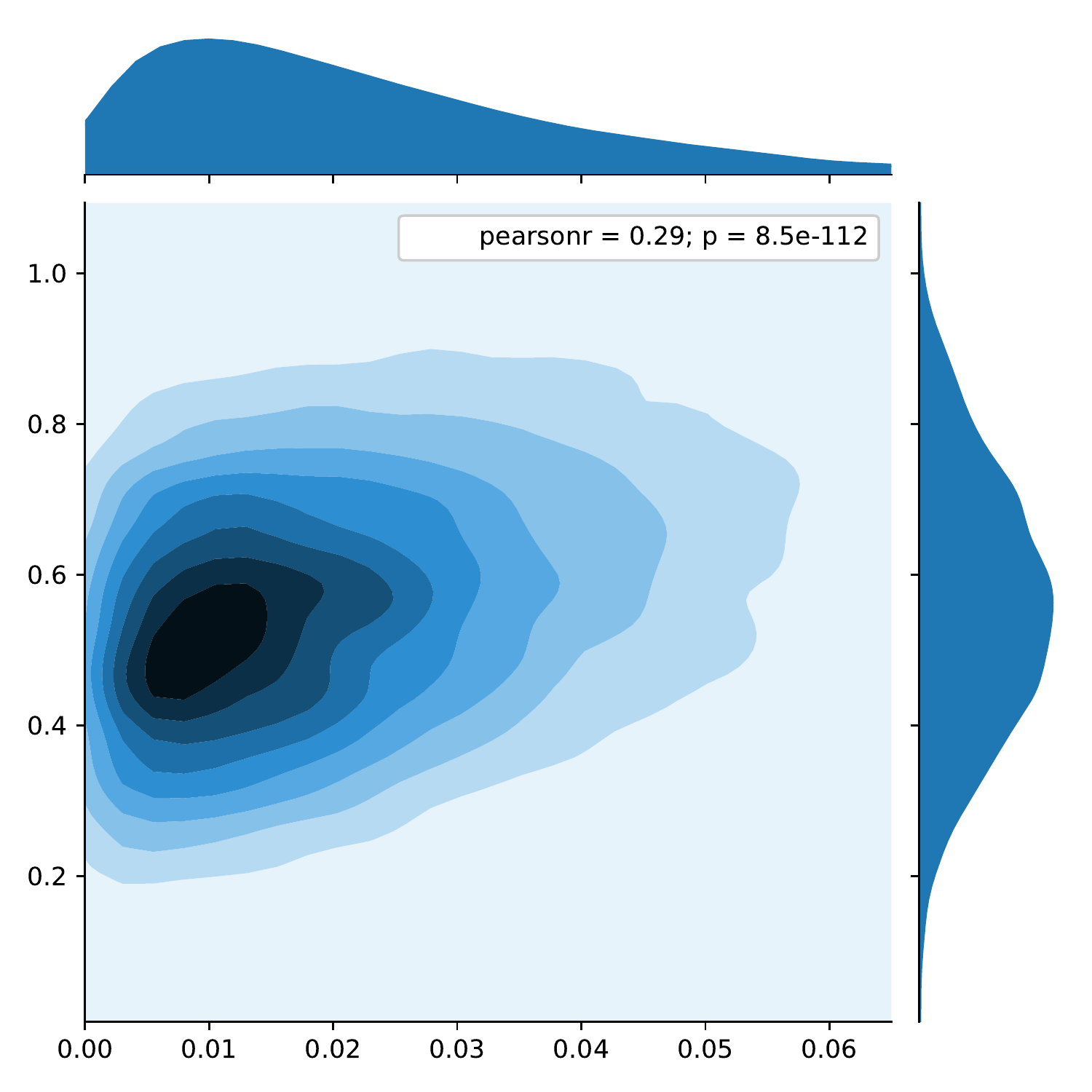}}&
  \subfloat[$f_{16}^{\text{max}}$]{
    \includegraphics[height=1.5in]{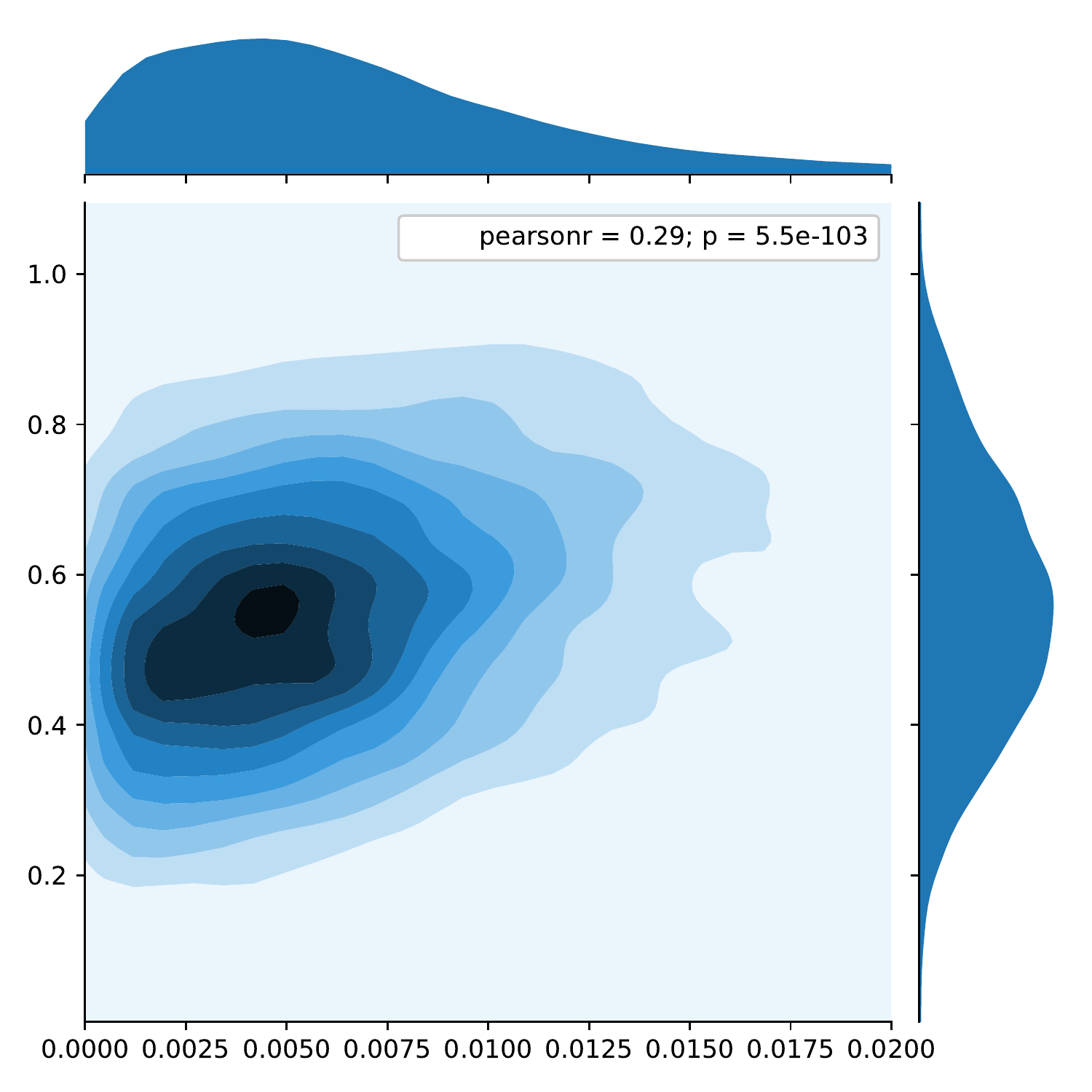}}&
  \subfloat[$f_{16}^{\text{mean}}$]{
    \includegraphics[height=1.5in]{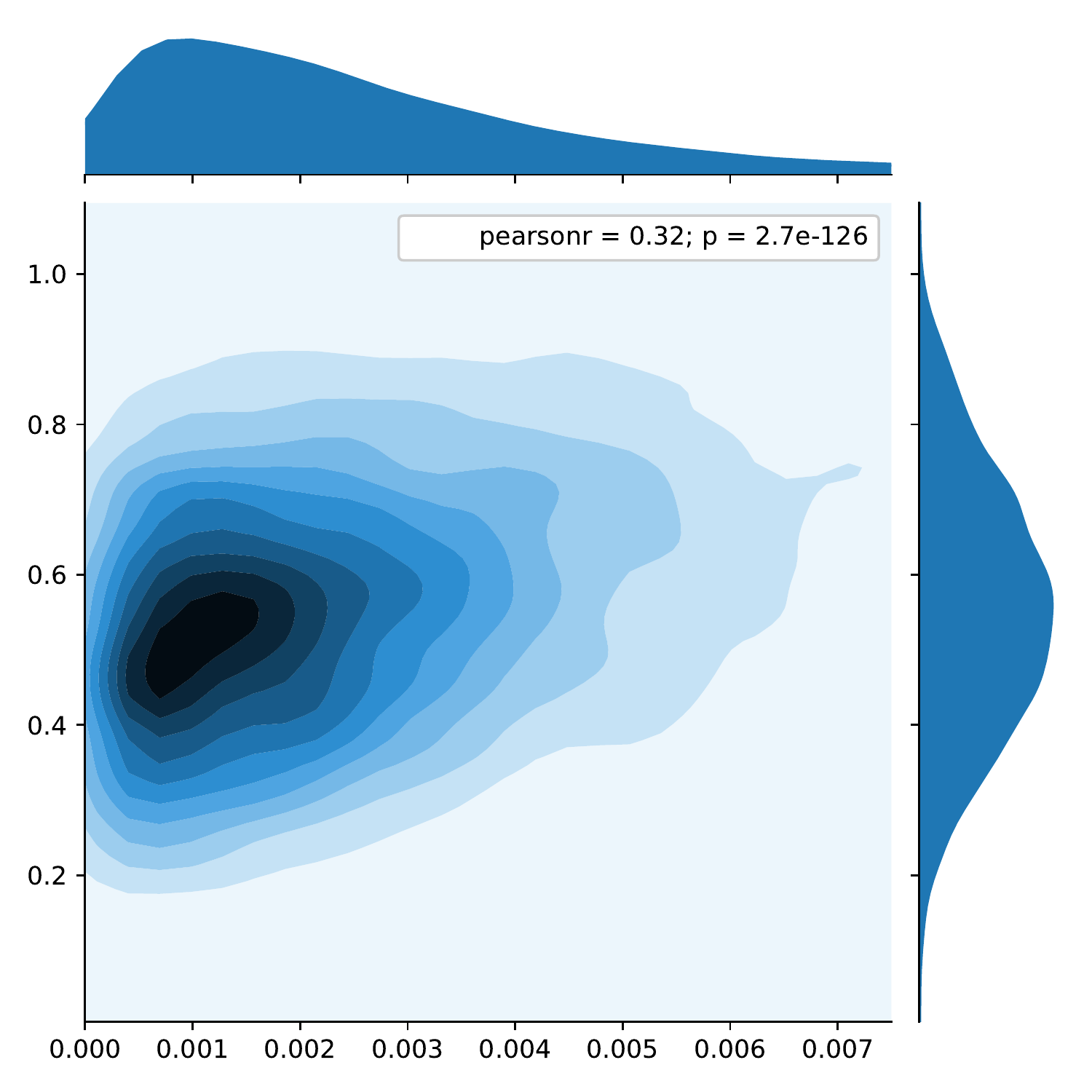}}\\
  \subfloat[$f_{30}^{\text{mean}}$]{
    \includegraphics[height=1.5in]{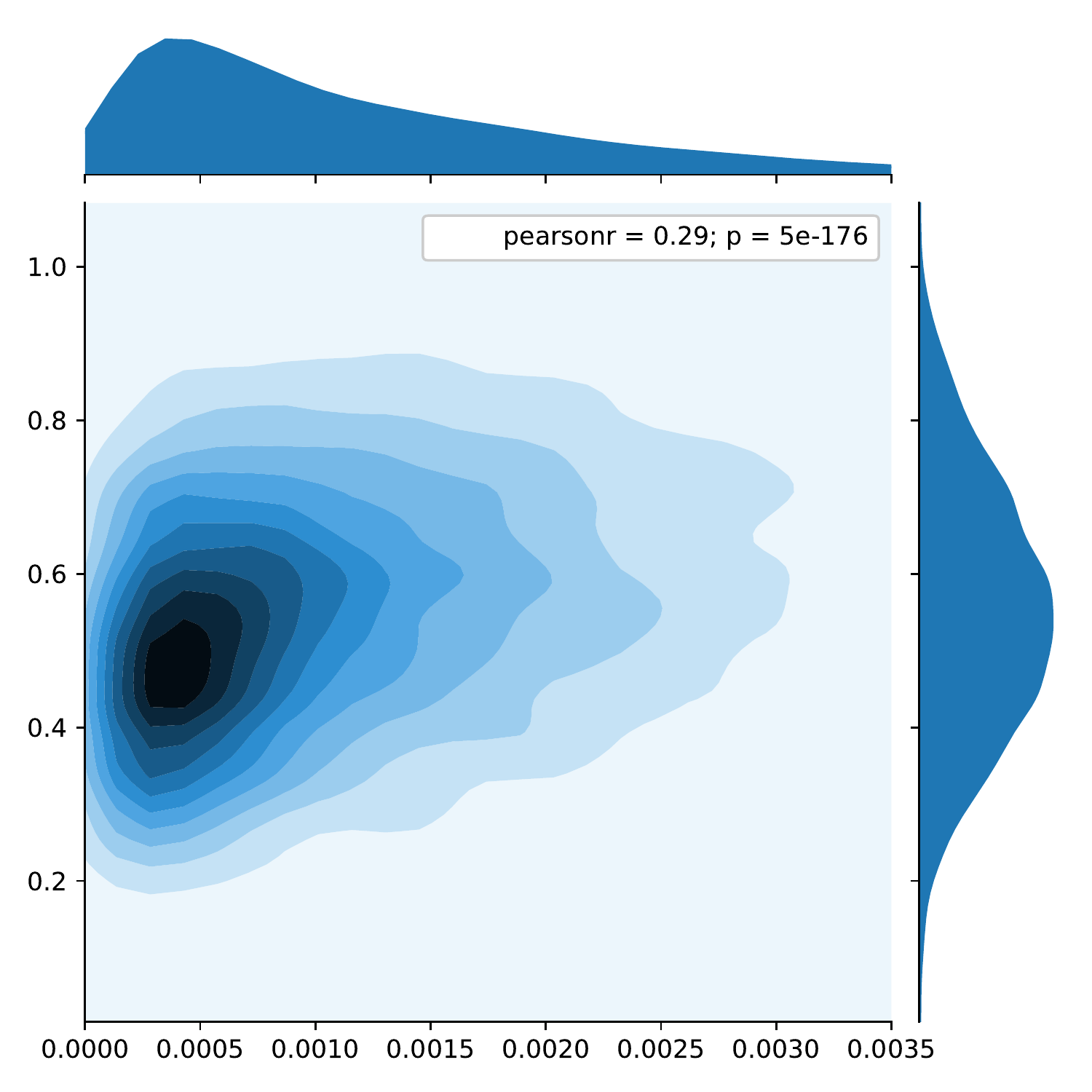}}&
  \subfloat[$f_{31}^{\text{mean}}$]{
    \includegraphics[height=1.5in]{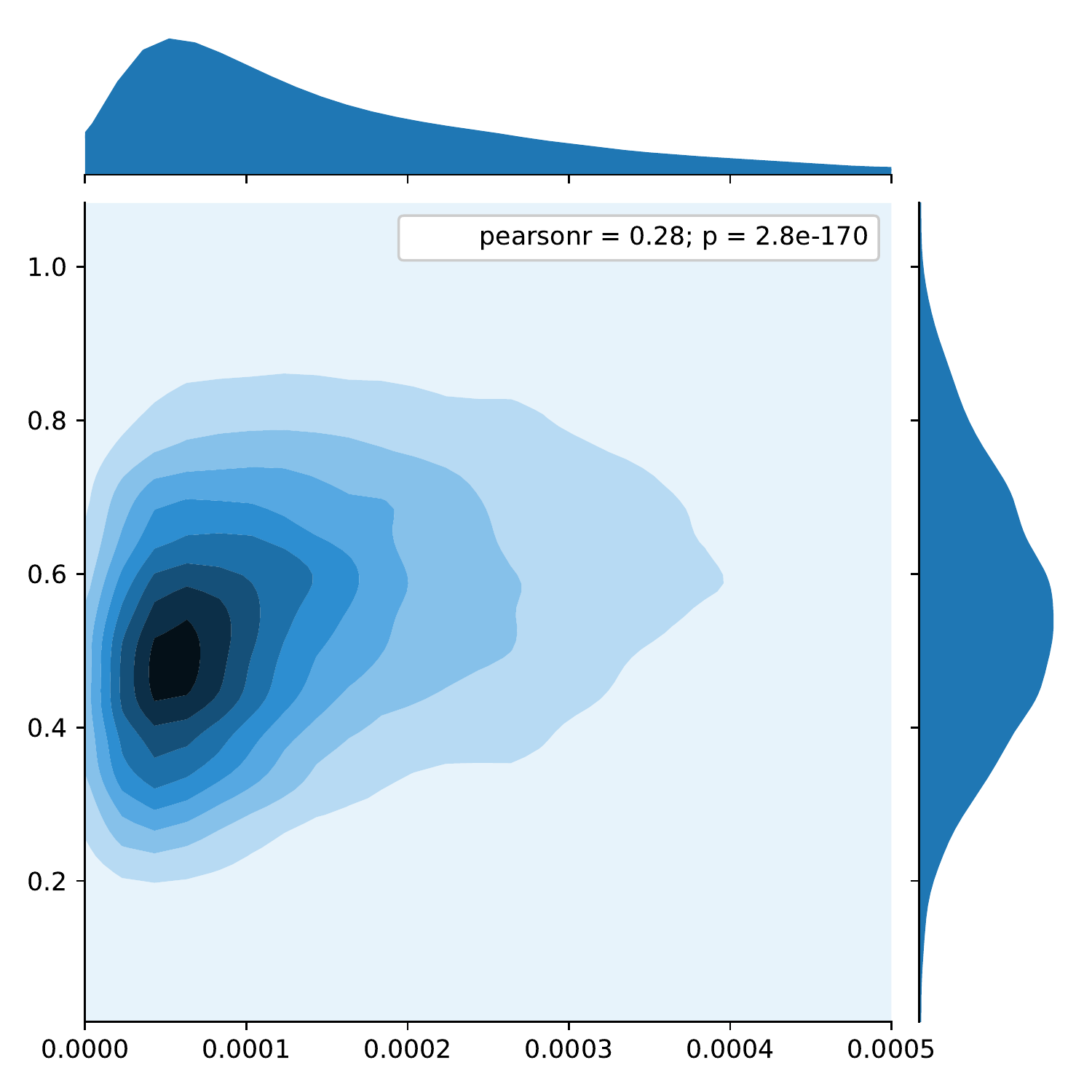}}&
  \subfloat[$f_{33}^{\text{mean}}$]{
    \includegraphics[height=1.5in]{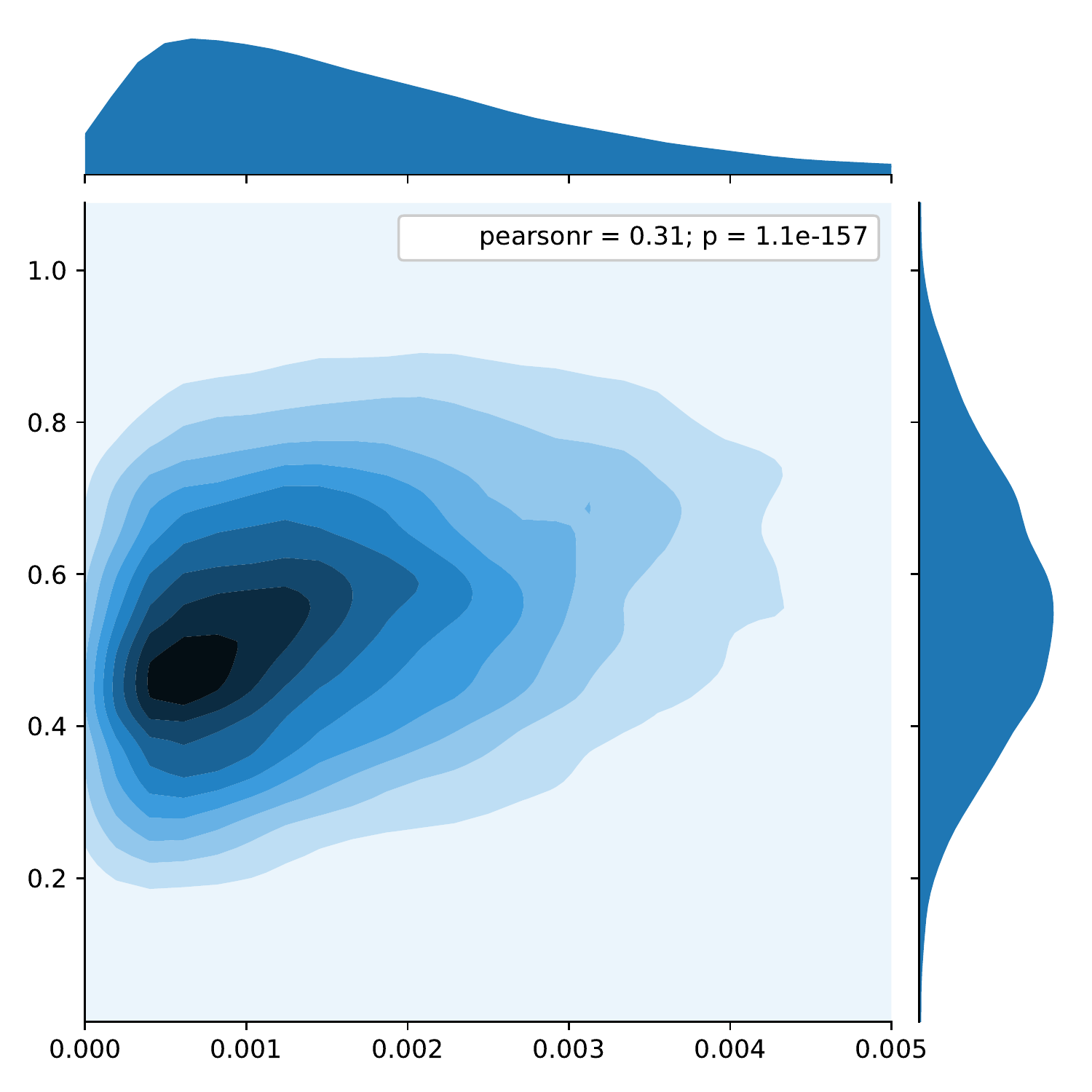}}\\
  \subfloat[$f_{34}^{\text{mean}}$]{
    \includegraphics[height=1.5in]{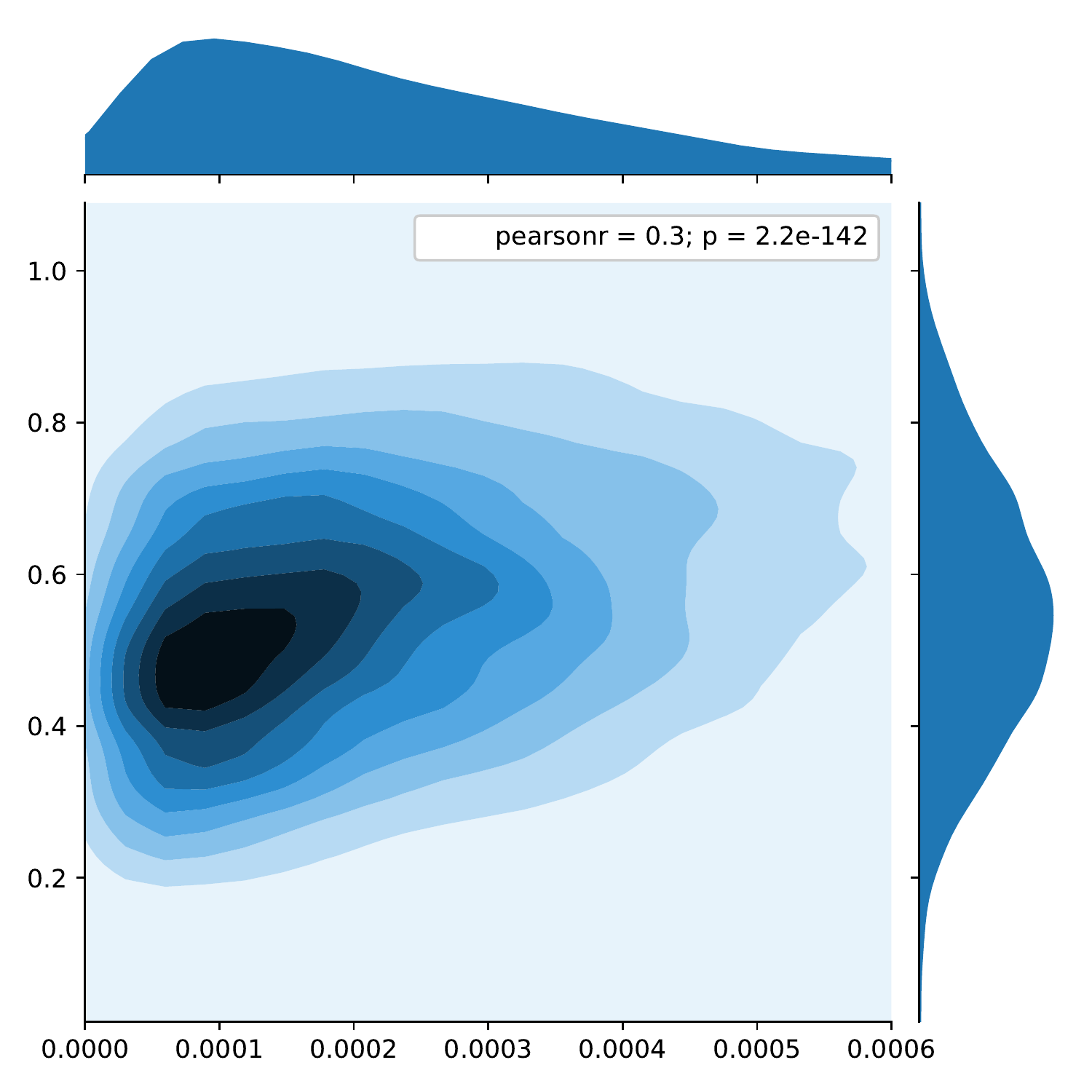}}&
  \subfloat[$f_{40}^{\text{max}}$]{
    \includegraphics[height=1.5in]{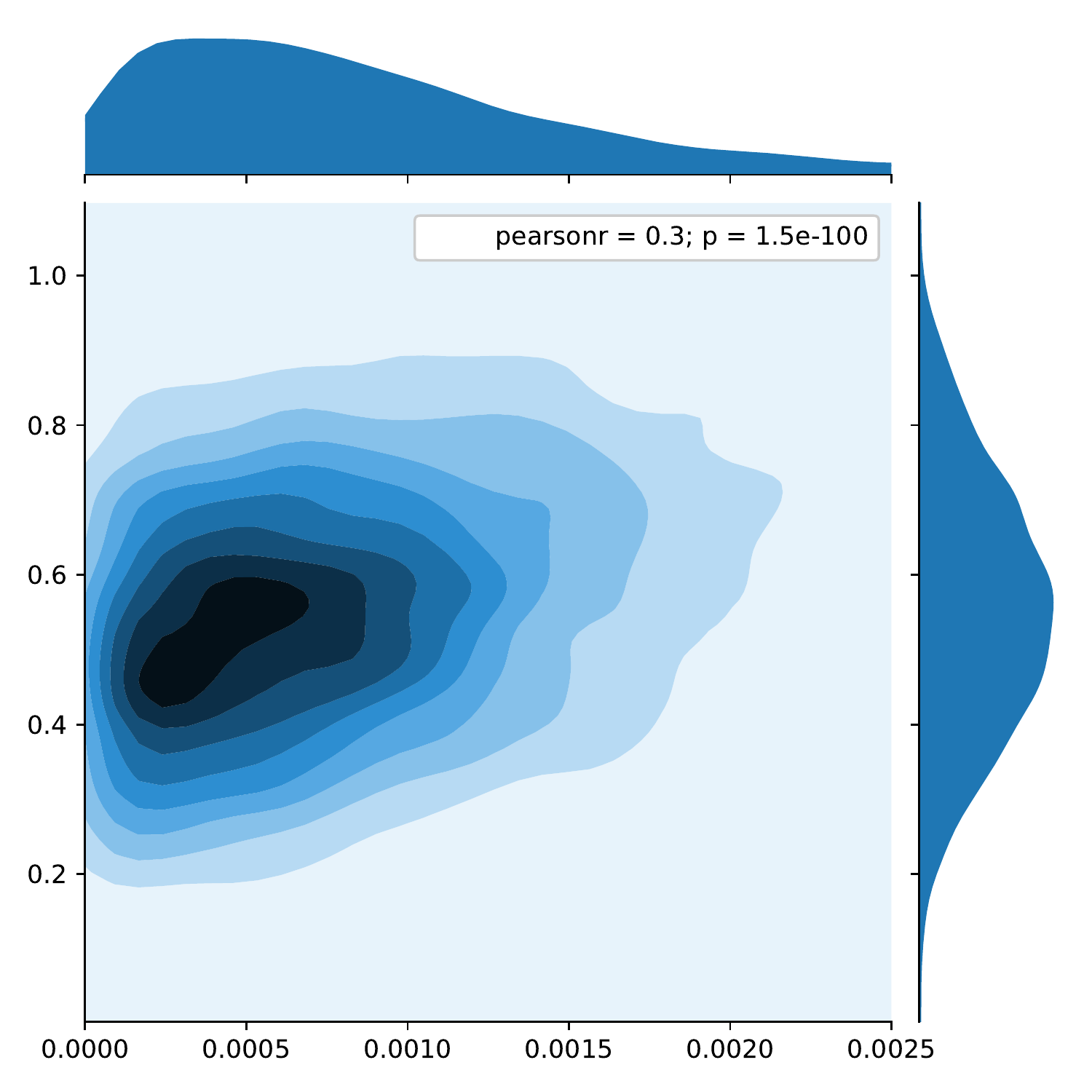}}&
  \subfloat[$f_{40}^{\text{mean}}$]{
    \includegraphics[height=1.5in]{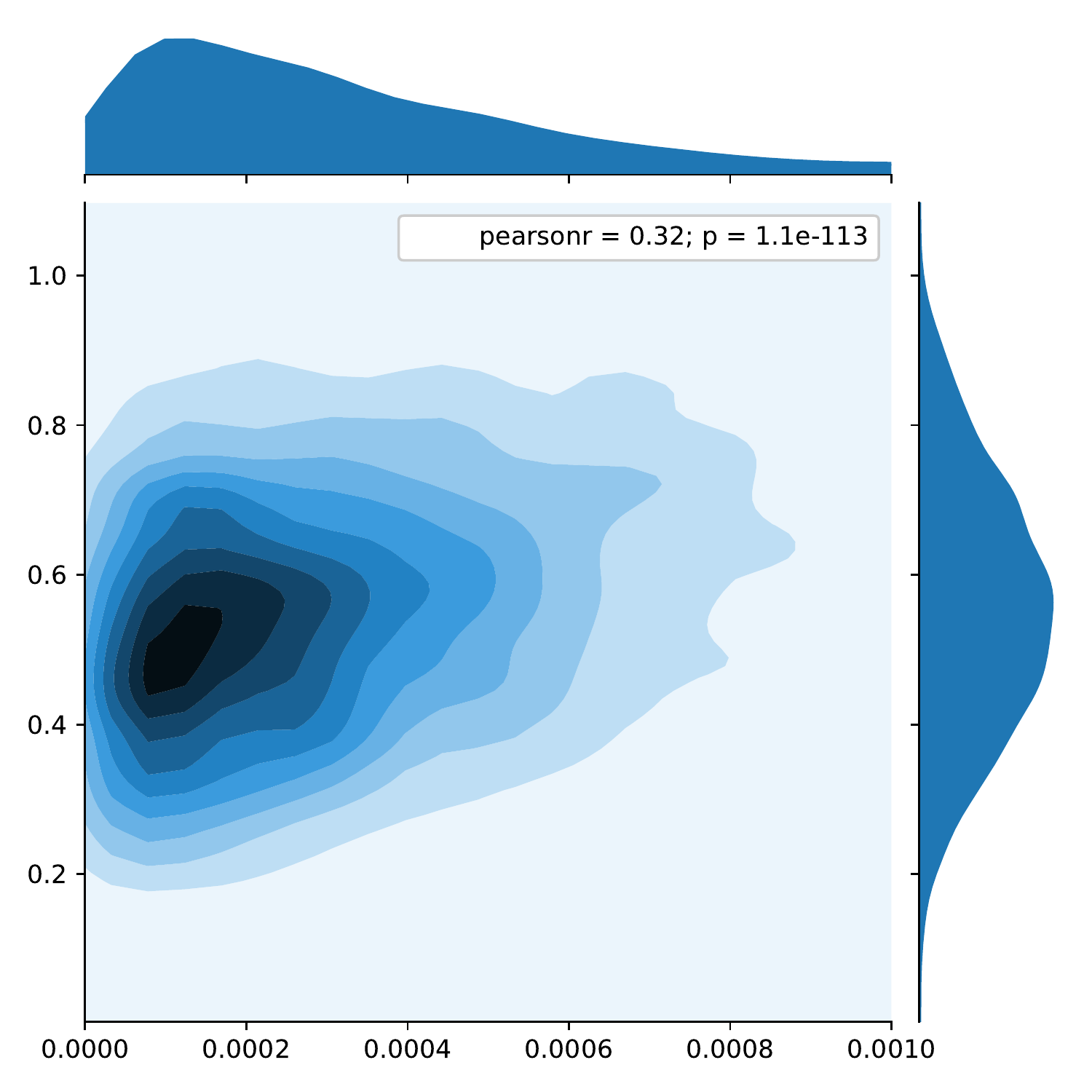}}
   \end{tabular}
  
   \caption{Kernel density estimation plots on selected LMA features that have high correlation with arousal.}
    \label{fig:lma_sig_feature}
\end{figure*}

We implemented two-stream networks in PyTorch\footnote{\url{http://pytorch.org/}}. We used 101-layer ResNet as~\citep{he2016deep} as our network architecture. Optical flow was computed via TVL1 optical flow algorithm \citep{zach2007duality}. Both image and optical flow were cropped with the instance body centered. Since emotion understanding could be potentially related to color, angle, and position, we did not apply any data augmentation strategies. The training procedure is identical to the work of~\citet{simonyan2014two}, where the learning rate is set to $0.01$. We used resnet-101 model pretrained on ImageNet to initialize our network weights. The training takes around 8 minutes for one epoch with an NVIDIA Tesla K40 card. \revise{The training time is short because only one frame is sampled input for each video in the RGB stream, and $10$ frames are concatenated along the channel dimension in the optical flow stream.} We used the validation set to choose the model of the lowest loss. \revise{We name this model as TS-ResNet101.}

\revise{Besides the original two-stream network, we also evaluated its two other state-of-the-art variants of action recognition. For temporal segment networks (TSN) \citep{wang2016temporal}, each video is divided into $K$ segments. One frame is randomly sampled for each segment during the training stage. Video classification result is averaged over all sampled frames. In our task, learning rate is set to $0.001$ and batch size is set to $128$. For two-stream inflated 3D ConvNet (I3D) \citep{carreira2017quo}, 3D convolution replaces 2D convolution in the original two-stream network. With 3D convolution, the architecture can learn spatiotemporal features in an end-to-end fashion. This architecture also leverages recent advances in image classification by duplicating weights of pretrained image classification model over the temporal dimension and using them as initialization. In our task, learning rate is set to $0.01$ and batch size is set to $12$. Both experiments are conducted on a server with two NVIDIA Tesla K40 cards. Other training details are the same as the original work~\citep{wang2016temporal, carreira2017quo}.}

\subsection{Results}

\subsubsection{Evaluation Metrics}
We evaluated all methods on the test set. For categorical emotion, we used average precision (AP, area under precision recall curve) and area under receiver operating characteristic curve (ROC AUC) to evaluate the classification performance. For dimensional emotion, we used $R^2$ to evaluate regression performance. Specifically, a random baseline of AP is the proportion of the positive samples (P.P.). ROC AUC could be interpreted as the possibility of choosing the correct positive sample among one positive sample and one negative sample; a random baseline for that is $0.5$. \revise{To compare performance of different models, we also report mean $R^2$ score (m$R^2$) over three dimensional emotion, mean average precision (mAP), and mean ROC AUC (mRA) over $26$ categories of emotion. For the ease of comparison, we define emotion recognition score (ERS) as follows and use it to compare performance of different methods:}
\begin{equation}
    \text{ERS} = \frac{1}{2}\left (\text{m}R^2 + \frac{1}{2}(\text{mAP} + \text{mRA})\right )\;. 
\end{equation}

\subsubsection{LMA Feature Significance Test}

For each categorical emotion and dimension of VAD, we conducted linear regression tests on each dimension of features listed in Table~\ref{table:feature}. 
All tests were conducted using the BoLD training set.
We did not find strong correlations ($R^2 < 0.02$) over LMA features and emotion dimensions other than arousal, {\it i.e.}, categorical emotion and valence and dominance. Arousal, however, seems to be significantly correlated with LMA features. 
Fig.~\ref{fig:lma_sig_feature} shows the kernel density estimation plots of features with top $R^2$ on arousal. Hands-related features are good indicators for arousal. With hand acceleration, $f_{16}^{\text{mean}}$ alone, $R^2$ can be achieved as $0.101$. Other significant features for predicting arousal are hands velocity, shoulders acceleration, elbow acceleration, and hands jerk. 

\begin{table}[t!]
    \caption{\revise {Dimensional emotion regression and categorical emotion classification performance on the test set. m$R^2$ = mean of $R^2$ over dimensional emotions, mAP(\%)= average precision / area under precision recall curve (PR AUC) over categorical emotions, mRA(\%) = mean of area under ROC curve (ROC AUC) over categorical emotions, and ERS = emotion recognition score. Baseline methods: ST-GCN~\citep{yan2018spatial}, TF~\citep{kantorov2014efficient}, TS-ResNet101~\citep{simonyan2014two}, I3D~\citep{carreira2017quo}, and TSN~\citep{wang2016temporal}. }}
{\renewcommand{\arraystretch}{1.35}
    \begin{tabular}{|C{2cm}|C{1.4cm}|C{0.9cm}|C{0.9cm}|C{0.9cm}|}
    \hline
        \multirow{2}{*}{Model}  &  Regression &  \multicolumn{2}{c|}{Classification}  & \multirow{2}{*}{ERS}\\ \cline{2-4}
                    &  m$R^2$   & mAP       & mRA       &                \\ \hline
             \end{tabular}
}
\vskip 0.05in
{\renewcommand{\arraystretch}{1.35}
    \begin{tabular}{|C{2cm}|C{1.4cm}|C{0.9cm}|C{0.9cm}|C{0.9cm}|}
        \multicolumn{5}{l}{A Random Method based on Priors:} \\ \hline
        Chance      & $0$      & $10.55$    & $50$      & $0.151$      \\ \hline 
            \end{tabular}
}
\vskip 0.05in
{\renewcommand{\arraystretch}{1.35}
    \begin{tabular}{|C{2cm}|C{1.4cm}|C{0.9cm}|C{0.9cm}|C{0.9cm}|}
        \multicolumn{5}{l}{Learning from Skeleton:} \\ \hline
        ST-GCN       & $0.044$  & $12.63$    & $55.96$   & $0.194$      \\ \hline 
        LMA         & $\mathbf{0.075}$ & $\mathbf{13.59}$ & $\mathbf{57.71}$ & $\mathbf{0.216}$ \\ \hline
            \end{tabular}
            }
             \vskip 0.05in
{\renewcommand{\arraystretch}{1.35}
    \begin{tabular}{|C{2cm}|C{1.4cm}|C{0.9cm}|C{0.9cm}|C{0.9cm}|}
        \multicolumn{5}{l}{Learning from Pixels:}  \\ \hline
        TF          & $-0.008$ & $10.93$    & $50.25$   & $0.149$     \\ \hline
        TS-ResNet101 & $0.084$  & $\mathbf{17.04}$    & $62.29$   & $0.240$   \\ \hline
        I3D        & $\mathbf{0.098}$  & $15.37$    & $61.24$   & $0.241$     \\ \hline
        TSN         & $0.095$  & $17.02$    & $\mathbf{62.70}$   & $\mathbf{0.247}$     \\ \hline
        TSN-Spatial & $0.048$  & $15.34$    & $60.03$   & $0.212$     \\ \hline
        TSN-Flow    & $0.098$  & $15.78$    & $61.28$   & $0.241$              \\ \hline
    \end{tabular}
    }
    \label{table:model_performance}
\end{table}

\begin{figure*}[t!]
    \centering
    \includegraphics[width=0.83\linewidth]{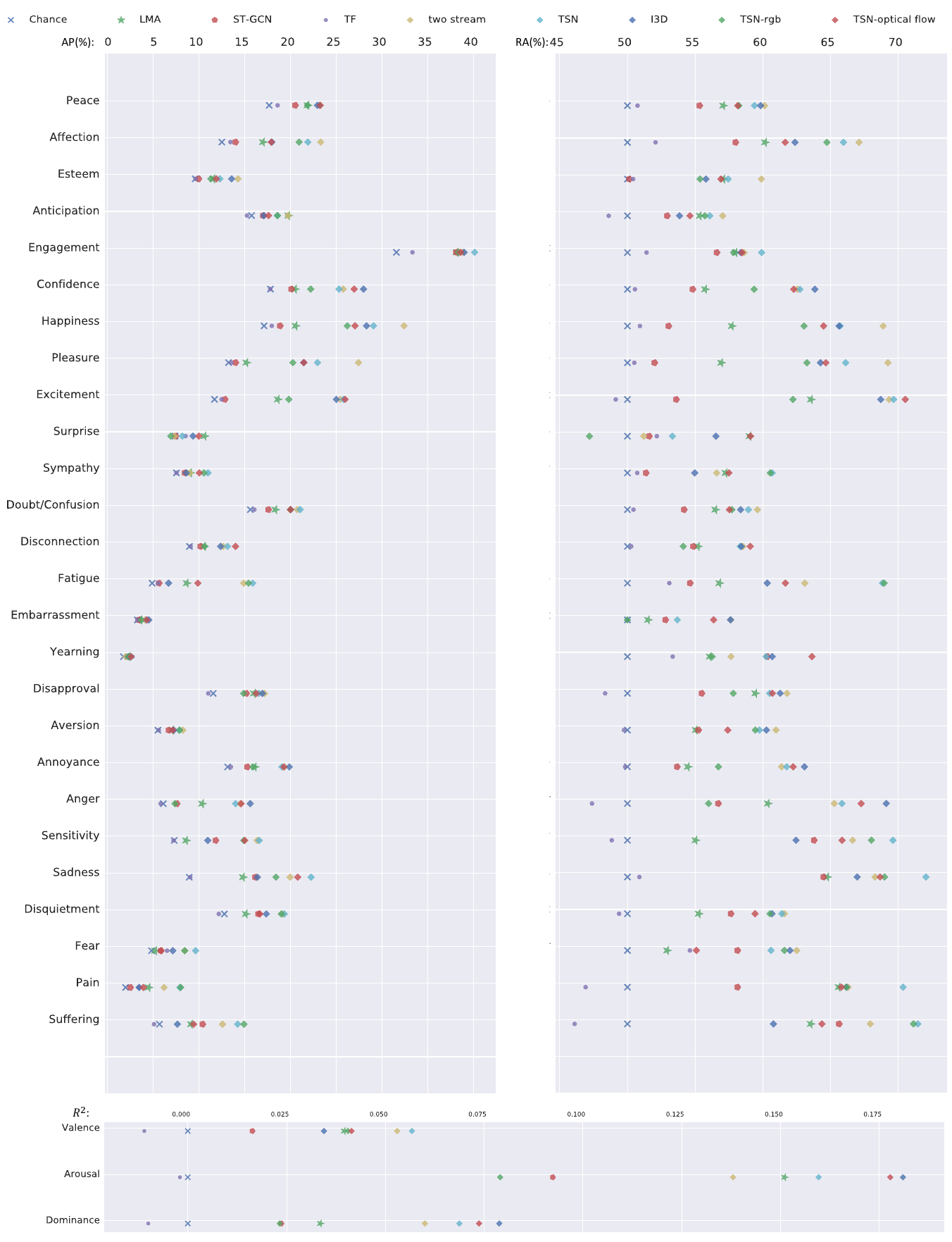}
    \caption{Classification performance (AP: average precision on the top left, RA: ROC AUC on the top right) and regression performance ($R^2$ on the bottom) of different methods on each categorical and dimensional emotion.}
    \label{fig:model_performance}
\end{figure*}

\subsubsection{Model Performance}
\revise{Table~\ref{table:model_performance} shows the results on the emotion classification and regression tasks. TSN achieves the best performance, with a mean $R^2$ of $0.095$, a mean average precision of $17.02\%$, a mean ROC AUC of $62.70\%$, and an ERS of $0.247$. Fig.~\ref{fig:model_performance} presents detailed metric comparisons over all methods of each categorical and dimensional emotion.}

\revise{For the pipeline that learns from the skeleton, both LMA and ST-GCN achieved above-chance results. Our handcrafted LMA features performs better than end-to-end ST-GCN under all evaluation metrics. For the pipeline that learns from pixels, trajectory-based activity features did not achieve above-chance results for both regression and classification task. 
However, two-stream network-based methods achieved significant\break above-chance results for both regression and classification tasks. As shown in Fig.~\ref{fig:model_performance} and Table~\ref{table:kappa}, most top-performance categories, such as affection, happiness, pleasure, excitement, sadness, anger, and pain, receive high agreement ($\kappa$) among annotators. Similar to the results from skeleton-based methods, two-stream network-based methods show better regression performance over arousal than for valence and dominance. However, as shown in Fig.~\ref{fig:worker_score}, workers with top $10\%$ performance has $R^2$ score of $0.48$, $-0.01$, and $0.16$ for valence, arousal, and dominance, respectively. Apparently, humans are best at recognizing valence and worst at recognizing arousal, and the distinction between human performance and model performance may suggest that there could be other useful features that the model has not explored.}

\subsection{Ablation Study}

\revise{To further understand the effectiveness of the two-stream-based model on our task, we conducted two sets of experiments to diagnose 1) if our task could leverage learned filters from pretrained activity-recognition model, and 2) how much a person's face contributed to the performance in the model. Since TSN has shown the best performance among all two-stream-based models, we conducted all experiments with TSN in this subsection. For the first set of experiments, we used different pretrained models, {\it i.e.}, image-classification model pretrained on ImageNet~\citep{deng2009imagenet} and action recognition model pretrained on Kinetics~\citep{kay2017kinetics}, to initialize TSN. Table~\ref{table:ablation_pretrain} shows the results for each case. The results demonstrate that initializing with pretrained ImageNet model leads to slightly better\break emotion-recognition performance. For the second set of experiments, we train TSN with two other different input types, {\it i.e.,} face only and faceless body. Our experiment in the last section crops the whole human body as the input. For face only, we crop the face for both spatial branch (RGB image) and temporal branch (optical flow) during both the training and testing stages. Note that for the face-only setting, orientation of faces in our dataset may be inconsistent, {\it i.e,} facing forward, facing backward, or facing to the side. For the faceless body, we still crop the whole body, but we also mask the region of face by imputing pixel value with a constant $128$. Table~\ref{table:ablation_face} shows the results for each setting. We can see from the results that the performance of using either the face or the faceless body as input is comparable to that of using the whole body as input. This result suggests both face and the rest of the body contribute significantly to the final prediction. Although the ``whole body'' setting of TSN performs better than any of the single model do, it does so by leveraging both facial expression and bodily expression.}

\begin{table}[t!]
    \caption{Ablation study on the effect of pretrained models.}
{\renewcommand{\arraystretch}{1.35}
    \begin{tabular}{|C{2cm}|C{1.4cm}|C{0.9cm}|C{0.9cm}|C{0.9cm}|}
    \hline
        \multirow{2}{*}{\shortstack{Pretrained\\ Model}}  &  Regression &  \multicolumn{2}{c|}{Classification}  & \multirow{2}{*}{ERS} \\ \cline{2-4}
            &  m$R^2$   & mAP       & mRA    &       \\ \hline 
        ImageNet            & $0.095$  & $17.02$    & $62.70$   & $0.247$  \\ \hline
        Kinetics    & $0.093$   & $16.77 $  &$62.53$ & $0.245$    \\ \hline
        \end{tabular}
        }
    \label{table:ablation_pretrain}
\end{table}

\begin{table}[t!]
    \caption{Ablation study on the effect of face.}
{\renewcommand{\arraystretch}{1.35}
    \begin{tabular}{|C{2cm}|C{1.4cm}|C{0.9cm}|C{0.9cm}|C{0.9cm}|}
    \hline
        \multirow{2}{*}{Input Type}  &  Regression &  \multicolumn{2}{c|}{Classification}  & \multirow{2}{*}{ERS}  \\ \cline{2-4}
                    &  m$R^2$   & mAP       & mRA   &        \\ \hline
        whole body         & $0.095$  & $17.02$    & $62.70$ & $0.247$       \\ \hline
        face only         &    $0.092$       &  $16.21$          & $62.18$        & $0.242$\\ \hline
        faceless body     & $0.088$         &   $16.61$          &  $62.30$       & $0.241$ \\ \hline
        \end{tabular}
        }
    \label{table:ablation_face}
\end{table}

\begin{table}[t!]
    \caption{Ensembled results.}
{\renewcommand{\arraystretch}{1.35}
    \begin{tabular}{|C{2cm}|C{1.4cm}|C{0.9cm}|C{0.9cm}|C{0.9cm}|}
    \hline
        \multirow{2}{*}{Model}  &  Regression &  \multicolumn{2}{c|}{Classification}  & \multirow{2}{*}{ERS}  \\ \cline{2-4}
                    &  m$R^2$   & mAP       & mRA   &        \\ \hline
        TSN-body         & $0.095$  & $17.02$    & $62.70$ & $0.247$       \\ \hline
        TSN-body + LMA   & $0.101$  & $16.70$    & $62.75$ & $0.249$       \\ \hline
        TSN-body + TSN-face         & $0.101$  & $17.31$    & $63.46$ & $0.252$       \\ \hline
        TSN-body + TSN-face + LMA         &    $0.103$       &  $17.14$          & $63.52$        & $0.253$\\ \hline
        \end{tabular}
        }
    \label{table:ensemble}
\end{table}

\begin{table}[t!]
  \begin{center}
  \caption{Retrieval results of our deep model. \break P@K(\%) = precision at K, R-P(\%)=R-Precision.}
{\setlength{\tabcolsep}{0.7em}\renewcommand{\arraystretch}{1.35}
    \begin{tabular}{| c !{\color{light-gray}\vrule}
 c!{\color{light-gray}\vrule}  c !{\color{light-gray}\vrule}
 c | 
 } 
    \hline
    {\bf Category} & {\bf P@10} & {\bf P@100} & {\bf R-P} \\ \hline 

 Peace & $40$ & $33$ & $28$ \\ \arrayrulecolor{light-gray}\hline\arrayrulecolor{black}
 Affection & $50$ & $32$  & $26$ \\ \arrayrulecolor{light-gray}\hline\arrayrulecolor{black}
 Esteem & $30$ & $14$ & $12$ \\ \arrayrulecolor{light-gray}\hline\arrayrulecolor{black}
 Anticipation & $30$ & $24$  & $20$ \\ \arrayrulecolor{light-gray}\hline\arrayrulecolor{black}
 Engagement & $50$ & $46$ & $42$ \\ \arrayrulecolor{light-gray}\hline\arrayrulecolor{black}
 Confidence & $40$ & $33$ & $31$ \\ \arrayrulecolor{light-gray}\hline\arrayrulecolor{black}
 Happiness & $30$ & $36$ & $31$ \\ \arrayrulecolor{light-gray}\hline\arrayrulecolor{black}
 Pleasure & $40$ & $25$ & $23$ \\ \arrayrulecolor{light-gray}\hline\arrayrulecolor{black}
 Excitement & $50$ & $41$ & $31$\\ \arrayrulecolor{light-gray}\hline\arrayrulecolor{black}
 Surprise & $20$ & $6$ & $8$ \\ \arrayrulecolor{light-gray}\hline\arrayrulecolor{black}
 Sympathy & $10$ & $14$ & $12$ \\ \arrayrulecolor{light-gray}\hline\arrayrulecolor{black}
 Doubt/Confusion & $20$ & $33$ & $25$ \\ \arrayrulecolor{light-gray}\hline\arrayrulecolor{black}
 Disconnection & $20$ & $20$ & $18$ \\ \arrayrulecolor{light-gray}\hline\arrayrulecolor{black}
 Fatigue & $40$ & $20$  & $17$ \\ \arrayrulecolor{light-gray}\hline\arrayrulecolor{black}
 Embarrassment & $0$ & $5$ & $5$ \\ \arrayrulecolor{light-gray}\hline\arrayrulecolor{black}
 Yearning & $0$ & $2$ & $4$ \\ \arrayrulecolor{light-gray}\hline\arrayrulecolor{black}
 Disapproval & $30$ & $28$ & $22$ \\ \arrayrulecolor{light-gray}\hline\arrayrulecolor{black}
 Aversion & $10$ & $10$ & $11$  \\ \arrayrulecolor{light-gray}\hline\arrayrulecolor{black}
 Annoyance & $30$ & $28$ & $23$ \\ \arrayrulecolor{light-gray}\hline\arrayrulecolor{black} 
 Anger & $40$ & $24$  & $20$ \\ \arrayrulecolor{light-gray}\hline\arrayrulecolor{black}
 Sensitivity & $30$ & $19$ & $19$ \\ \arrayrulecolor{light-gray}\hline\arrayrulecolor{black}
  Sadness & $50$ & $34$ & $25$ \\ \arrayrulecolor{light-gray}\hline\arrayrulecolor{black}
 Disquietment & $10$ & $26$ & $25$ \\ \arrayrulecolor{light-gray}\hline\arrayrulecolor{black}
 Fear & $10$ & $8$ & $8$  \\ \arrayrulecolor{light-gray}\hline\arrayrulecolor{black}
 Pain & $20$ & $9$  & $12$ \\ \arrayrulecolor{light-gray}\hline\arrayrulecolor{black}
 Suffering & $10$ & $17$ & $18$ \\ \hline
 Average &$27$ &$23$ & $20$\\ \hline

    \end{tabular}
}
\label{table:retrieval}
  \end{center}      
\end{table}
\subsection{ARBEE: Automated Recognition of Bodily Expression of Emotion}
\revise{We constructed our emotion recognition system, ARBEE, by ensembling best models of different modalities. As suggested in the previous subsection, different modalities could provide complementary clues for emotion recognition. Concretely, we average the prediction from different models (TSN-body: TSN trained with whole body, TSN-face: TSN trained with face, and LMA: random forest model with LMA features) and evaluate the performance on the test set. Table~\ref{table:ensemble} shows the results of ensembled results. According to the table, combining all modalities, {\it i.e.,} body, face and skeleton, achieves the best performance. ARBEE is the average ensemble of the three models.}

We further investigated how well ARBEE retrieves instances in the test set given a specific categorical emotion as query. Concretely, we calculated precision at $10$, $100$, and R-Precision as summarized in Table \ref{table:retrieval}. R-Precision is computed as precision at $R$, where $R$ is number of positive samples. Similar to the classification results, happiness and pleasure can be retrieved with a rather high level of precision. 
\section{Conclusions and Future Work} \label{sec:conclusion}

We proposed a scalable and reliable video-data collection pipeline and collected a large-scale bodily expression dataset, the BoLD. We have validated our data collection via statistical analysis. To our knowledge, our effort is the first quantitative  investigation of human performance on emotional expression recognition with thousands of people, tens of thousands of clips, and thousands of characters. Importantly, we found significant predictive features regarding the computability of bodily emotion, \textit{i.e.,} hand acceleration for emotional expressions along the dimension of arousal. Moreover, for the first time, our deep model demonstrates decent generalizability for bodily expression recognition in the wild. 

Possible directions for future work are numerous. First, our model's regression performance of arousal is clearly better than that of valence, yet our analysis shows humans are better at recognizing valence. The inadequacy in feature extraction and modeling, especially for valence, suggests the need for additional investigation. Second, our analysis has identified demographic factors in emotion perception between different ethnic groups. Our current model has largely ignored these potentially useful factors. Considering characters' demographics in the inference of bodily expression can be a fascinating research direction. 
Finally, although this work has focused on bodily expression, the BoLD dataset we have collected has several other modalities useful for emotion recognition, including audio and visual context. An integrated approach to study these will likely lead to exciting real-world applications.


\begin{acknowledgements}
This material is based upon work supported in part by The Pennsylvania State University.
This work used the Extreme Science and Engineering Discovery Environment (XSEDE), which is supported by National Science Foundation grant No. ACI-1548562~\citep{towns2014xsede}.
The work was also supported through a GPU gift from the NVIDIA Corporation.
The authors are grateful to the thousands of 
Amazon Mechanical Turk independent contractors for their 
time and dedication in providing 
invaluable emotion ground truth labels for the 
video collection. Hanjoo Kim contributed in some of the discussions. 
Jeremy Yuya Ong supported the data collection and visualization effort. We thank Amazon.com, Inc. for supporting the expansion of this line of research.
\end{acknowledgements}

\bibliographystyle{spbasic}      
\bibliography{references}   


\end{document}